\documentclass[10pt,twocolumn,letterpaper]{article}

\usepackage[pagenumbers]{cvpr} 

\definecolor{cvprblue}{rgb}{0.21,0.49,0.74}
\usepackage[pagebackref,breaklinks,colorlinks,citecolor=cvprblue]{hyperref}
\usepackage{multirow}
\usepackage{graphicx}
\usepackage{amsmath}
\usepackage{pifont}
\usepackage{xcolor,colortbl}
\usepackage[accsupp]{axessibility}

\newcommand{\cmark}{\ding{51}}%
\def\paperID{852} 
\def\confName{CVPR}
\def\confYear{2025}

\newif\ifdraft
\draftfalse
\drafttrue

\usepackage{xcolor}
\definecolor{orange}{rgb}{1,0.5,0}
\definecolor{violet}{RGB}{150,0,170}

\ifdraft
 \newcommand{\MS}[1]{{\color{green}{\bf MS: #1}}}
 
 \newcommand{\YH}[1]{{\color{blue}{\bf YH: #1}}}
 
 \newcommand{\Yang}[1]{{\color{violet}{\bf Yang: #1}}}
 
\else
 \newcommand{\MS}[1]{}
 
 \newcommand{\YH}[1]{}
 
 \newcommand{\Yang}[1]{}
\fi

\newcommand{\bI}{\mathbf{I}}

\newcommand{\bF}{\mathbf{F}}

\definecolor{best}{HTML}{FFCCC9}
\definecolor{secondbest}{HTML}{FFFFD4}

\title{Hierarchical Flow Diffusion for Efficient Frame Interpolation}

\author{%
	{Yang Hai $^1$, \quad  Guo Wang $^1$, \quad Tan Su $^1$, \quad Wenjie Jiang $^1$, \quad Yinlin Hu $^{2}$} \\
	{\normalsize $^1$ Insta360 Research \quad \quad \quad \quad $^2$ MagicLeap} \\
}
\begin{document}

\twocolumn[{
\renewcommand\twocolumn[1][]{#1}
\maketitle
\vspace{-5mm}  
\begin{center}
    \setlength{\tabcolsep}{2pt}

\begin{tabular}{cccccc}
    \includegraphics[width=0.16\linewidth]{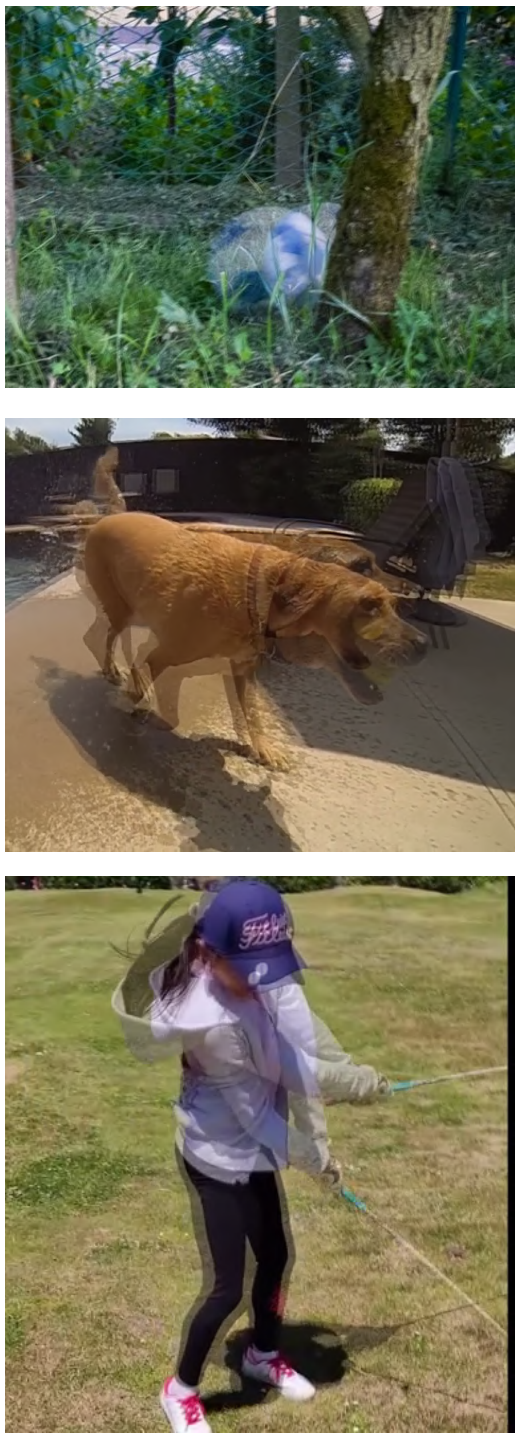} &
    \includegraphics[width=0.157\linewidth]{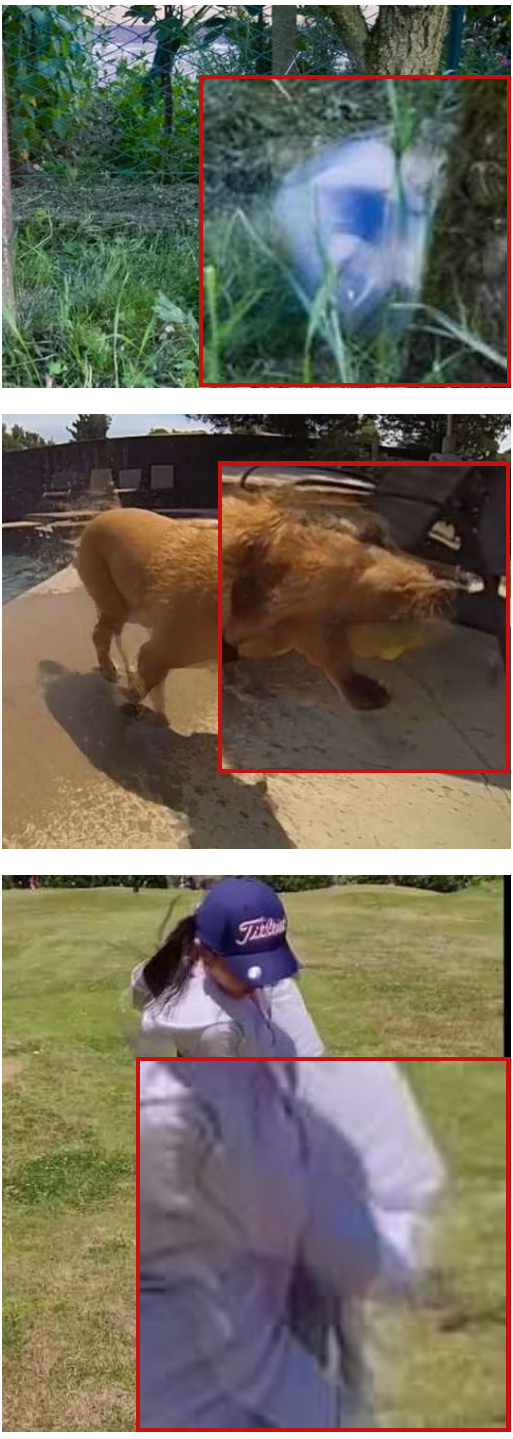} &
    \includegraphics[width=0.16\linewidth]{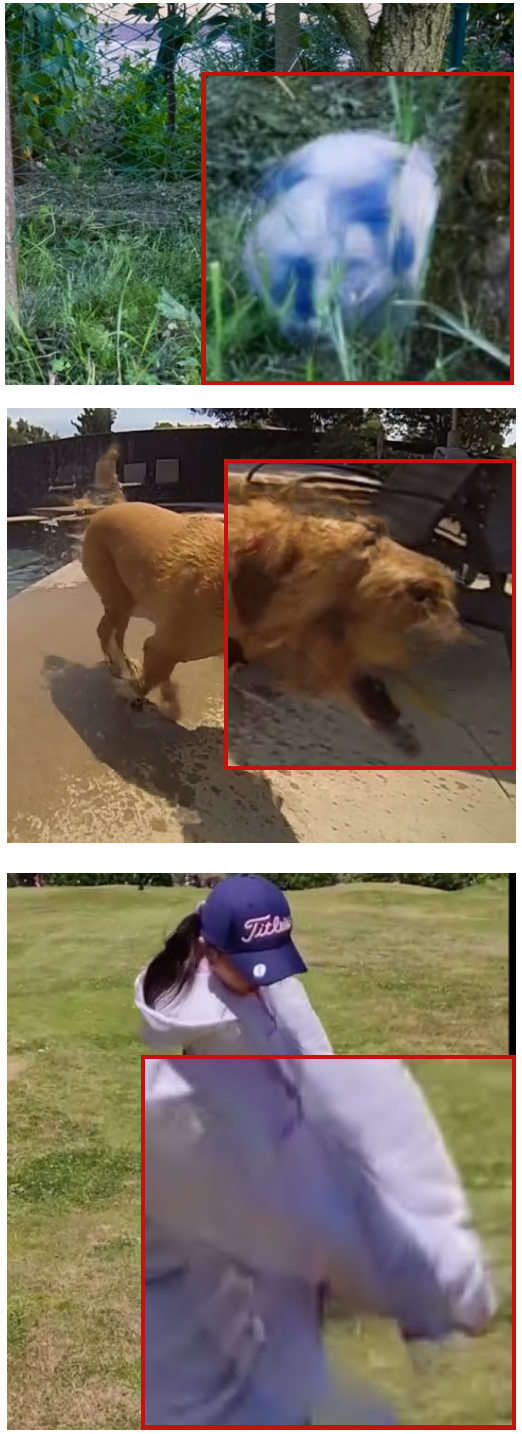} &
    \includegraphics[width=0.16\linewidth]{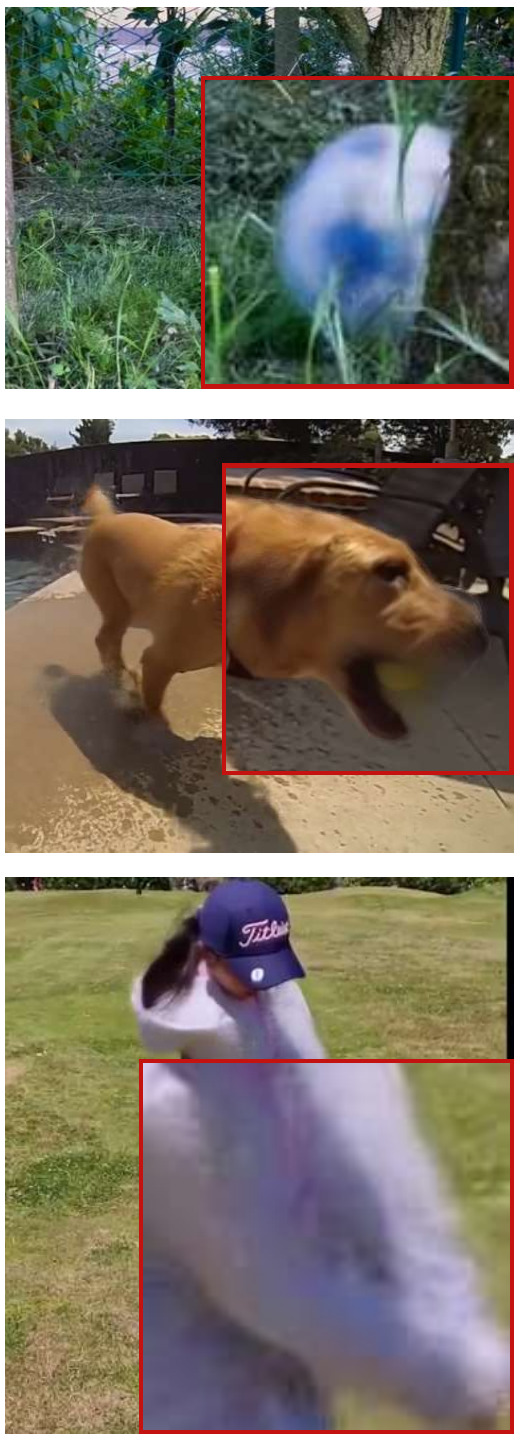} &
    \includegraphics[width=0.16\linewidth]{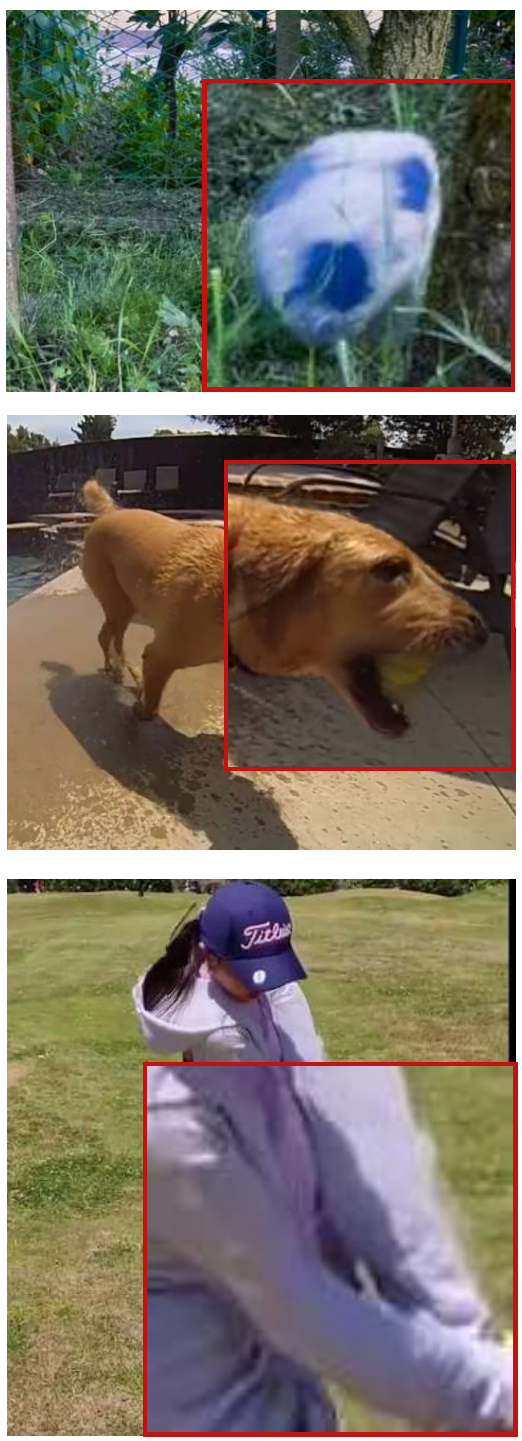} &
    \includegraphics[width=0.16\linewidth]{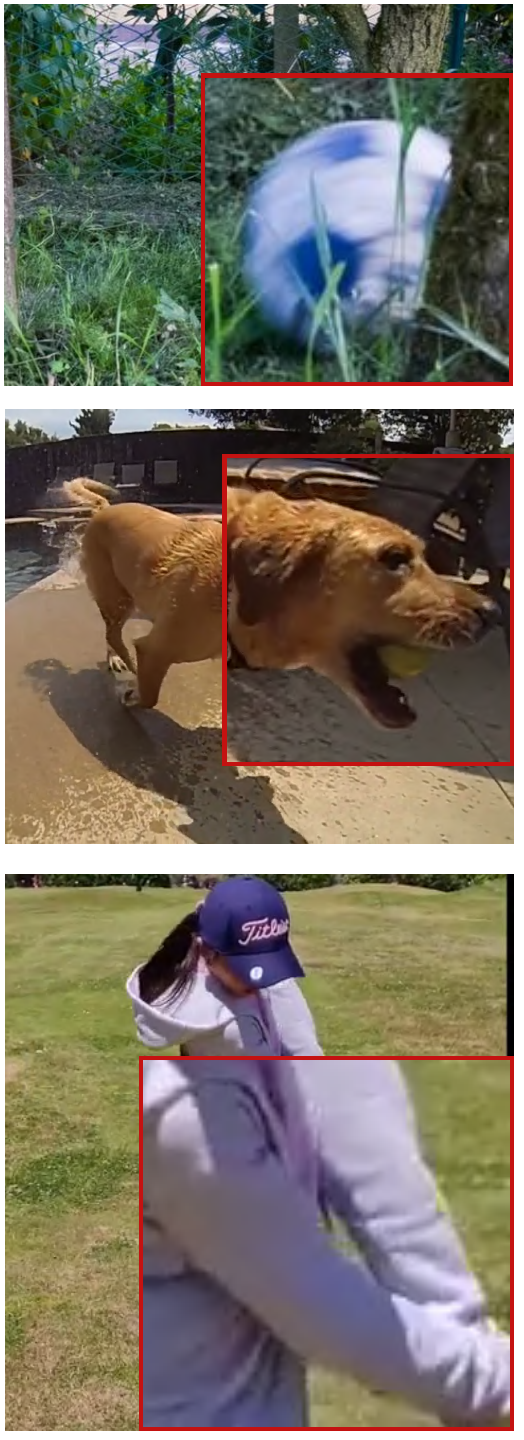} \\

         {\small Input overlay}  & {\small LDMVFI (8.3s)} & {\small CBBD (2.1s)}  & {\small SGM-VFI (0.19s)}   & {\small {\bf Ours} (0.20s)}   & {\small Ground truth} \\
\end{tabular}
    
    \vspace{-2mm}
    \captionof{figure}{
        \textbf{Different methods for video frame interpolation.} Most diffusion-based~\cite{ho2020ddpm, song2021ddim} interpolation methods (LDMVFI~\cite{danier2024ldmvfi}, CBBD~\cite{lyu2024cbbd}) still have a large gap from non-diffusion-based methods (SGM-VFI~\cite{liu2024sgm-vfi}), in both accuracy and efficiency. We propose a diffusion-based model that is 10+ times faster than other diffusion-based methods, and on par with SGM-VFI in efficiency. More importantly, we achieve significantly better accuracy than all baselines. Note how the details and large motions are missed in the baselines, but recovered with our method. We report the inference seconds on the same RTX-4090 GPU with a typical 1024$\times$1024 image pair.
    }
    \label{fig:teaser}
\end{center}
}]

\begin{abstract}
Most recent diffusion-based methods still show a large gap compared to non-diffusion methods for video frame interpolation, in both accuracy and efficiency. Most of them formulate the problem as a denoising procedure in latent space directly, which is less effective caused by the large latent space. We propose to model bilateral optical flow explicitly by hierarchical diffusion models, which has much smaller search space in the denoising procedure. Based on the flow diffusion model, we then use a flow-guided images synthesizer to produce the final result. We train the flow diffusion model and the image synthesizer end to end. Our method achieves state of the art in accuracy, and 10+ times faster than other diffusion-based methods. The project page is at: \url{https://hfd-interpolation.github.io}.
\end{abstract}

\begin{figure*}
    \centering
    \setlength{\tabcolsep}{10pt}
    \begin{tabular}{cc}
        \includegraphics[width=0.4\linewidth]{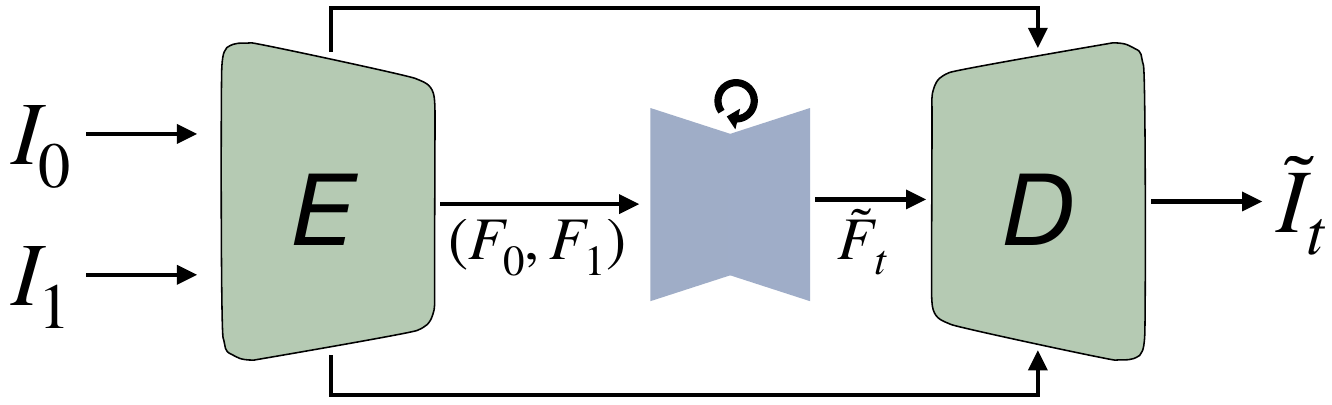} &
        \includegraphics[width=0.55\linewidth]{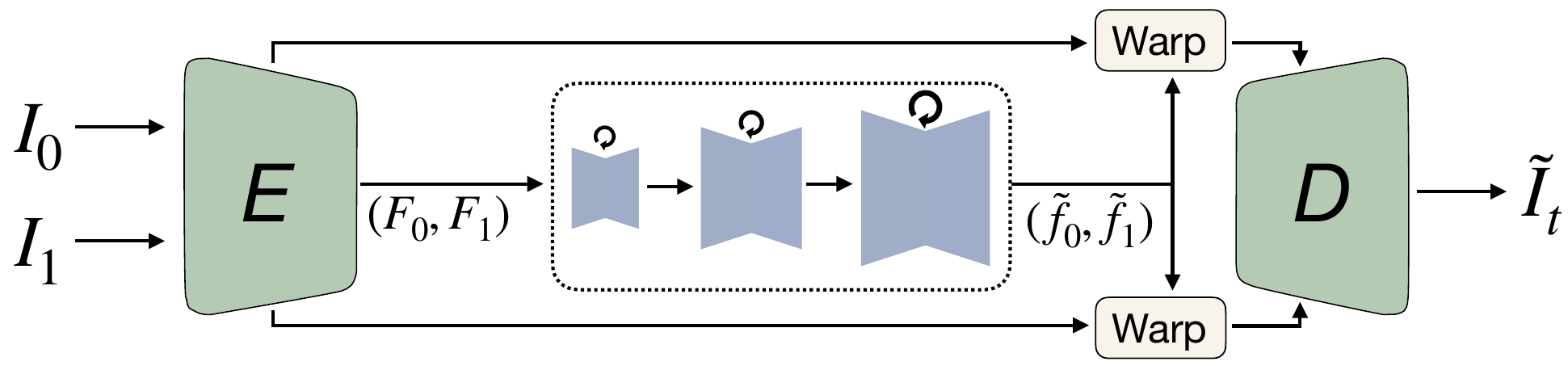} \\
        {\small (a) Baseline diffusion-based strategy} &
        {\small (b) Our method with hierarchical flow diffusion } \\
    \end{tabular}
    \vspace{-1.5mm}
    \caption{{\bf Different strategies with diffusion models for video frame interpolation.} Given an image pair ($I_0$, $I_1$), our goal is to predict the intermediate frame $\tilde{I}_t$.
    {\bf (a)} Most diffusion-based methods~\cite{danier2024ldmvfi, lyu2024cbbd, huang2024madiff} formulate the problem as a denoising process in the latent space ($\tilde{F}_t$) directly, and train the diffusion network and the encode-decoder (``E'' and ``D'') network separately. This strategy is less effective caused by the large latent space. On the other hand, this method cannot handle complex motions and large displacement.
    {\bf (b)} 
    We use a hierarchical strategy with explicit flow modeling. We first train a flow based encoder-decoder for image synthesizer with image pairs and the ground truth optical flow. Then, unlike most diffusion-based methods that denoise the latent space directly, we use a hierarchical diffusion model, conditioned on the encoder feature ($F_0$, $F_1$), to explicitly denoise optical flow from coarse to fine. We use the predicted bilateral flow ($\tilde{f}_0$, $\tilde{f}_1$) to warp image features for the synthesizer, and finally fine-tune the synthesizer and the diffusion models jointly.
    }
    \vspace{-2mm}
    \label{fig:arch_compare} 
\end{figure*}

\section{Introduction}
\label{src:intro}
Video frame interpolation aims to generate intermediate frames given a pair of consecutive frames from video. It is a fundamental video understanding task in computer vision~\cite{voleti2022mcvd}, and has many real applications such as slow-motion generation~\cite{jiang2018superslomo}, video compression~\cite{wu2018video}, and novel view synthesis~\cite{li2021neural}.

Existing methods have made great progress based on an encoder-decoder paradigm with bilateral flow as an intermediate supervision signal~\cite{kong2022ifrnet, huang2022rife, lu2022vfiformer, li2023amt, zhang2023ema_vfi, liu2024sgm-vfi, wu2024pervfi, reda2022film}. However, since predicting bilateral flow between two frames is an ill-posed problem with many possible solutions in essence, most of them can only produce an over-smoothed mean solution, as shown in Fig.~\ref{fig:teaser} with SGM-VFI.

Some recent methods try to use diffusion techniques~\cite{ho2020ddpm, song2021ddim, dhariwal2021diffusion} for video frame interpolation~\cite{danier2024ldmvfi, lyu2024cbbd, huang2024madiff}, which formulates the frame interpolation as a denoising process. Although these diffusion-based methods usually generate sharper image results, they suffer from several issues. First, most of them conduct the denoising directly in the latent space, which is less effective because of the large latent space. On the other hand, most of them cannot handle complex motions and large displacement, limited by the representation capability of diffusion networks.

To address these problems, we first train a flow based encoder-decoder for image synthesizer using image pairs and the ground truth optical flow. Then, unlike most diffusion-based methods that model the latent space directly, we use hierarchical diffusion models, conditioned on the encoder feature, to explicitly model optical flow from coarse to fine, which is very efficient and can handle large motions. Finally, we simply upsample the flow as the input of the image synthesizer, and jointly fine-tune the synthesizer and the hierarchical flow diffusion model, as illustrated in Fig.~\ref{fig:arch_compare}.

\begin{figure*}
    \centering
    \includegraphics[width=0.99\linewidth]{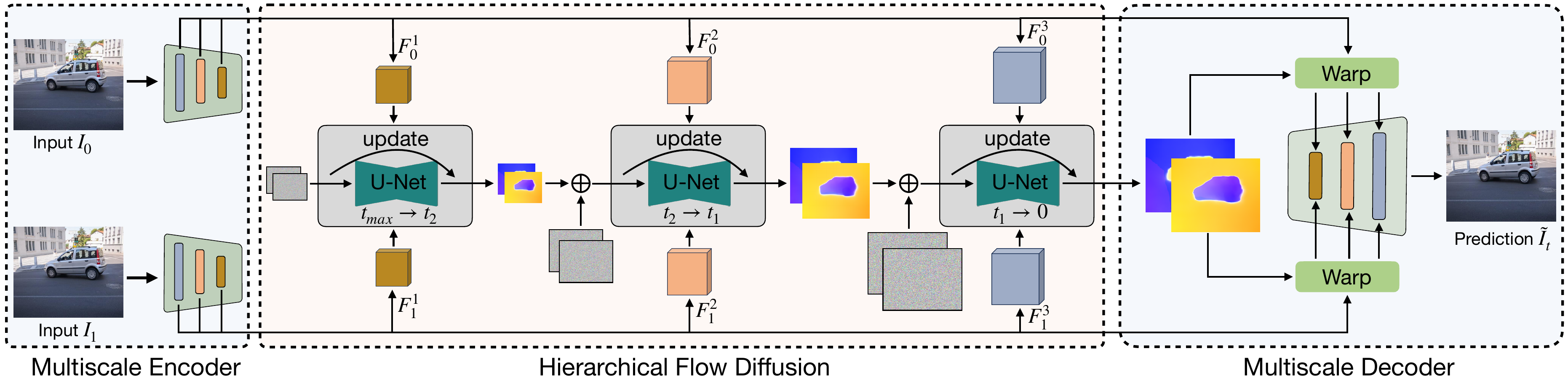} \\
    \vspace{-1.5mm}
    \caption{{\bf Overview of our method.}
    We first construct a flow-guided encoder-decoder with multiscale features as our image synthesizer, and then use diffusion to explicitly denoise optical flow in a coarse-to-fine manner, where the diffusion on each level will be conditioned on encoder features from the corresponding level. With the predicted intermediate optical flow, we use the flow to warp encoder features on each level, and use a multiscale decoder to synthesize the final target image.
    }
    \vspace{-3mm}
    \label{fig:overall}
\end{figure*}

We evaluate our method on multiple challenging benchmarks, including SNUFILM~\cite{choi2020snufilm}, Xiph~\cite{montgomery1994xiph}, DAVIS~\cite{perazzi2016davis}, and Vimeo~\cite{xue2019vimeo}. Our method achieves state of the art in accuracy, and 10+ times faster than other diffusion-based methods.

\section{Related Work}
\label{sec:relatedwork}

\noindent \textbf{Optical flow estimation} is a basic building block for video frame interpolation, and also a fundamental problem in computer vision, which aims to estimate pixel-level matches from the source image to the target image.
Traditionally, it is modeled as an energy optimization problem, built upon the assumption of brightness consistency and local smoothness~\cite{black1993framework, Fullflow_2016_cvpr, Hu_2016_CVPR, hu2017robust}.
Recently, learning-based methods have shown great progress, benefiting from large datasets~\cite{mayer2016large, sun2021autoflow, mehl2023spring} and advanced model architectures~\cite{sun2018pwc, teed2020raft, huang2022flowformer, hai2023scflow, hai2023pfcflow, wang2024sea}.
Some recent methods~\cite{saxena2024ddvm, luo2024flowdiffuser, nam2024diffmatch} apply diffusion models to optical flow, and generate promising results.
These diffusion models are specifically designed for optical flow estimation, and are trained with ground truth flow. However, in the context of video frame interpolation, we do not have ground truth flow. We propose a hierarchical flow diffusion model supervised by the pseudo flow predicted by a pretrained flow model.

\noindent \textbf{Video frame interpolation} has shown significant progress with the development of flow-based methods.
Most of them are based on an encoder-decoder paradigm with optical flow as an intermediate supervision signal, in either forward flow supervision~\cite{niklaus2020softmax, jin2023urpnet} or backward flow supervision~\cite{huang2022rife, reda2022film, kong2022ifrnet, li2023amt, zhang2023ema_vfi}. 
The most recent method, SGM-VFI~\cite{liu2024sgm-vfi}, combines the forward-flow and backward-flow techniques in a unified framework, and shows superior performance.
However, since the intermediate bilateral flow between two frames has many possible solutions in essence, and we do not have the ground truth bilateral flow for supervision, this type of method tends to produce an over-smoothed mean solution. We propose to only use bilateral flow as intermediate supervisions, and train a flow-guided image synthesizer at the same time, producing an end-to-end trainable system for video frame interpolation.

\noindent \textbf{Diffusion model} demonstrates great advantages for many image generation tasks, including text-to-image synthesis~\cite{dhariwal2021diffusion, rombach2022stablediffusion, peebles2023dit}, image restoration~\cite{yue2024resshift, zhou2023pyramid}, and image editing~\cite{hertz2023prompttoprompt, Kawar2023imagic}, as well as several high-level tasks, such as monocular depth estimation~\cite{ke2024marigold}, object detection~\cite{chen2023diffusiondet}, and image segmentation~\cite{chen2023Analogbits}.
Some recent methods~\cite{danier2024ldmvfi, lyu2024cbbd, huang2024madiff} apply diffusion models to video frame interpolation and show promising results.
However, most of them conduct the denoising procedure in latent space directly, which has a large search space and cannot handle complex motions and large displacement.
We propose a hierarchical diffusion model, which formulates the problem as denoising optical flow explicitly in a coarse-to-fine manner, achieving 10+ times faster than other diffusion-based methods.

\section{Approach}
\label{sec:approach}

Given a pair of consecutive frames, our goal is to generate the intermediate frame. Our method is built on top of the diffusion model~\cite{song2021ddim, ho2020ddpm}. However, unlike common diffusion-based methods, which are usually less efficient and cannot handle complex motions and large displacements, we introduce a hierarchical flow diffusion framework to address those problems.
Fig.~\ref{fig:overall} shows the overview of our framework. 

\subsection{Flow-Guided Image Synthesis}
\label{sec:auto-encoder}

We train a flow-guided encoder-decoder as our image synthesizer in the first stage. Formally, given a deep regressor $g$ with parameters $\Phi$, lets us write
\begin{equation}
    {\bI}_t =g({\bf I}_0, {\bf I}_1, f_0, f_1;\Phi),
\label{eq:local_prediction}
\end{equation}
where ${\bI}_t$ is the target synthesized image, ${\bf I}_0$ and ${\bf I}_1$ are the input images, and $f_0$ and $f_1$ are the corresponding optical flow from ${\bf I}_t$ to ${\bf I}_0$ and ${\bf I}_1$ respectively.
We implement $g$ with a multiscale encoder-decoder architecture~\cite{resnet}.

\begin{figure}
    \centering
    \setlength{\tabcolsep}{0pt}
    \includegraphics[width=0.99\linewidth]{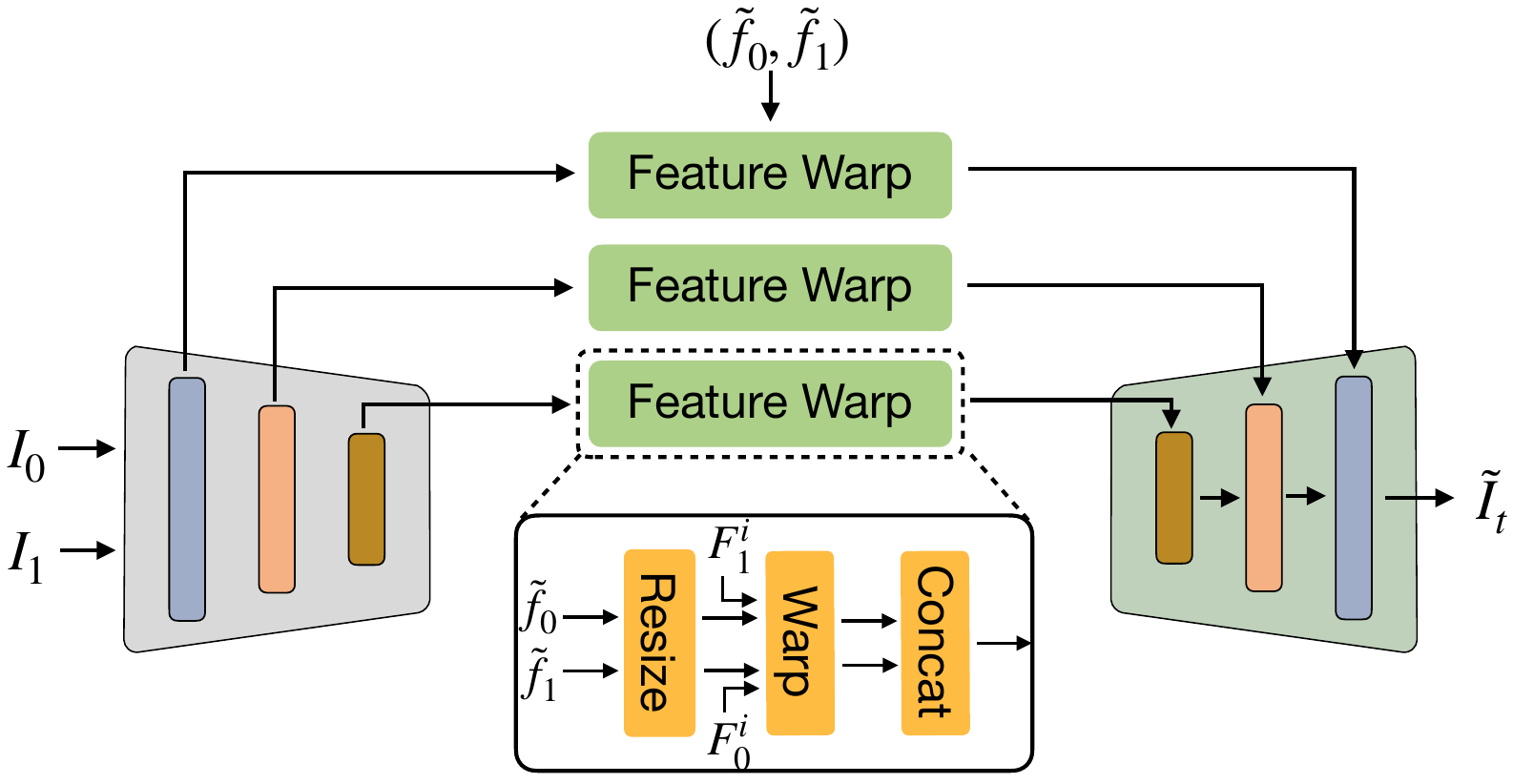}
    \vspace{-1.5mm}
    \caption{{\textbf{Illustration of flow-guided image synthesis.}}
    We train a multiscale encoder-decoder as our image synthesizer based on image pairs ($I_0$, $I_1$) and bilateral optical flow ($\tilde{f}_0$, $\tilde{f}_1$).
    }
    \label{fig:auto_encoder_arc}
\end{figure}

During training, since there is no ground truth bilateral flow ($f_0$, $f_1$) in most datasets for frame interpolation. We use a pretrained optical flow network~\cite{teed2020raft} to produce the pseudo bilateral flow ($\tilde{f}_0$, $\tilde{f}_1$).
We resize the bilateral flow to match the feature resolution on each level, and use the flow to warp encoder features accordingly. After combining the wrapped encoder features and decoder features on each level, the final synthesized result can be inferred from the output of the last decoder layer, as illustrated in Fig.~\ref{fig:auto_encoder_arc}. We use 4 channels in the last decoder layer. With one channel for a blending mask $M$, and three channels for an RGB residual map $\Delta\bI$, we have
\begin{equation}
    \tilde{\bI}_t = M \odot w(\bI_0, \tilde{f}_0) + (1 - M) \odot w(\bI_1, \tilde{f}_1) + \Delta\bI,
\end{equation}
where $w$ is the warp operation, and $\odot$ is the blending operation.
We train the encoder-decoder synthesizer by minimizing the photometric loss~\cite{reda2022film} between the ground truth target frame $\bI_t$ and the prediction $\tilde{\bI}_t$. The photometric loss is a combination of L1-based pixel-wise error $\mathcal{L}_{pixel}$, LPIPS-based perceptual reconstruction error $\mathcal{L}_{lpips}$~\cite{zhang2018lpips}, and the style loss $\mathcal{L}_{style}$~\cite{gatys2016style}:
\begin{equation}
    \mathcal{L}_{photo} = \mathcal{L}_{pixel} + \lambda_1\mathcal{L}_{lpips} + \lambda_2\mathcal{L}_{style},
    \label{eq:image_loss}
\end{equation}
where $\lambda_1$ and $\lambda_2$ are balancing parameters.

After finishing the training of the flow-guided encoder-decoder, our hierarchical flow diffusion will be conditioned on features extracted by the pretrained encoder, which will be discussed in the following section.

\subsection{Hierarchical Flow Diffusion}

\begin{figure}
    \centering
    \setlength{\tabcolsep}{1pt}
   \begin{tabular}{ccccc}

    \includegraphics[width=0.19\linewidth, trim=270 280 775 250, clip]{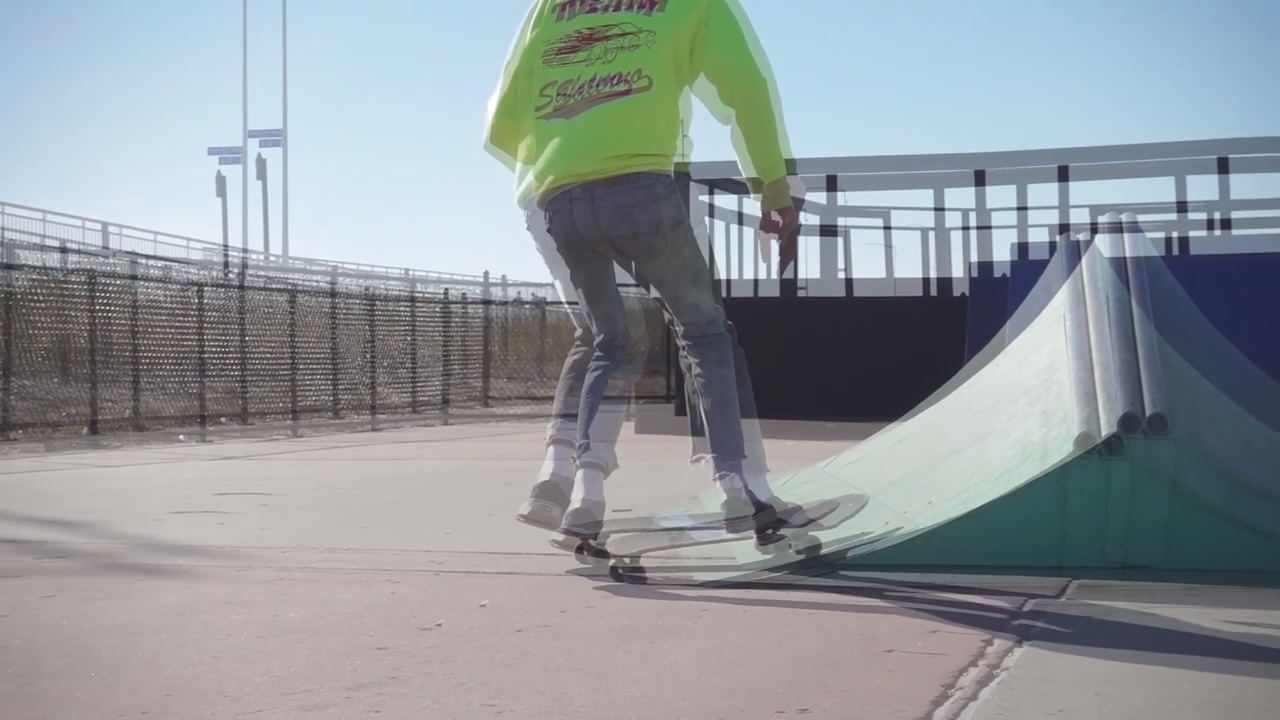} &
    \includegraphics[width=0.19\linewidth, trim=270 280 775 250, clip]{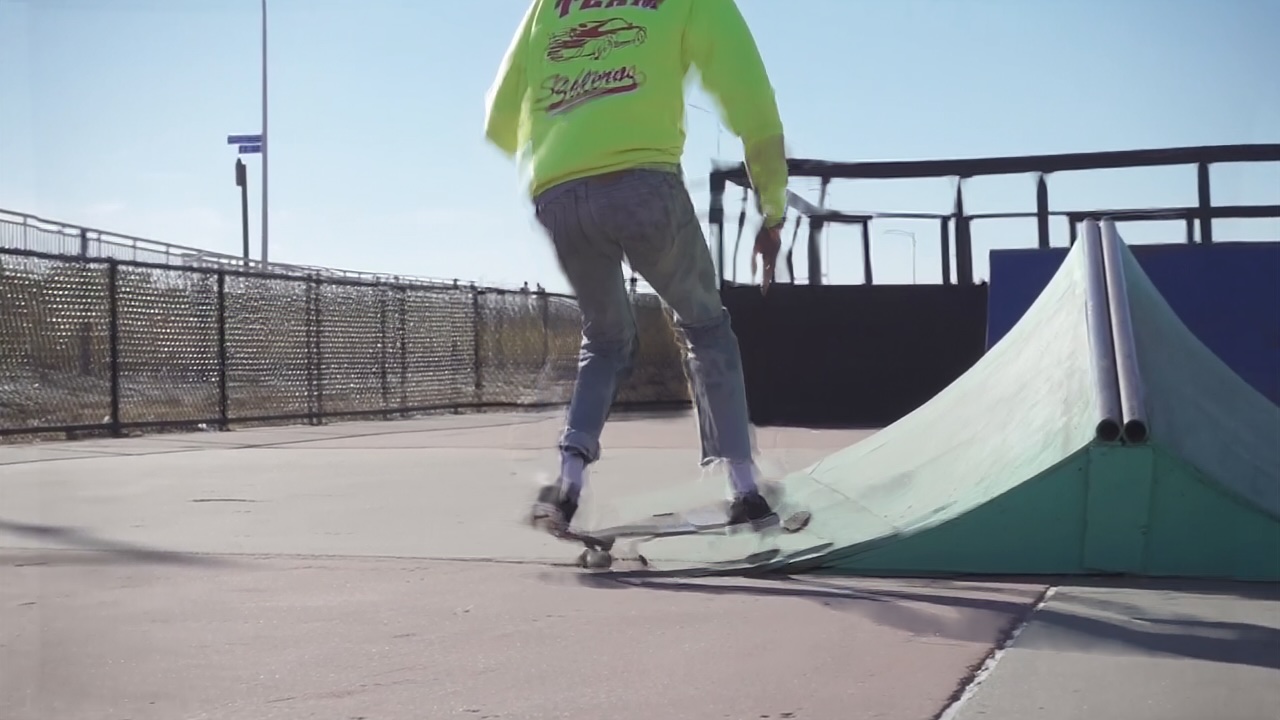} &
    \includegraphics[width=0.19\linewidth, trim=270 280 775 250, clip]{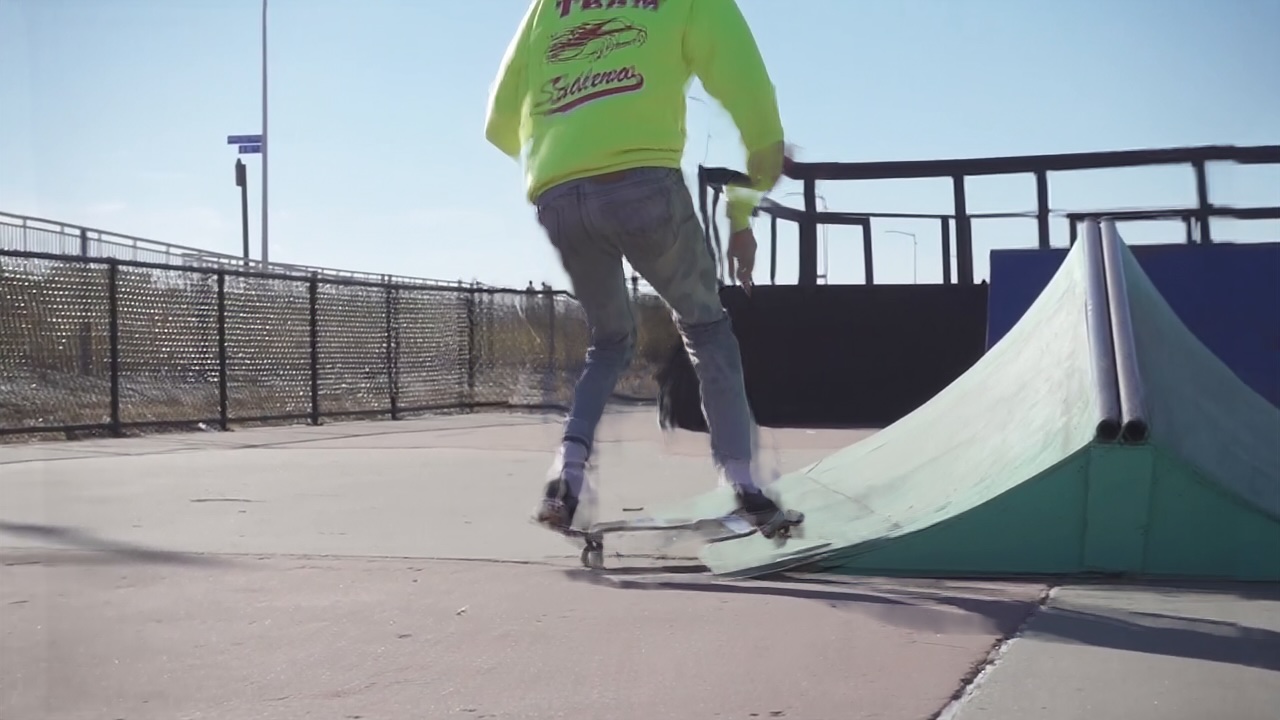}&
    \includegraphics[width=0.19\linewidth, trim=270 280 775 250, clip]{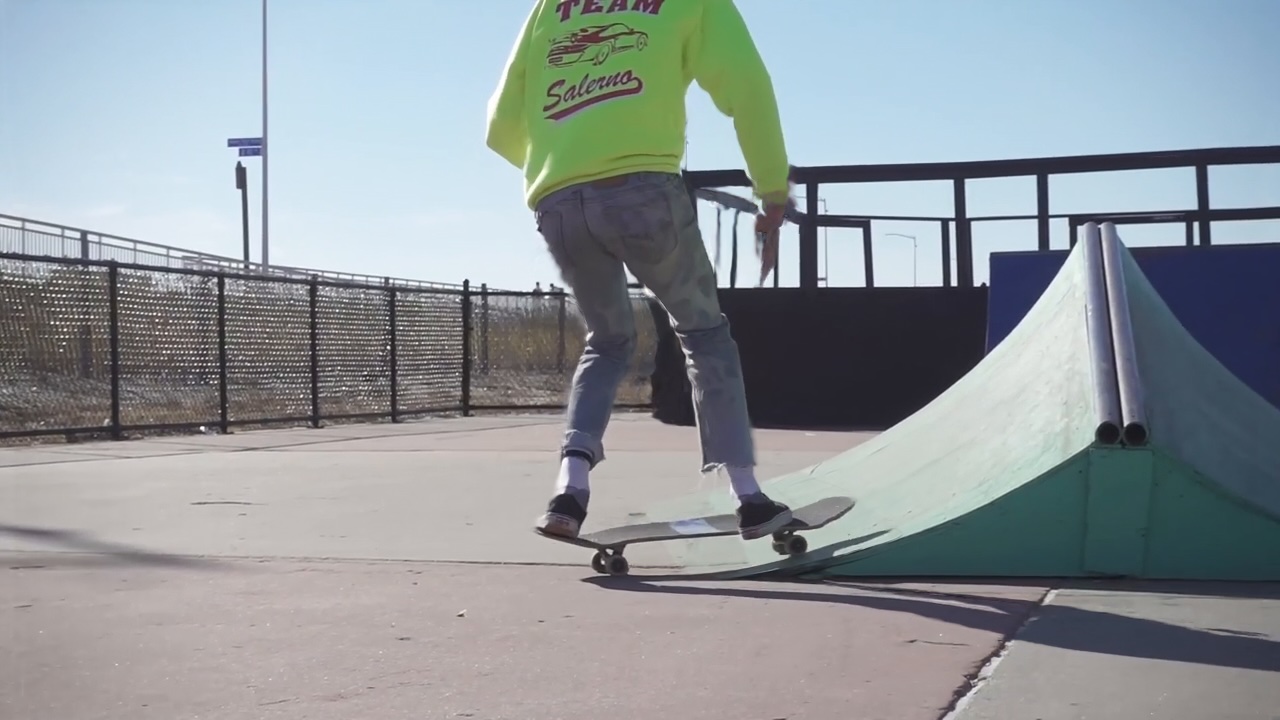} &
    \includegraphics[width=0.19\linewidth, trim=270 280 775 250, clip]{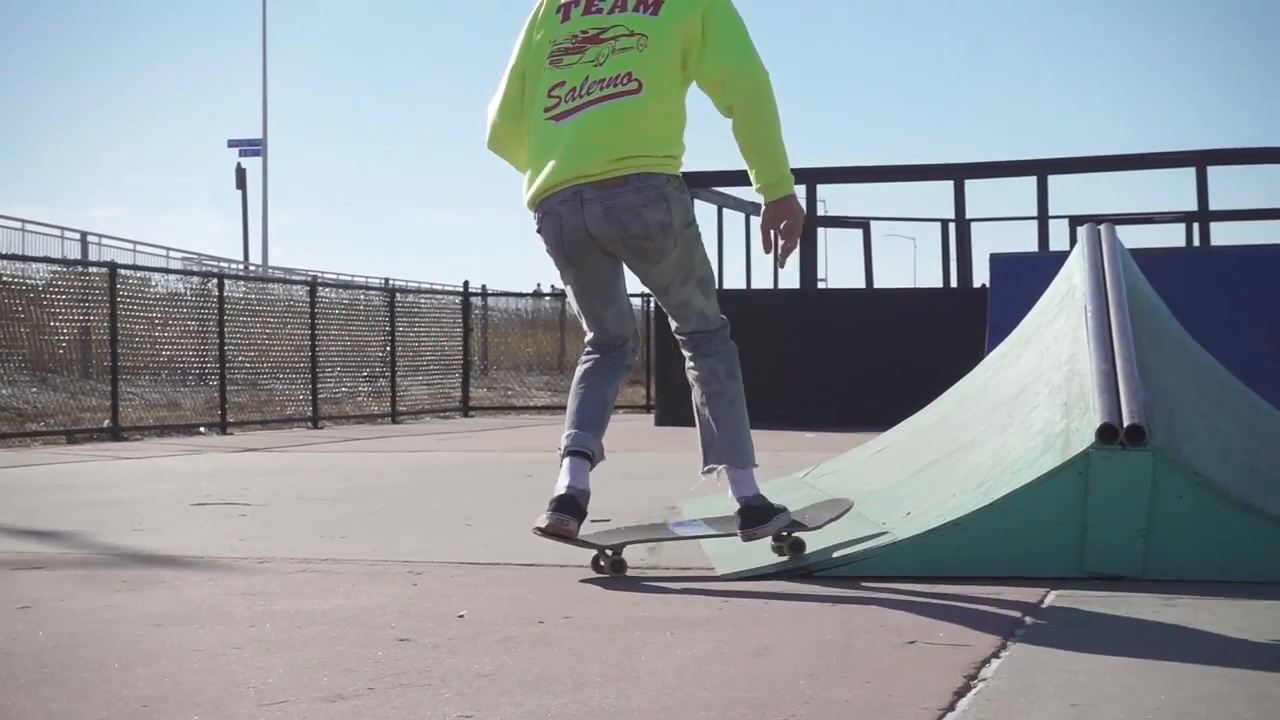} \\
    
    \includegraphics[width=0.19\linewidth, trim=270 130 775 350, clip]{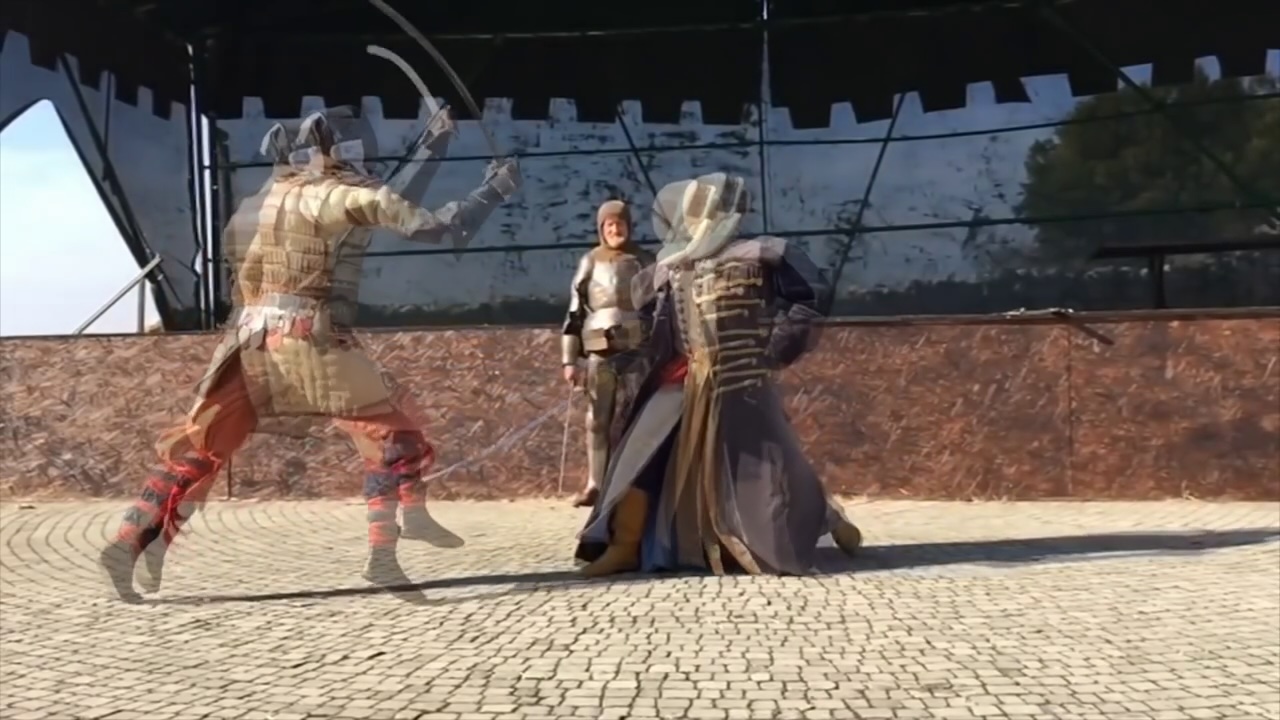} &
    \includegraphics[width=0.19\linewidth, trim=270 130 775 350, clip]{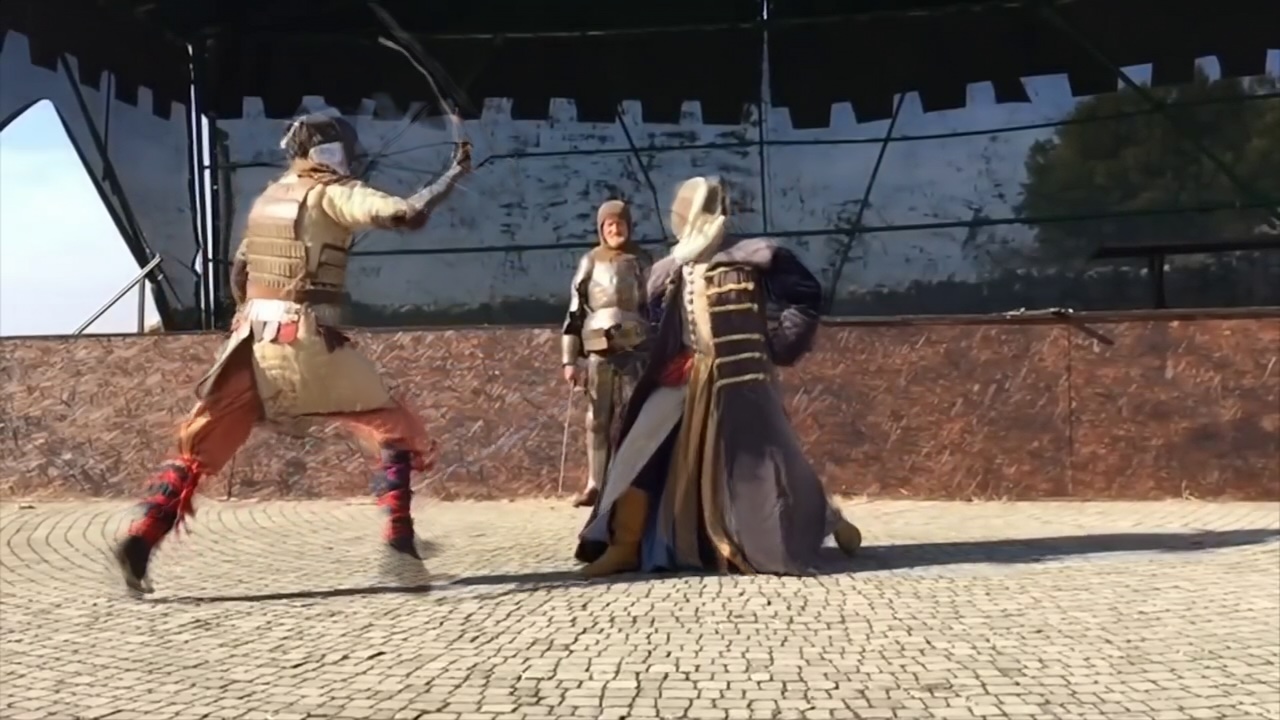} &
    \includegraphics[width=0.19\linewidth, trim=270 130 775 350, clip]{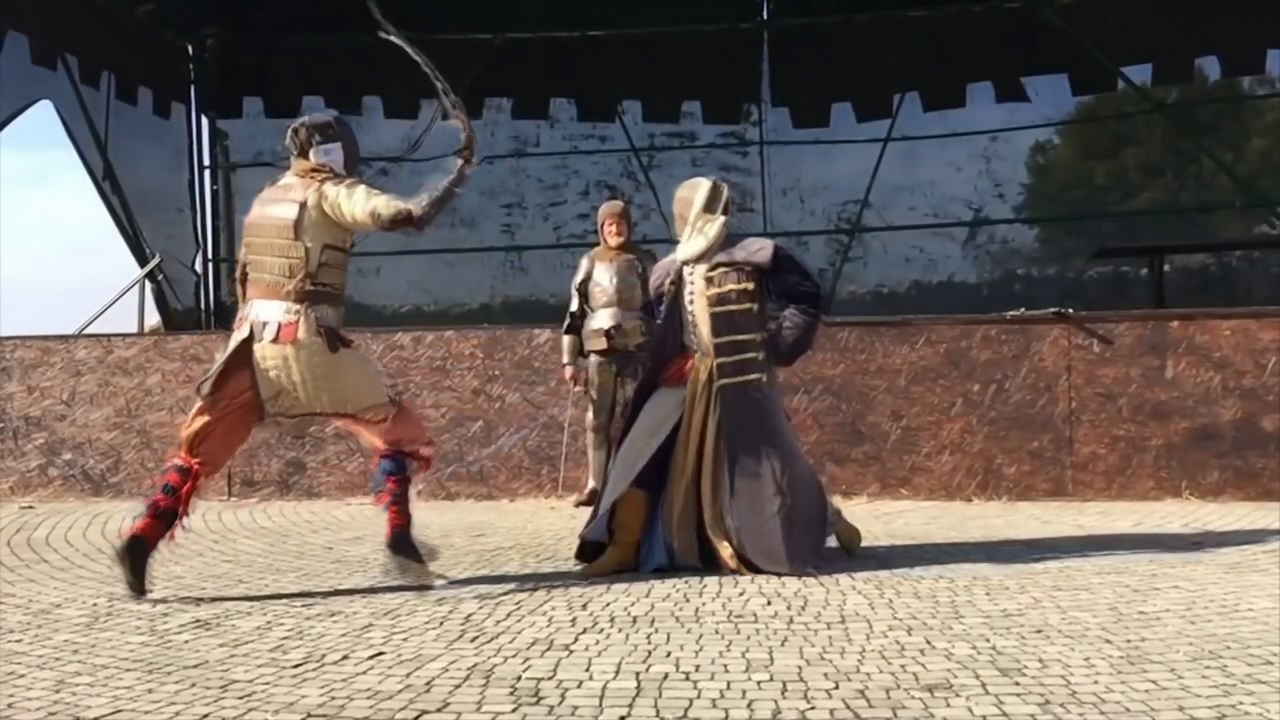}&
    \includegraphics[width=0.19\linewidth, trim=270 130 775 350, clip]{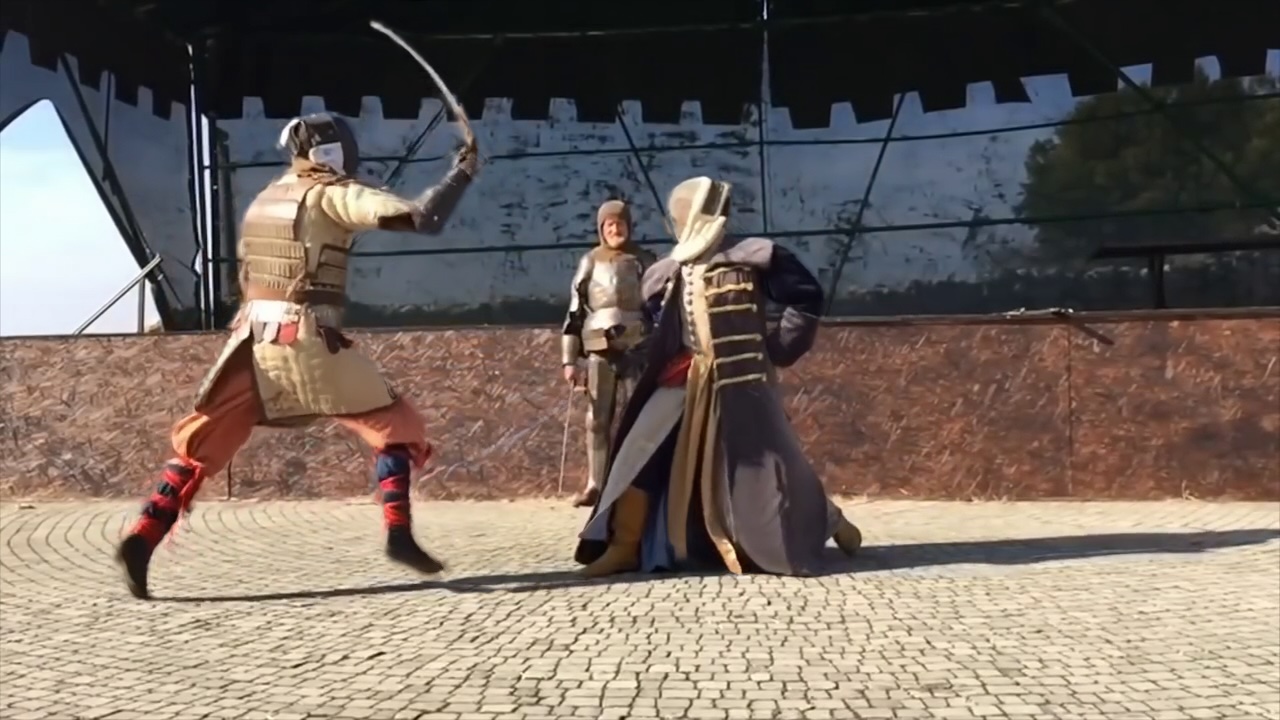} &
    \includegraphics[width=0.19\linewidth, trim=270 130 775 350, clip]{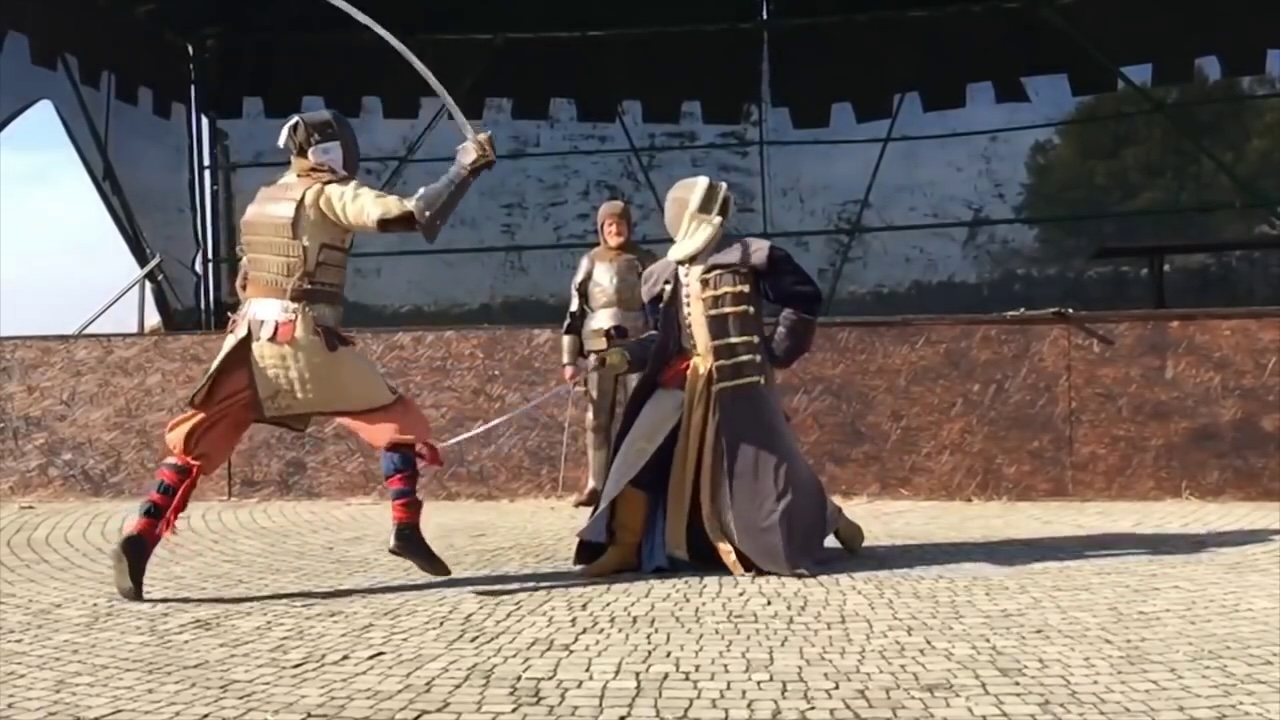} \\

    \includegraphics[width=0.19\linewidth]{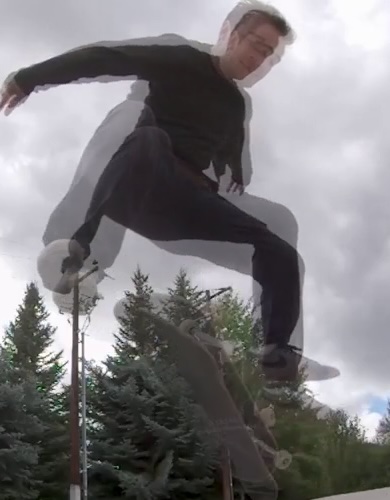} &
    \includegraphics[width=0.19\linewidth]{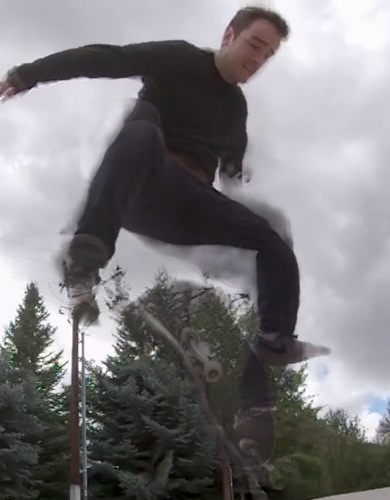} &
    \includegraphics[width=0.19\linewidth]{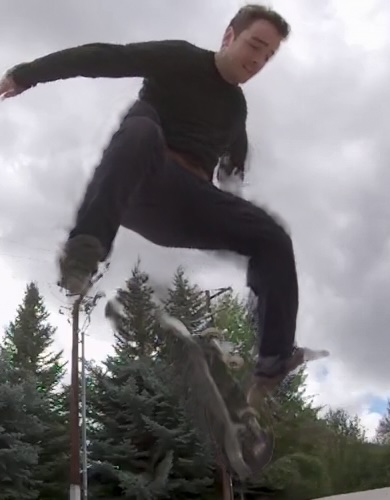} &
    \includegraphics[width=0.19\linewidth]{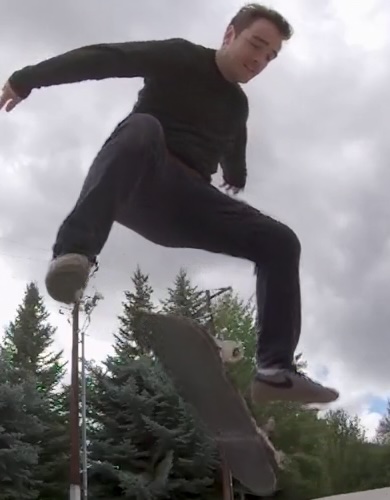} &
    \includegraphics[width=0.19\linewidth]{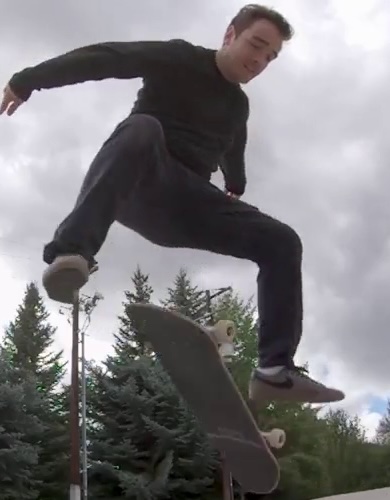} \\

    \includegraphics[width=0.19\linewidth, trim=100 160 100 325, clip]{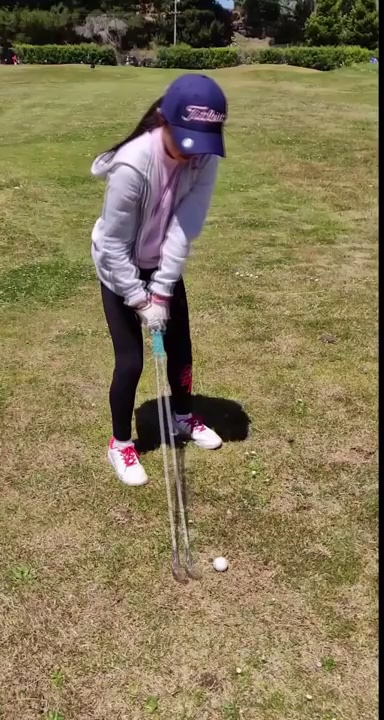} &
    \includegraphics[width=0.19\linewidth, trim=100 160 100 325, clip]{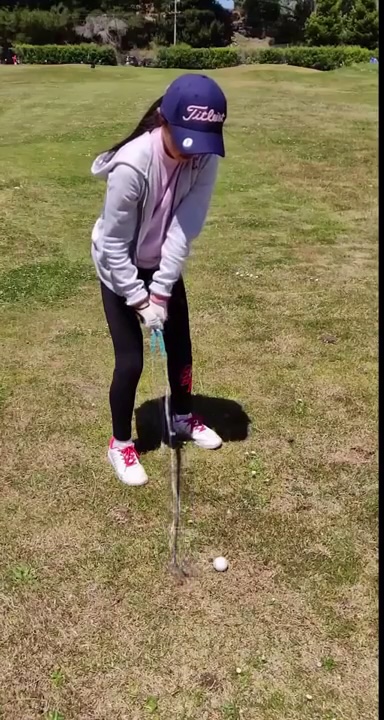} &
    \includegraphics[width=0.19\linewidth, trim=100 160 100 325, clip]{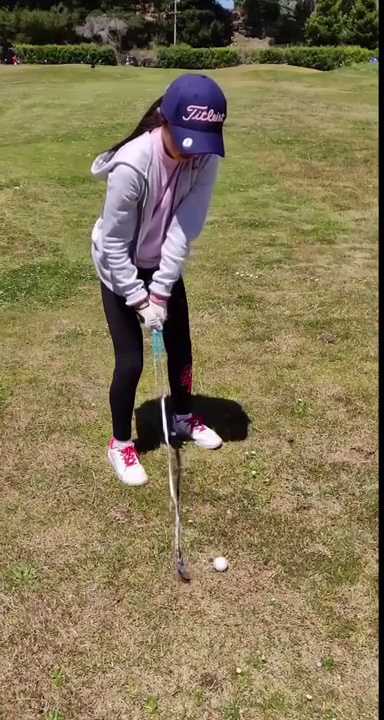} &
    \includegraphics[width=0.19\linewidth, trim=100 160 100 325, clip]{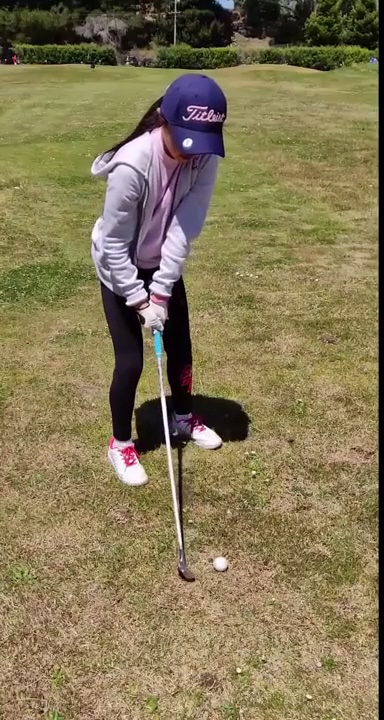} &
    \includegraphics[width=0.19\linewidth, trim=100 160 100 325, clip]{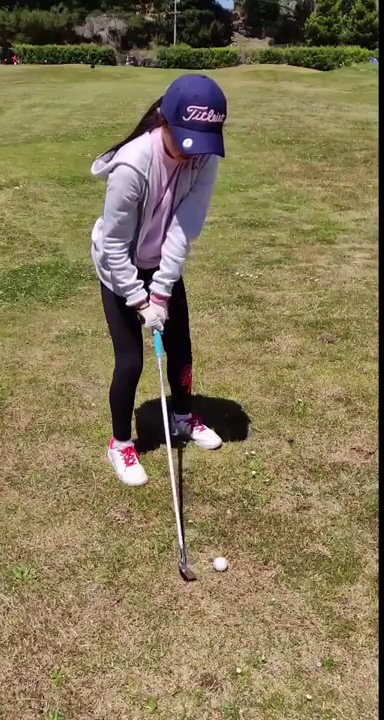} \\

    \small{Input} & \small{Scale 1/16}  & \small{Scale 1/8}  & \small{Scale 1/4} & \small{GT} \\
    
\end{tabular}
   \vspace{-1.5mm}
    \caption{{\bf Results of the hierarchical models on different scales.}
    We show the coarse-to-fine results from left to right in addition to the input and ground truth. With the proposed hierarchical diffusion models, the result becomes progressively better with finer resolution, making it capable of handling complex motions and large displacements.
    }
    \vspace{-3mm}
    \label{fig:vis_ms_compare}
\end{figure}

We use diffusion to denoise optical flow starting from a Gaussian noise. Unlike most existing diffusion models that are based on only a single fixed resolution, we propose to denoise the optical flow in multiple stages from coarse to fine, as shown in Fig.~\ref{fig:vis_ms_compare}.

With the input image pair $(\bI_0, \bI_1)$, we exploit the encoder mentioned above to extract the multiscale feature pair
\begin{equation}
    \{(\bF_0^i, \bF_1^i)\}, \ \ \ k_0 \leq i \leq k_1, 
\end{equation}
where the feature level $i$ has a resolution $1/2^{i}$ of the original image, and $k_0$ and $k_1$ denote the finest and coarsest feature level we used in hierarchical diffusion. 

At feature level $i$, our denoising U-Net is conditioned on the feature pair $(\bF_0^i, \bF_1^i)$,  and takes the noisy bilateral flow $(f_0^t, f_1^t)$ as input to predict the target flow $(\tilde{f}_0^i, \tilde{f}_1^i)$, both in the  $1/2^i$ of the original resolution.
By denoting the denoising U-Net as $u$ with parameter $\theta$, we can write as
\begin{equation}
    \tilde{f}_0^i, \tilde{f}_1^i = u(\bF_0^i, \bF_1^i, f_0^t, f_1^t, t; \theta).
\end{equation}
Before forwarding the denoising U-Net, we normalize the feature, and use a flow projector and feature projector to process the noisy flow and feature pair, respectively. 
We share the parameter of the diffusion model across different feature levels, except for the feature and flow projectors.

In the diffusion procedure, we uniformly divide the entire denoising process $t\sim \mathcal{U}(0, T)$ into multiple stages, while each stage corresponds to a feature level $i$. We express this as
\begin{equation}
    s_t = \{i|t_0^i \leq t  <  t_1^i\}, \ \  \   t_0^{k_0} = 0, \ \  t_1^{k_1} = T, 
\end{equation}
where $t_0^i$ and  $t_1^i$ is the starting and ending point, and $(\bF_0^i, \bF_1^i)$ is used as the condition of the denoising U-Net for interval $i$.

When time step $t-1$ and $t$ do not belong to the same denoising stage, i.e., $s_t > s_{t-1}$, we perform 2$\times$ bilinear upsample for the estimated $(\tilde{f}_0^{s_t}, \tilde{f}_1^{s_t})$ , and apply the forward function in DDPM~\cite{ho2020ddpm} to approximate the input for timestep $t-1$~\cite{zhou2023pyramid}.
We summarize this as
\begin{equation}
    f^{t-1} = \sqrt{\alpha_{t-1}} \uparrow \mkern-6mu \tilde{f}^{s_t} + \sqrt{1 - \alpha_{t-1}} \epsilon, 
    \label{eq:forward}
\end{equation}
where $\uparrow \mkern-6mu \tilde{f}^{s_t}$ is 2$\times$ upsampled flow,  $\alpha_t \in \{\alpha_1, ..., \alpha_T\}$ is a predefined noise schedule, and $\epsilon \sim \mathcal{N}(0, 1)$ is the gaussian noise.

When the previous time step $t-1$ and current time step $t$ falls into the same interval, we update $(f_0^{t-1}, f_1^{t-1})$ using the reverse function in DDPM~\cite{ho2020ddpm}
\begin{equation}
    f^{t-1} = \sqrt{\alpha_{t-1}} \tilde{f}^{s_t}+ \sqrt{1 - \alpha_{t-1} - \sigma^2_t} \tilde{\epsilon}_t + \sigma_t \epsilon,
    \label{eq:denoising}
\end{equation}
where $\tilde{\epsilon}_t$ is the estimated noise, which is
\begin{equation}
    \tilde{\epsilon}_t = (f^t - \sqrt{\alpha_t}\tilde{f}^{s_t})/{\sqrt{1 - \alpha_t}}.
\end{equation}
We apply Eq.~\ref{eq:forward} and Eq.~\ref{eq:denoising} for $\tilde{f}_0^{s_t}$ and $\tilde{f}_1^{s_t}$ to yield $f_0^{t-1}$ and $f_1^{t-1}$, respectively.

We train hierarchical diffusion models simultaneously at all feature levels.
At each level $i$, we randomly sample a time step $t_i$, construct the noisy flow input $(f_0^{t_i}, f_1^{t_i})$ with the resized ground truth flow $(f_0^i, f_1^i)$ matching the resolution of the level $i$, and apply L1 loss to supervise their prediction $(\tilde{f}_0^i, \tilde{f}_1^i)$
\begin{equation}
    \mathcal{L}_{flow} = \sum_{i=k_0}^{k_1}||\tilde{f}_0^i- f_0^i||_1 + ||\tilde{f}_1^i - f_1^i||_1. 
\end{equation}

\begin{table*}
    \centering
        
\begin{tabular}{l|cc|cc|cc|cc}
    \toprule
     \multirow{2}{*}{Method} & \multicolumn{2}{c|}{easy} & \multicolumn{2}{c|}{medium} & \multicolumn{2}{c|}{hard}  & \multicolumn{2}{c}{extreme}\\
     ~                              & LPIPS    & FID           & LPIPS      & FID       & LPIPS     & FID           & LPIPS     & FID       \\
     \midrule
     AMT-G~\cite{li2023amt}          & 0.0325    & 6.139         & 0.0447     & 11.039    & 0.0680     & 20.810        & 0.1128     & 40.075    \\
     SGM-VFI~\cite{liu2024sgm-vfi}   & 0.0191    & 5.854         & 0.0329     & 10.945    & 0.0611    & 22.004        & 0.1182     & 41.078   \\
     EMA-VFI~\cite{zhang2023ema_vfi} & 0.0186    & 5.882         & 0.0325     & 11.051    & 0.0579     & 20.679        & 0.1099     & 39.051    \\
     URPNet-LARGE~\cite{jin2023urpnet}& 0.0179   & 5.669         & 0.0389     & 10.983     & 0.0604     & 22.127        & 0.1115     & 40.098    \\
     Per-VFI~\cite{wu2024pervfi}     & 0.0166   & 6.654          & 0.0263      & 11.509        & 0.0480    & 19.855 & 0.0901   & 34.182 \\
     LDMVFI~\cite{danier2024ldmvfi}  & 0.0145   & 5.752         & 0.0284     & 12.485    & 0.0602     & 26.520        & 0.1226     & 47.042    \\
     MADIFF~\cite{huang2024madiff}   & 0.0130    & 5.334         & 0.0270     & 11.022    & 0.0580     &22.707         & 0.1180     & 44.923   \\
     CBBD~\cite{lyu2024cbbd}         & \underline{0.0112} & \underline{4.791}   & \underline{0.0274}     & \underline{9.039}     & \underline{0.0467}     & \underline{18.589}        & \underline{0.1040}     & \underline{36.729}    \\
     \textbf{Ours}                   &  \textbf{0.0098}      & \textbf{4.541}           & \textbf{0.0191}          & \textbf{8.499}       & \textbf{0.0405}          & \textbf{15.320} &\textbf{0.0839} &\textbf{27.032}\\
     \bottomrule
\end{tabular}
    \vspace{-2.5mm}
    \caption{\textbf{Comparison on SNU-FILM ~\cite{choi2020snufilm} benchmark.}
    Our method outperforms the current SOTA methods significantly, especially in the hard and extreme subset of SNU-FILM. 
    }
    \vspace{-3mm}
    \label{tab:compre_snufilm}
\end{table*}

\begin{table}
    \centering
    
        \setlength{\tabcolsep}{4pt}
        \begin{tabular}{l|cc|cc}
    \toprule
     \multirow{2}{*}{Method} & \multicolumn{2}{c|}{2K} & \multicolumn{2}{c}{4K}\\
     ~                              &LPIPS                  & FID           & LPIPS     & FID\\
     \midrule
     URPNet-LARGE~\cite{jin2023urpnet}& 0.1010              & 14.209        & 0.2150    & 32.003   \\
     EMA-VFI~\cite{zhang2023ema_vfi} & 0.1024               & 12.332        & 0.2258    &  30.675  \\
     AMT-G~\cite{li2023amt}          & 0.1061               & 13.089        & 0.2054    & 29.512  \\
     SGM-VFI~\cite{liu2024sgm-vfi}   & 0.1000               & 12.375        & 0.2172     & 27.334    \\
     LDMVFI~\cite{danier2024ldmvfi}  & 0.0420               & 11.385        & 0.0859    & 21.272 \\
     CBBD~\cite{lyu2024cbbd}         & \underline{0.0272} & \underline{10.168}        & \underline{0.0634}     & \underline{24.621}\\
     \textbf{Ours}                   & \textbf{0.0264}    & \textbf{7.940} &\textbf{0.0614} &\textbf{14.132}       \\
     \bottomrule
\end{tabular}
    \vspace{-2.5mm}
    \caption{{\bf Comparison on Xiph~\cite{montgomery1994xiph} benchmark.}
    Our method achieves the best performance across all evaluation settings, especially in the more challenging 4K setting.
    }
    \vspace{-3mm}
    \label{tab:compare_xiph}
\end{table}

\begin{table}
    \centering

\begin{tabular}{l|cc|cc}
    \toprule
     \multirow{2}{*}{Method}    & \multicolumn{2}{c|}{DAVIS}       & \multicolumn{2}{c}{Vimeo-90k} \\
     ~                          & LPIPS     & FID                   & LPIPS     & FID           \\
     \midrule
     VFIformer~\cite{lu2022vfiformer}   & 0.1272     & 14.407            & 0.0212        &  3.341          \\
     EMA-VFI~\cite{zhang2023ema_vfi}    & 0.1324     & 15.186              & 0.0213      & 3.819               \\
     AMT-G~\cite{li2023amt}             & 0.1091     & 13.018               & 0.0208     &   3.172            \\
     LDMVFI~\cite{danier2024ldmvfi}     & 0.1070    & 12.554            & 0.0234              &  2.744         \\
     MADIFF~\cite{huang2024madiff}      & 0.0960     & 11.089            &  -    &    -     \\
     Per-VFI~\cite{wu2024pervfi}        & \underline{0.0819}     & 8.813                  & 0.0180          &   2.314        \\
     CBBD~\cite{lyu2024cbbd}           & 0.0919     & \underline{9.220}            & \underline{0.0123}           &\underline{1.961}       \\
     
     \textbf{Ours}              &  \textbf{0.0753}        & \textbf{7.237}           & \textbf{0.0120}          &\textbf{1.712}               \\
     \bottomrule
     
\end{tabular}
    \vspace{-2.5mm}
    \caption{\textbf{Comparison on DAVIS~\cite{perazzi2016davis} and Vimeo-90k~\cite{xue2019vimeo}.}
    Our method outperforms all other competitors, consistently.}
    \vspace{-3mm}
    \label{tab:compare_others_benchmark}
\end{table}

\subsection{End-to-end Training with Joint Fine-tuning}
\label{sec:joint_optimize}

After finishing the training of the encoder-decoder synthesizer and the hierarchical diffusion models separately, we propose to fine-tune these two components jointly, producing an end-to-end interpolation framework.
Given multiscale bilateral flow $(\tilde{f}_0^i, \tilde{f}_1^i)$ predicted by the hierarchical flow diffusion models, we use them to warp the corresponding features $(\bF_0^i, \bF_1^i)$ from the encoder to construct the flow-guided features for the decoder.
We use the photometric loss in Eq.~\ref{eq:image_loss} to supervise the target image generated by the decoder.
We will show in experiments that this end-to-end training strategy improves the performance.

\begin{figure*}
    \centering
     \setlength{\tabcolsep}{2pt}
    \resizebox{0.99\linewidth}{!}{
        \begin{tabular}{ccccccc}

    \includegraphics[width=0.20\linewidth, trim=200 400 210 0, clip]{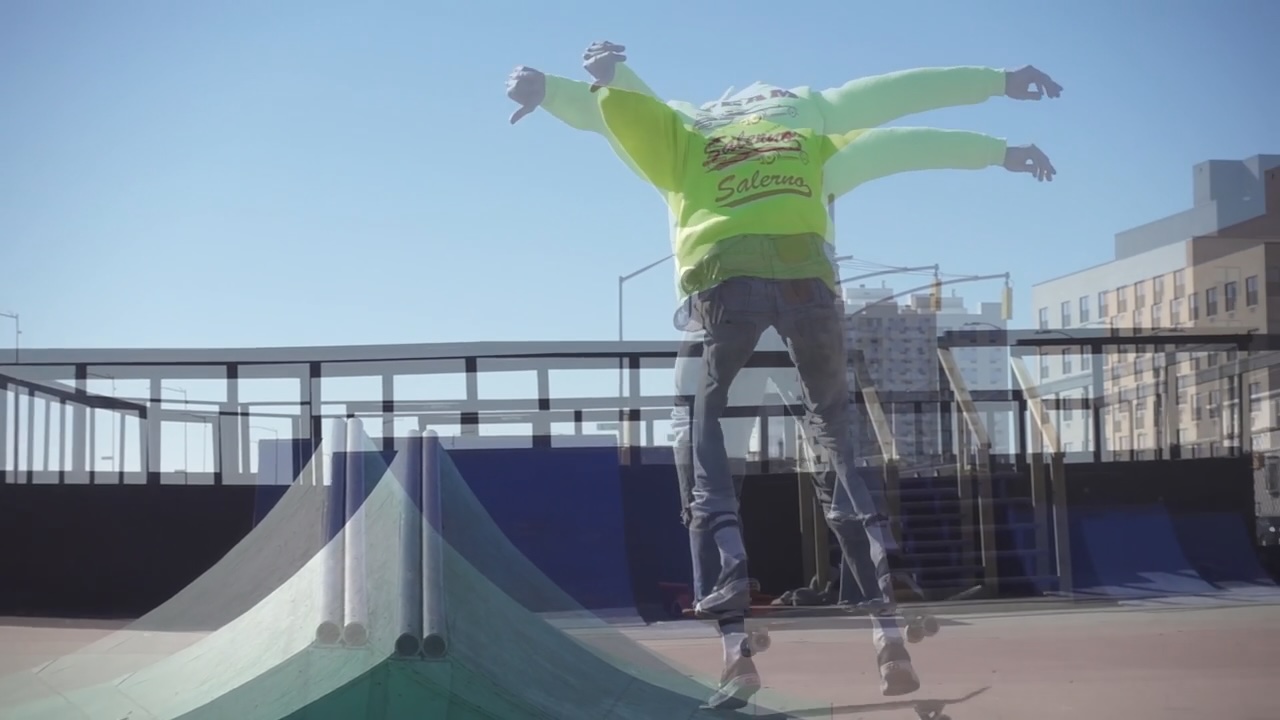} &
    \includegraphics[width=0.13\linewidth, trim=500 400 210 0, clip]{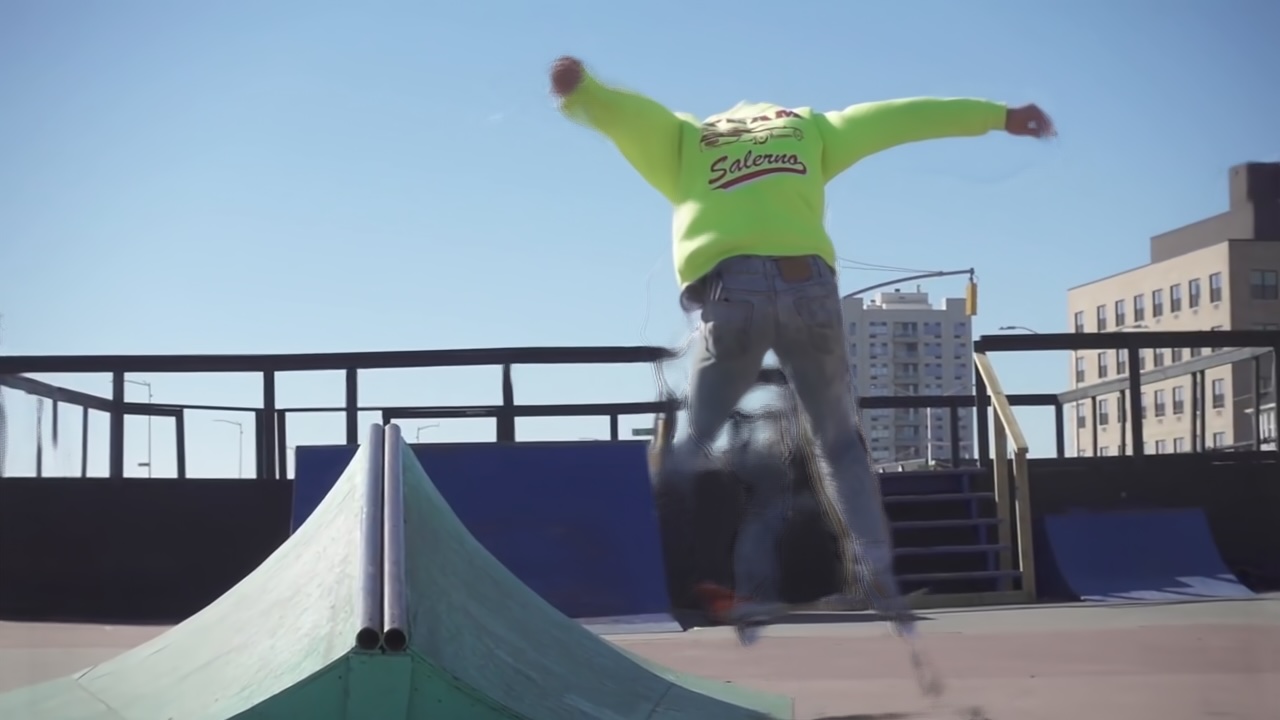}&
    \includegraphics[width=0.13\linewidth, trim=500 400 210 0, clip]{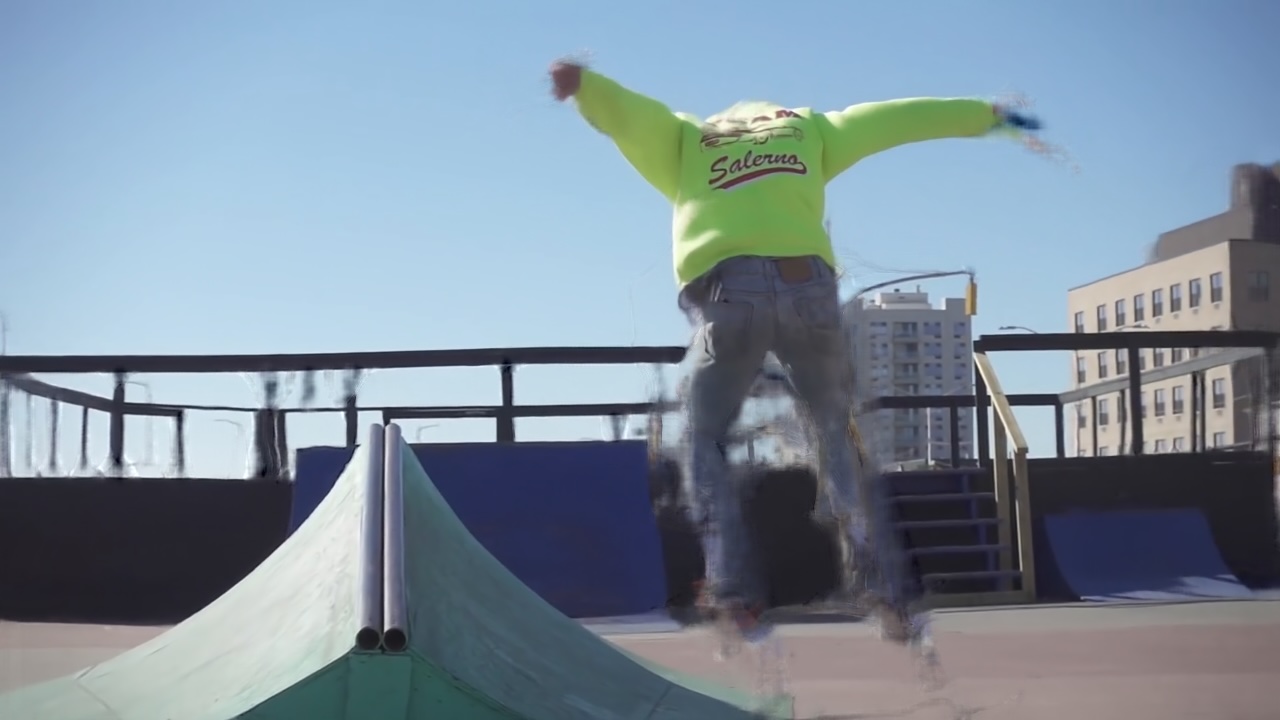}&
    \includegraphics[width=0.13\linewidth, trim=500 400 210 0, clip]{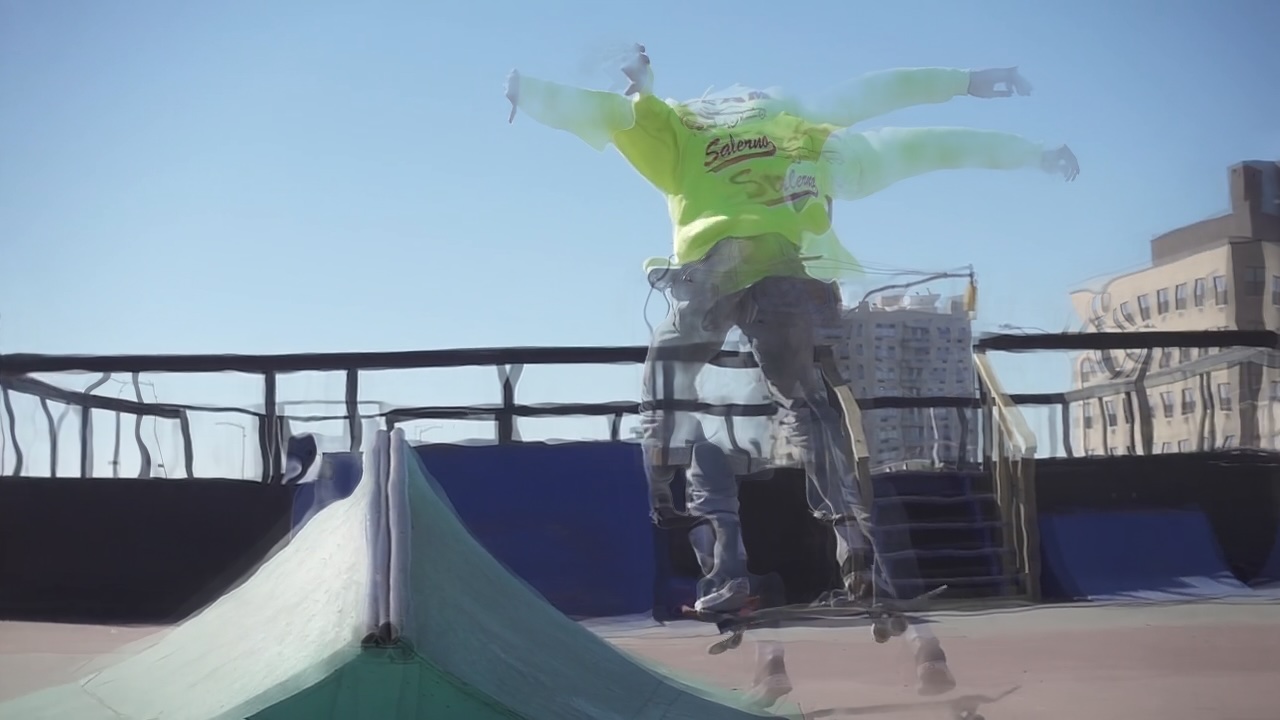}&
    \includegraphics[width=0.13\linewidth, trim=500 400 210 0, clip]{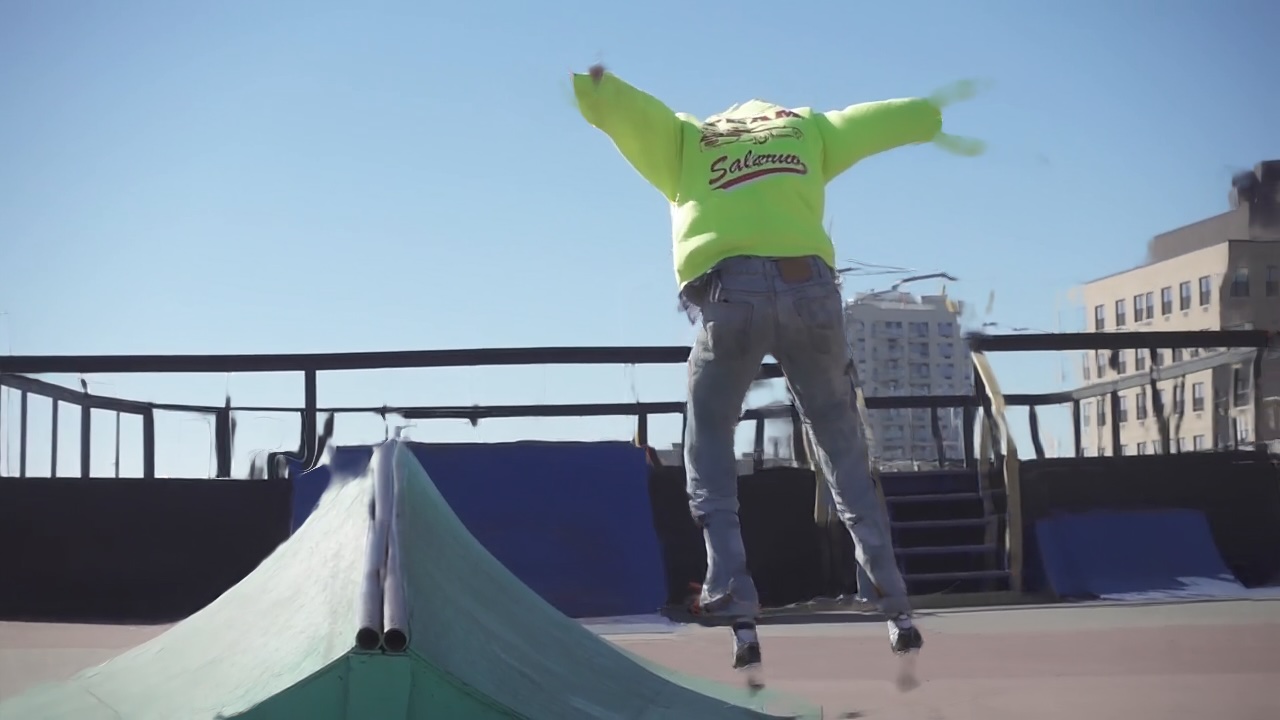}&
    \includegraphics[width=0.13\linewidth, trim=500 400 210 0, clip]{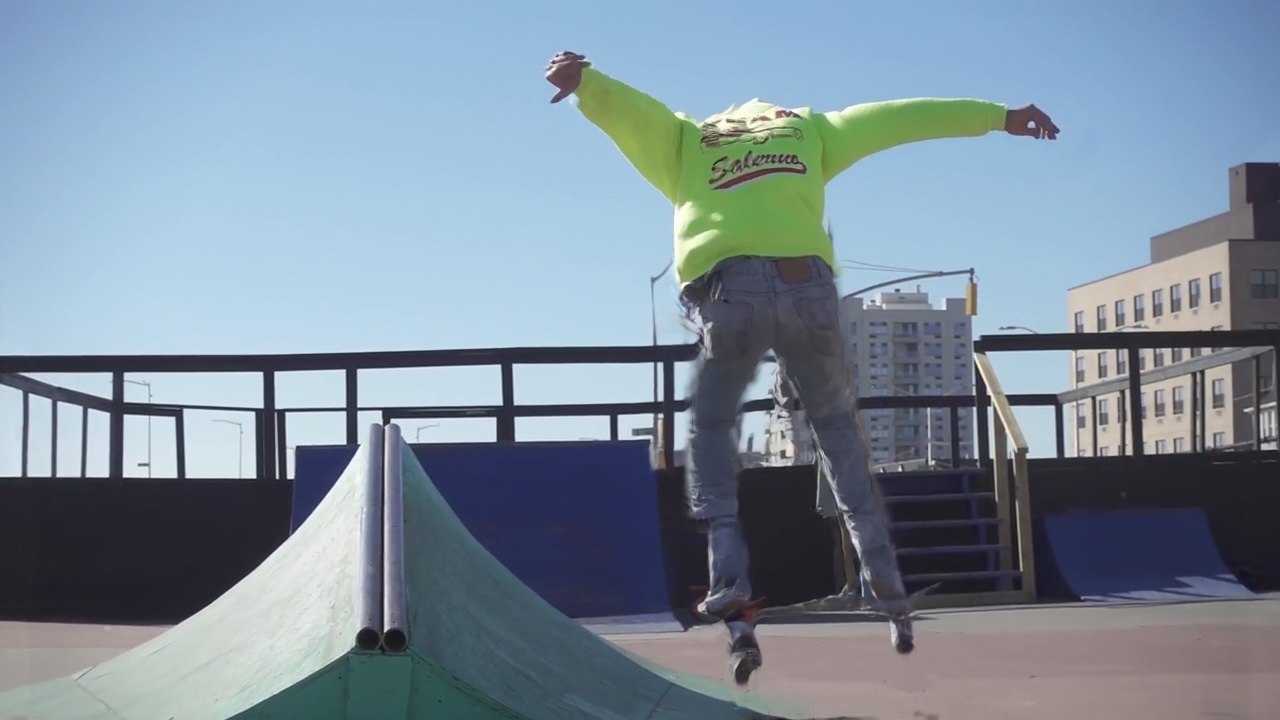}&
    \includegraphics[width=0.13\linewidth, trim=500 400 210 0, clip]{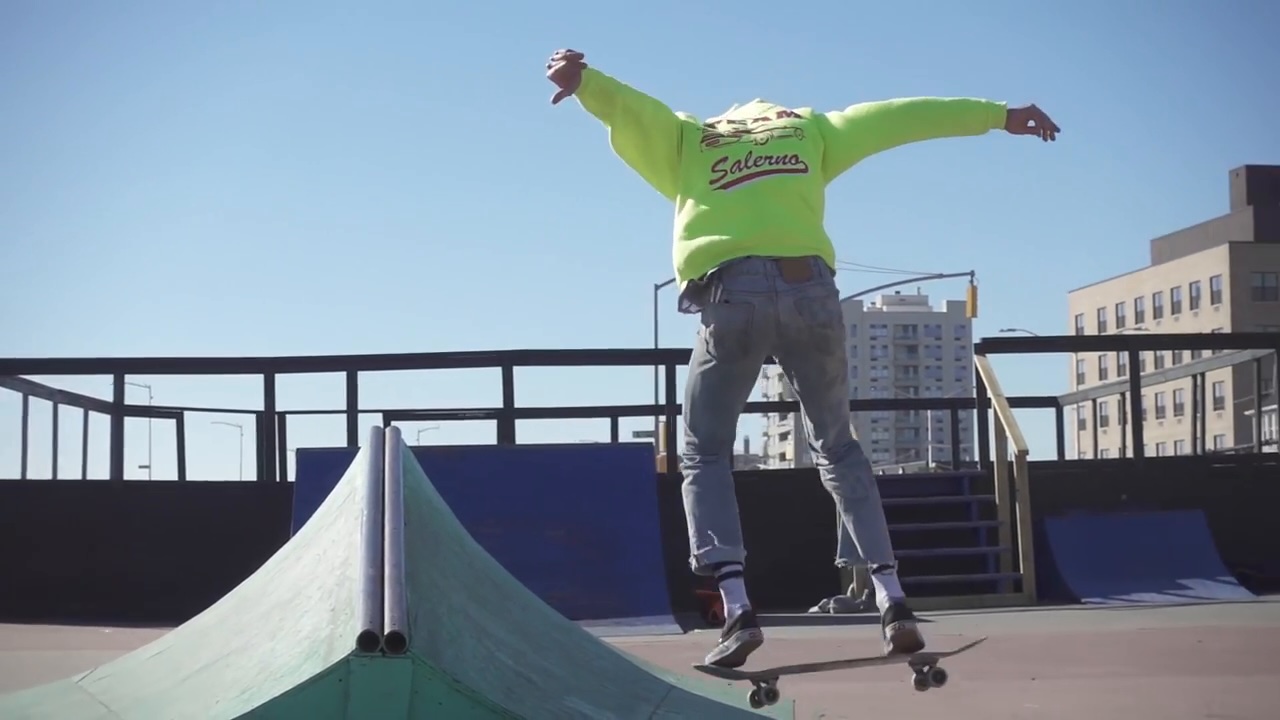}\\

\includegraphics[width=0.2\linewidth, trim=0 150 0 0, clip]{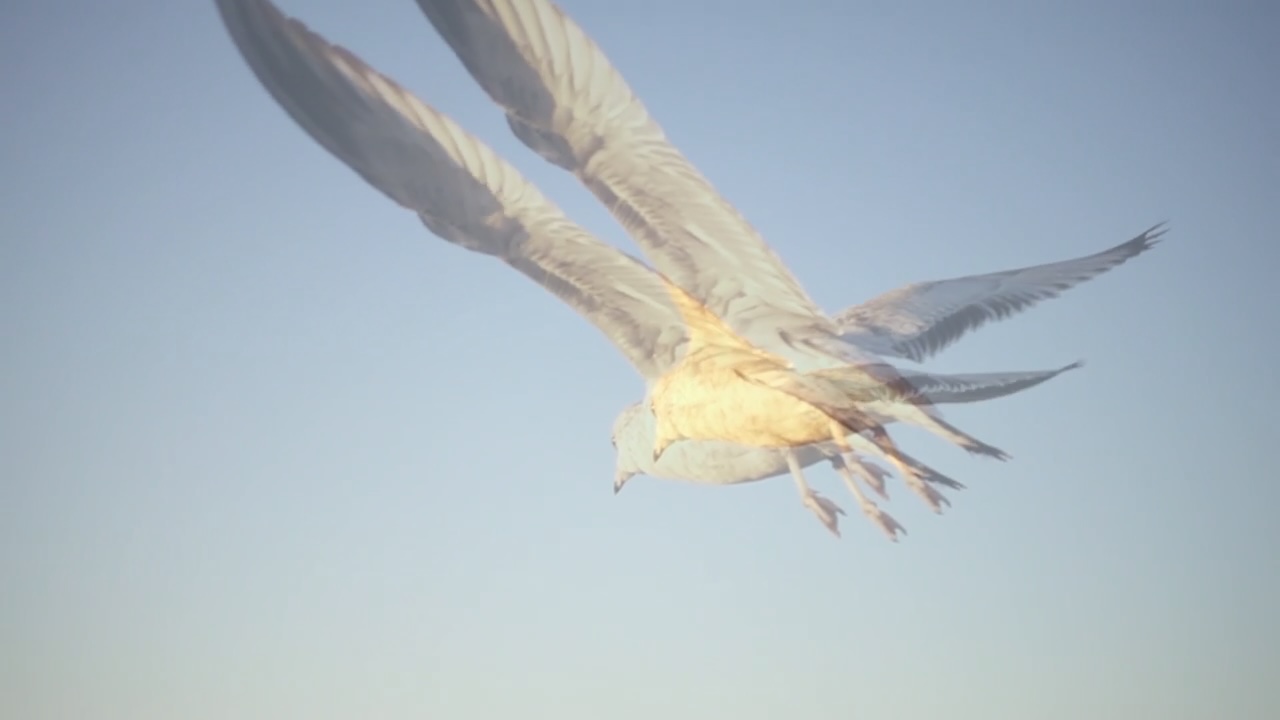} &
\includegraphics[width=0.13\linewidth,trim=380 170 100 0, clip]{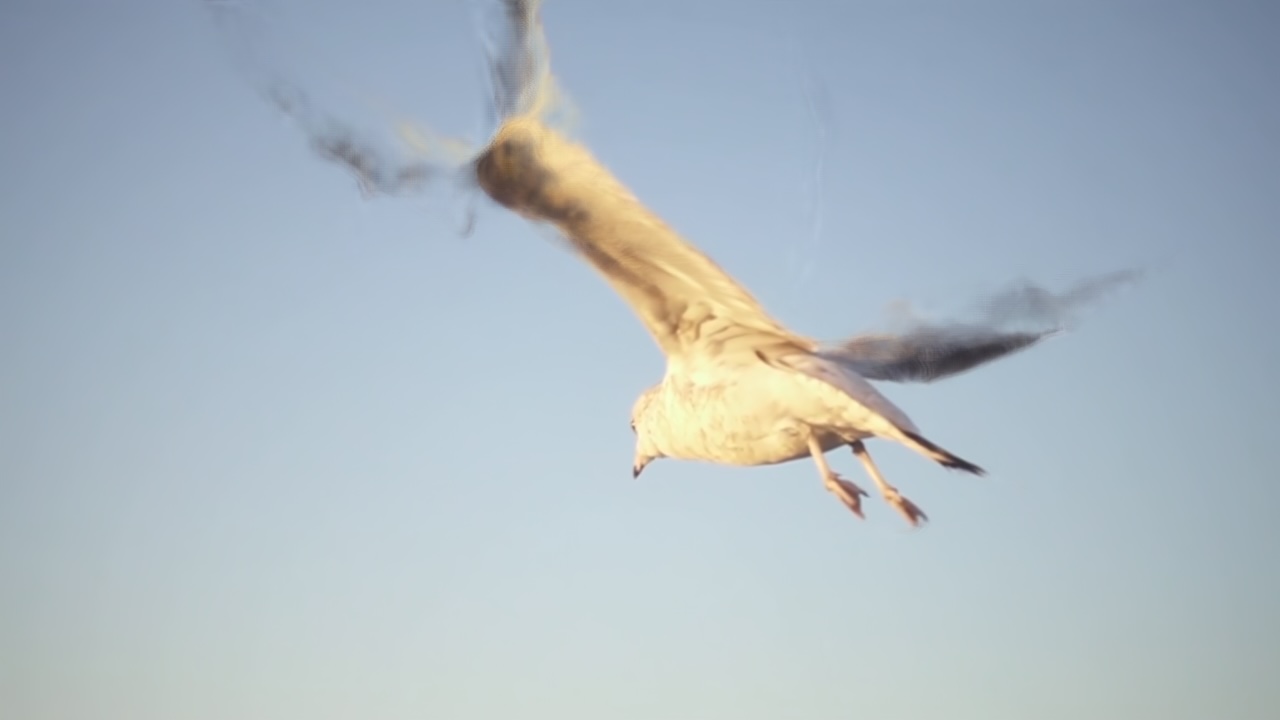}&
\includegraphics[width=0.13\linewidth,trim=380 170 100 0, clip]{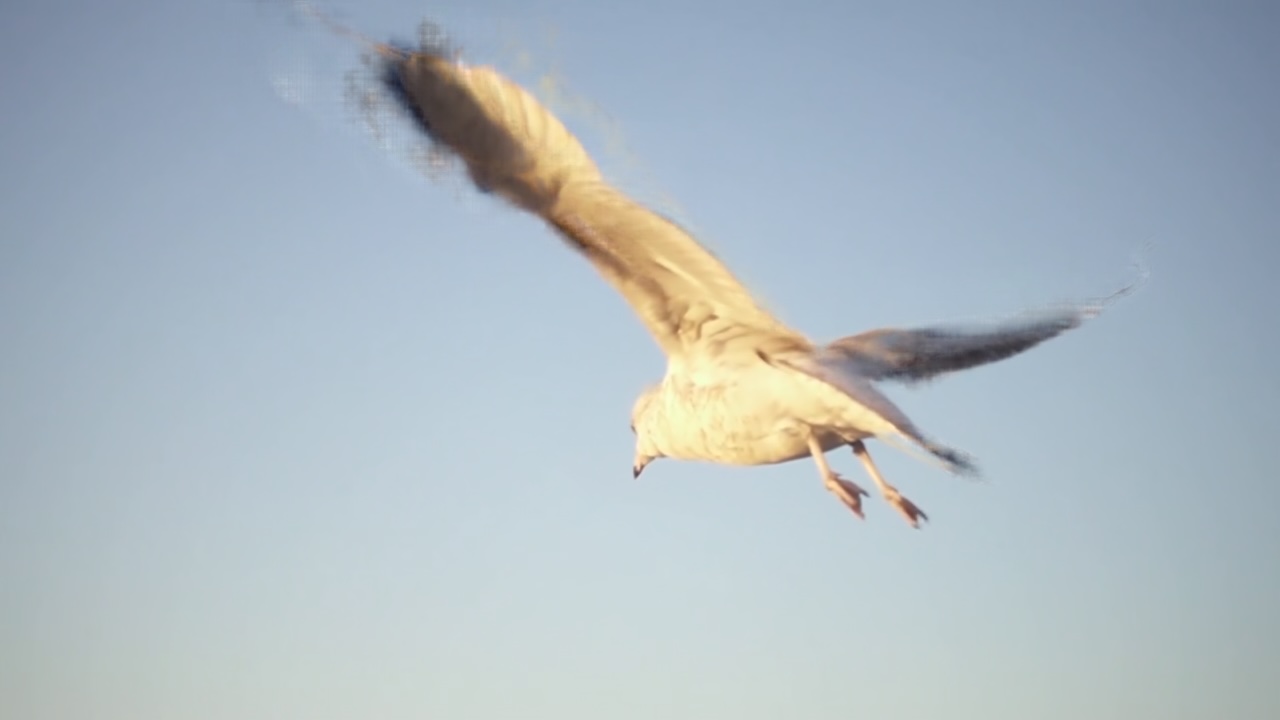}&
\includegraphics[width=0.13\linewidth,trim=380 170 100 0, clip]{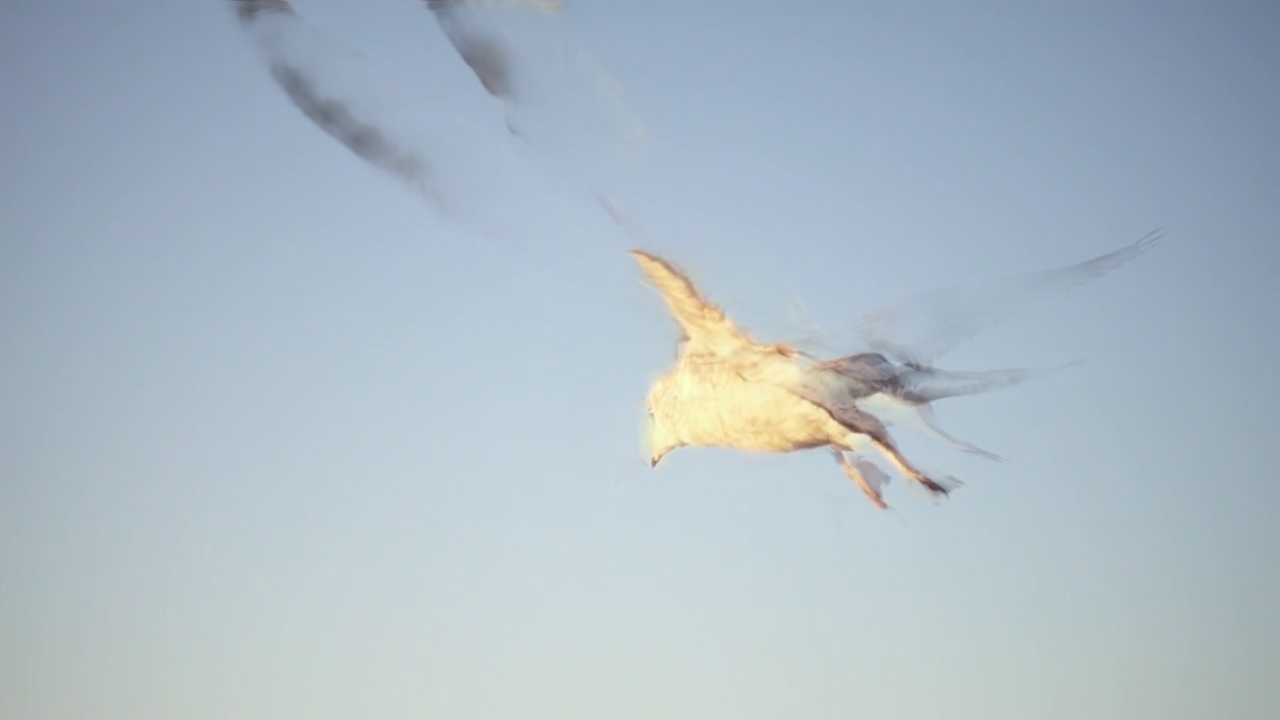}&
\includegraphics[width=0.13\linewidth,trim=380 170 100 0, clip]{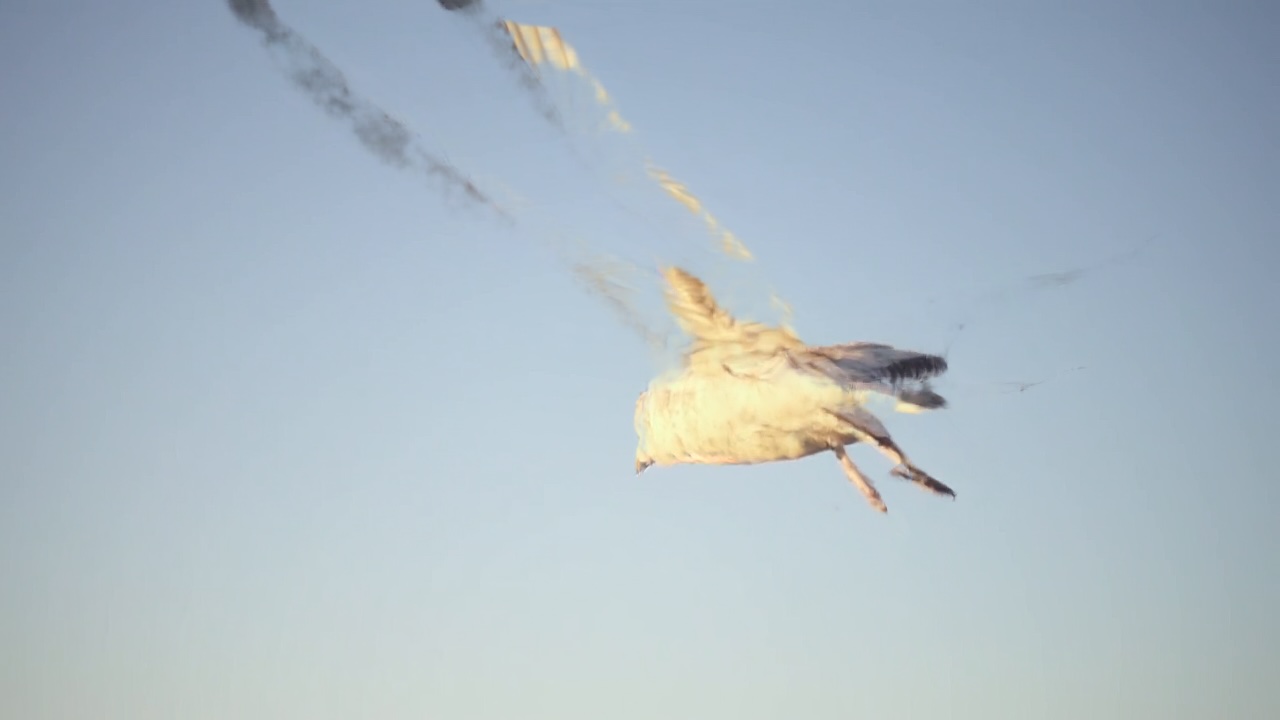}&
\includegraphics[width=0.13\linewidth,trim=380 170 100 0, clip]{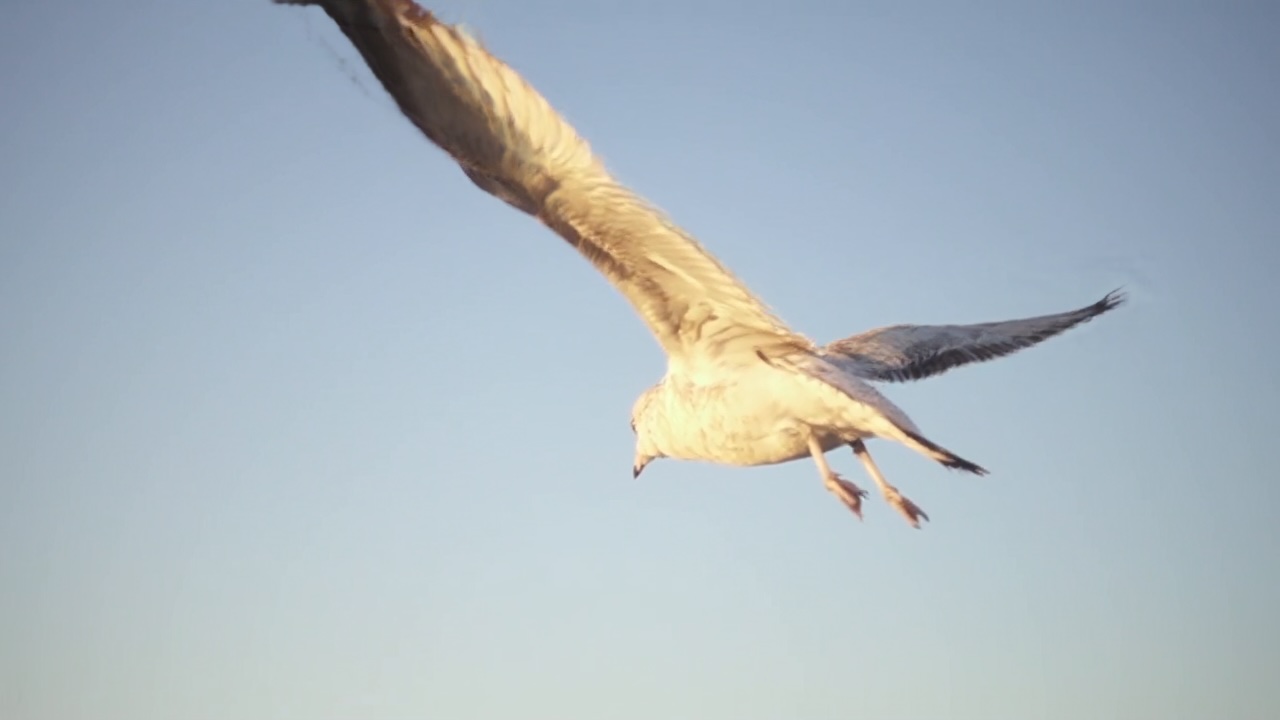}&
\includegraphics[width=0.13\linewidth,trim=380 170 100 0, clip]{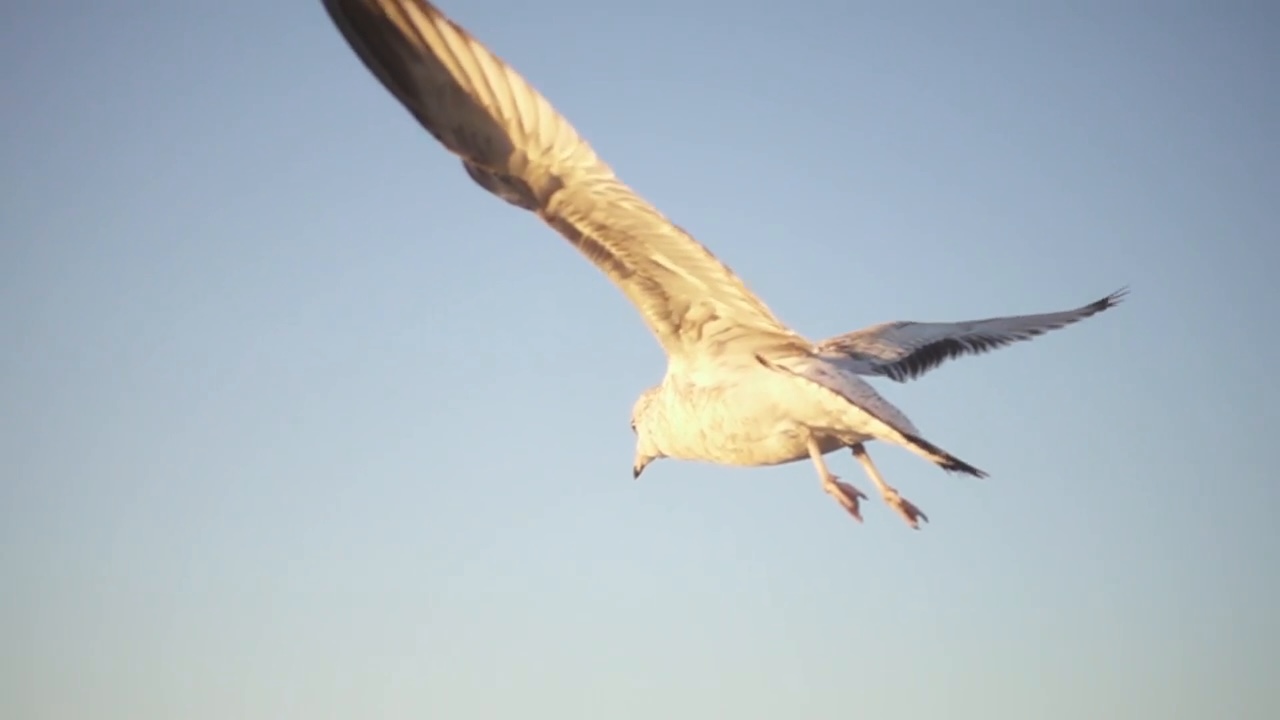}\\

\includegraphics[width=0.2\linewidth, trim=50 150 275 50, clip]{figures/compare_results/snufilm_extreme/263/input_263.jpg} &
\includegraphics[width=0.13\linewidth, trim=200 400 700 0, clip]{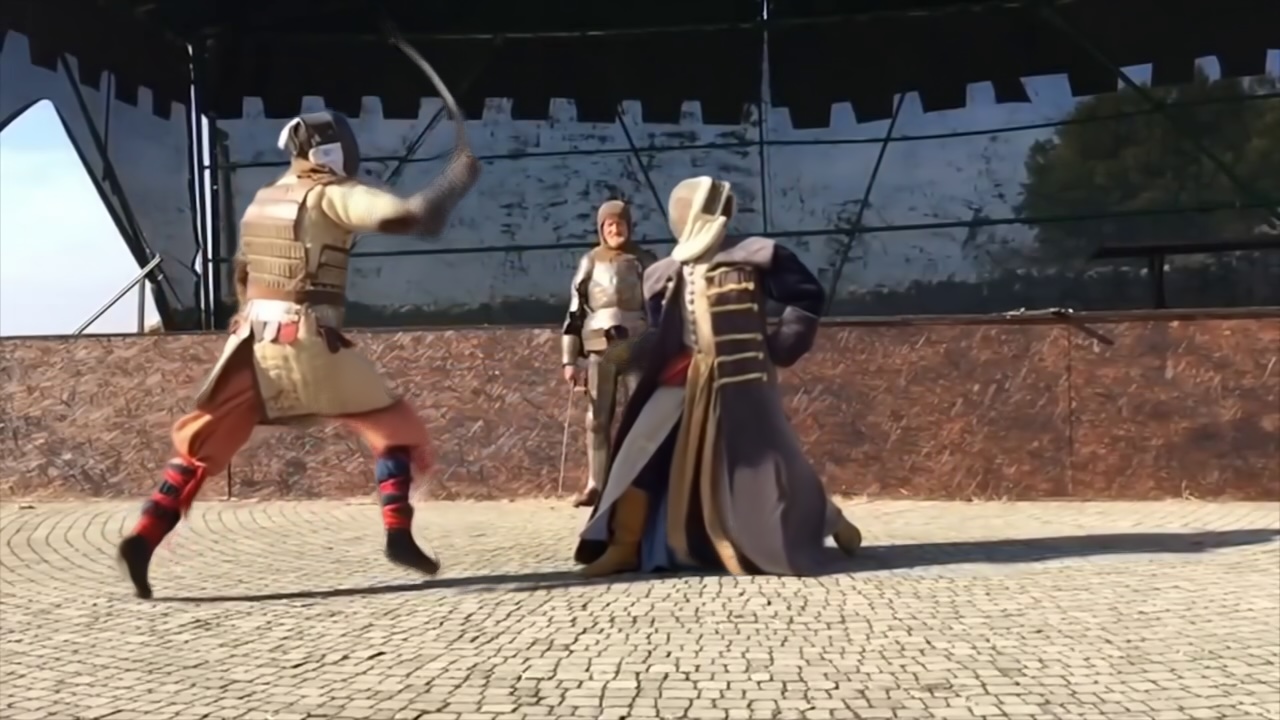}&
\includegraphics[width=0.13\linewidth, trim=200 400 700 0, clip]{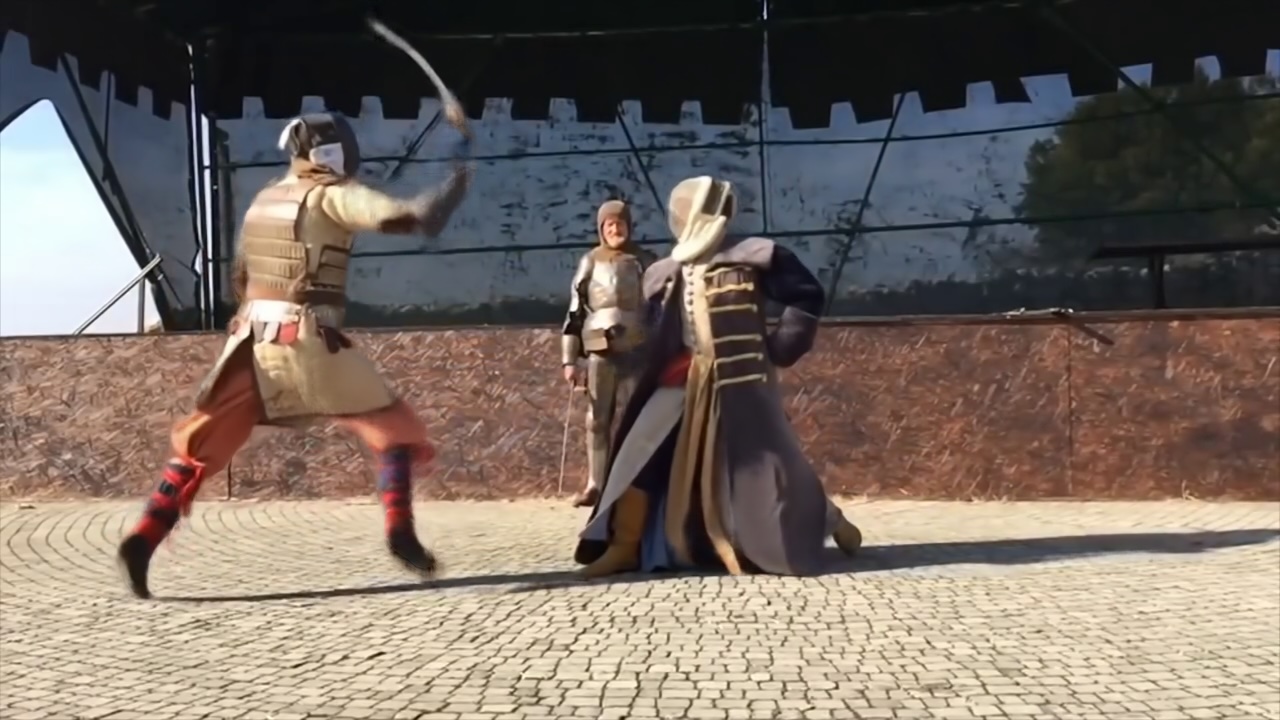}&
\includegraphics[width=0.13\linewidth, trim=200 400 700 0, clip]{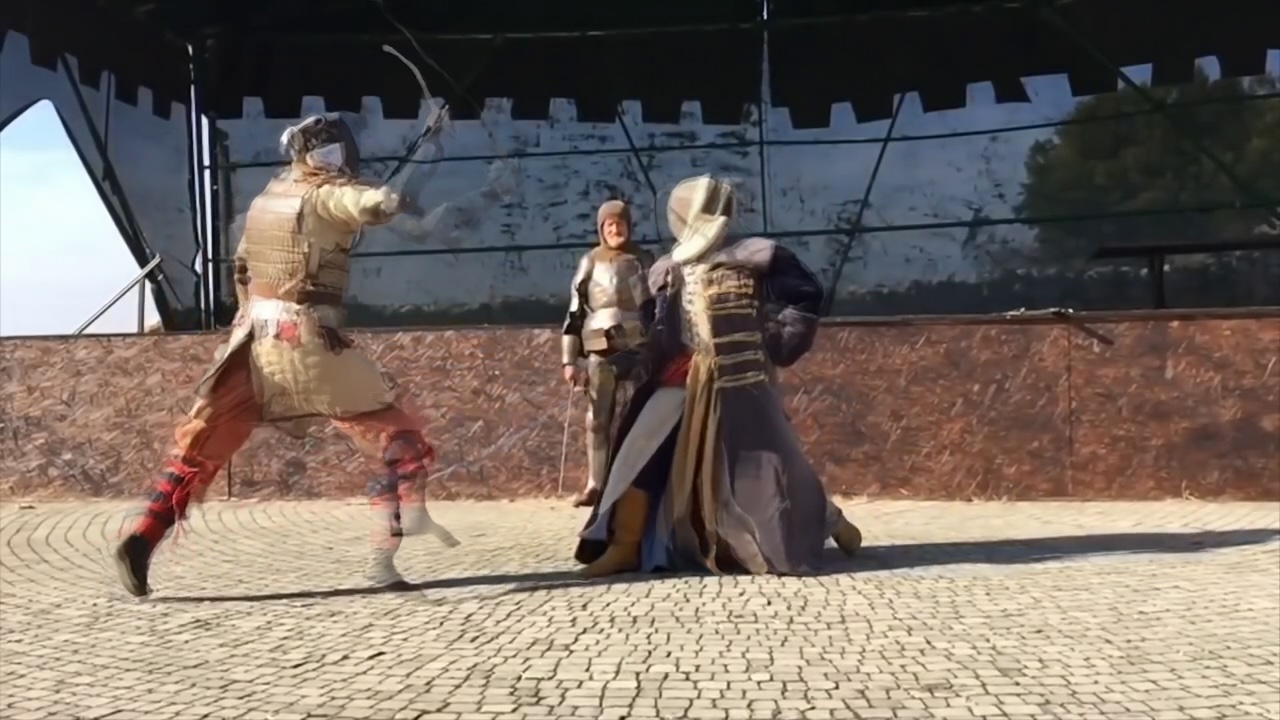}&
\includegraphics[width=0.13\linewidth, trim=200 400 700 0, clip]{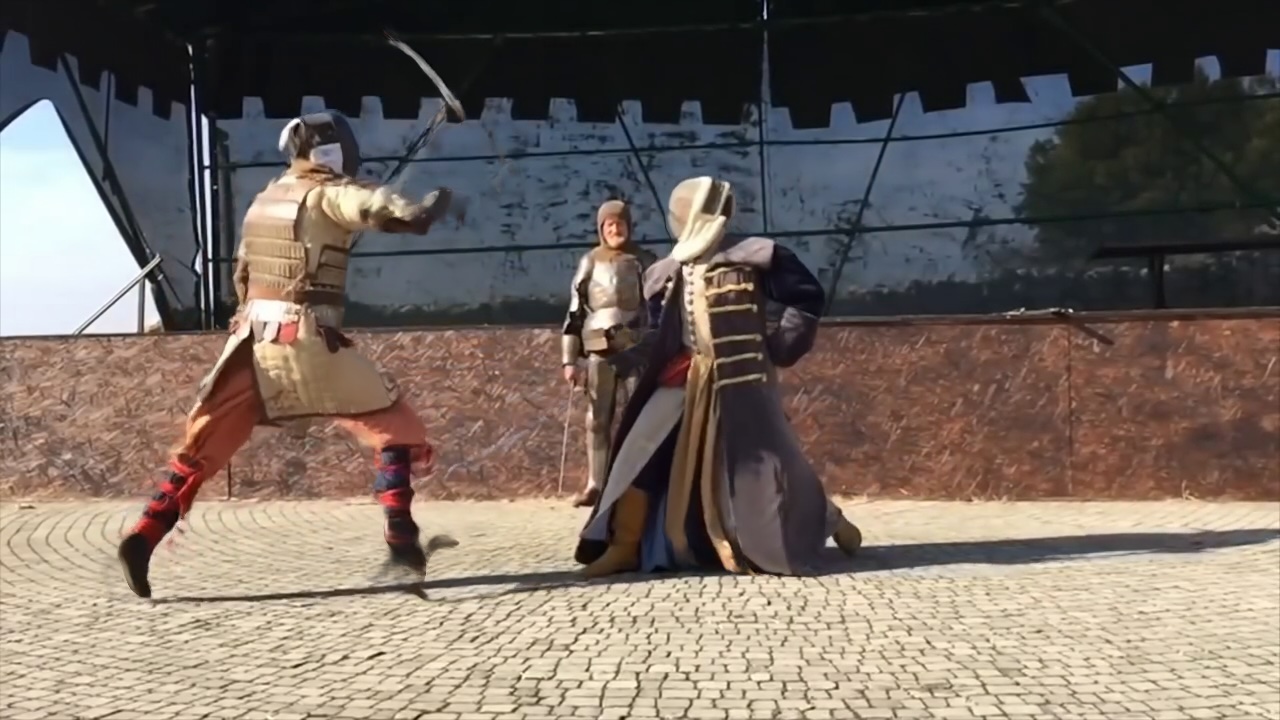}&
\includegraphics[width=0.13\linewidth, trim=200 400 700 0, clip]{figures/compare_results/snufilm_extreme/263/ours_263.jpg}&
\includegraphics[width=0.13\linewidth, trim=200 400 700 0, clip]{figures/compare_results/snufilm_extreme/263/gt_263.jpg}\\

\includegraphics[width=0.2\linewidth, trim=0 200 0 150, clip]{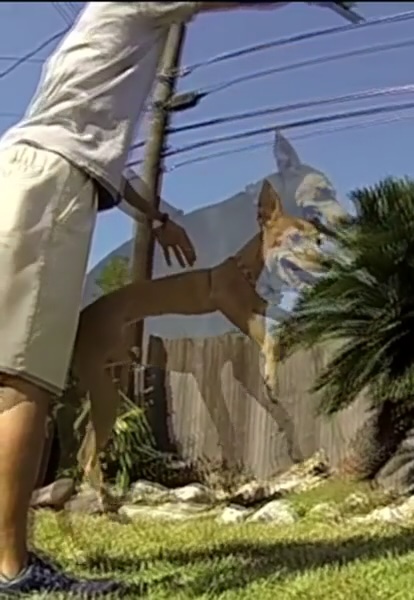} &
\includegraphics[width=0.13\linewidth, trim=200 300 50 150, clip]{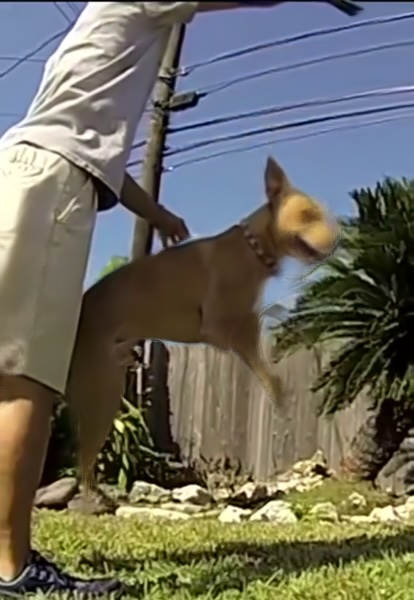}&
\includegraphics[width=0.13\linewidth, trim=200 300 50 150, clip]{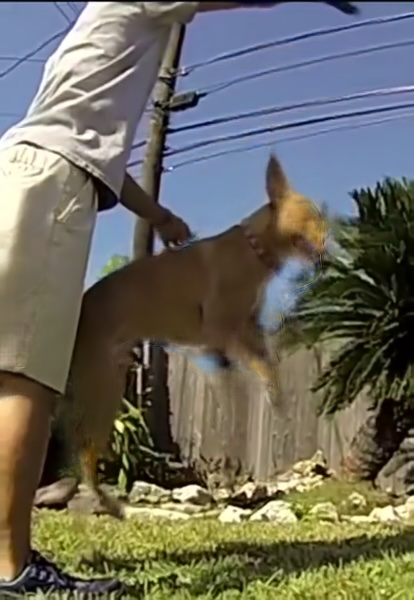}&
\includegraphics[width=0.13\linewidth, trim=200 300 50 150, clip]{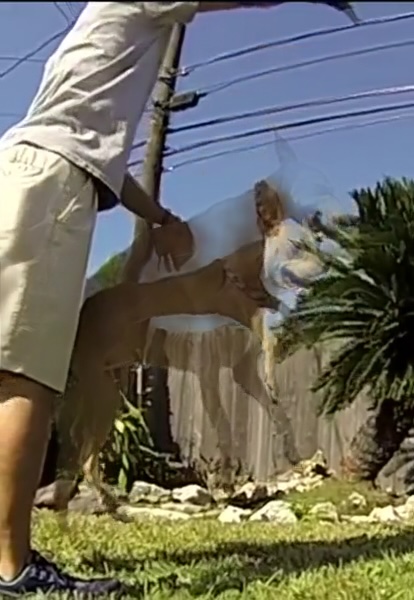}&
\includegraphics[width=0.13\linewidth, trim=200 300 50 150, clip]{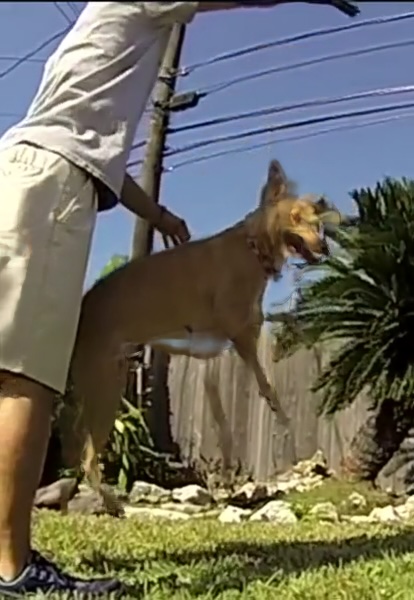}&
\includegraphics[width=0.13\linewidth, trim=200 300 50 150, clip]{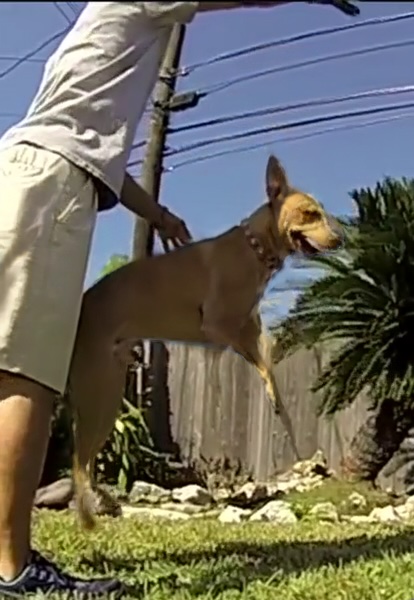}&
\includegraphics[width=0.13\linewidth, trim=200 300 50 150, clip]{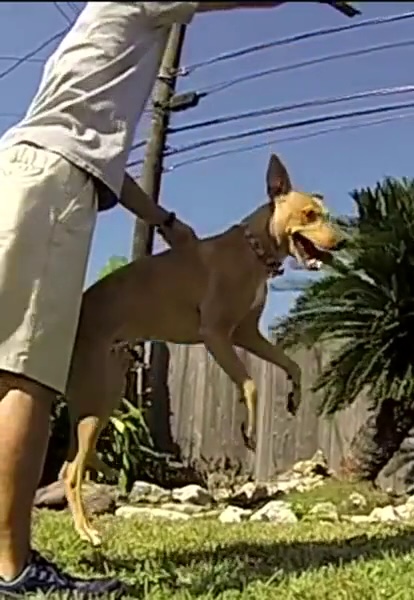}\\

\includegraphics[width=0.20\linewidth, trim=0 0 0 0, clip]{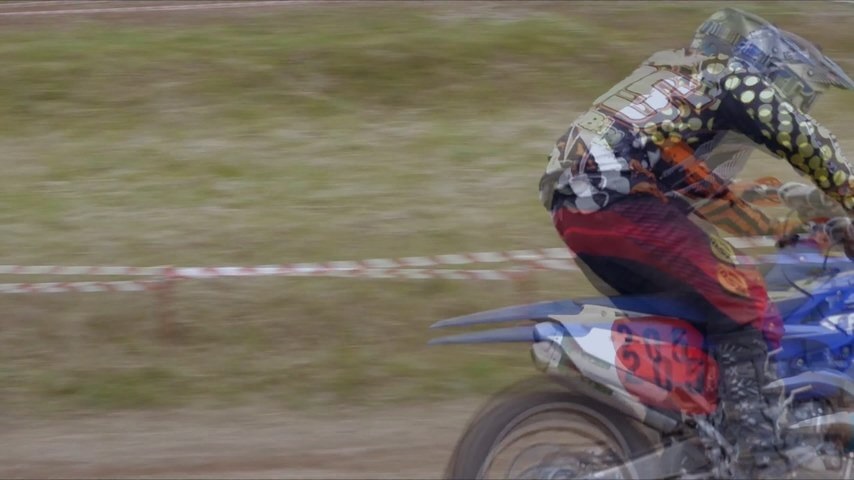} &
\includegraphics[width=0.13\linewidth, trim=300 0 0 0, clip]{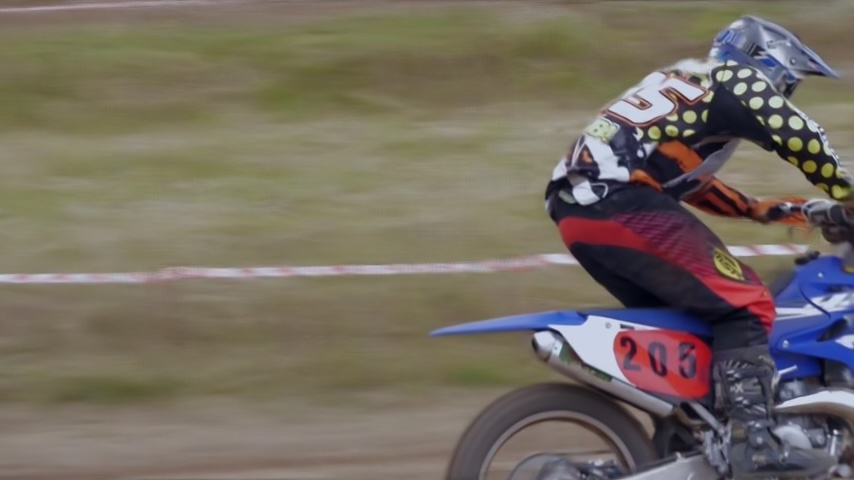}&
\includegraphics[width=0.13\linewidth, trim=300 0 0 0, clip]{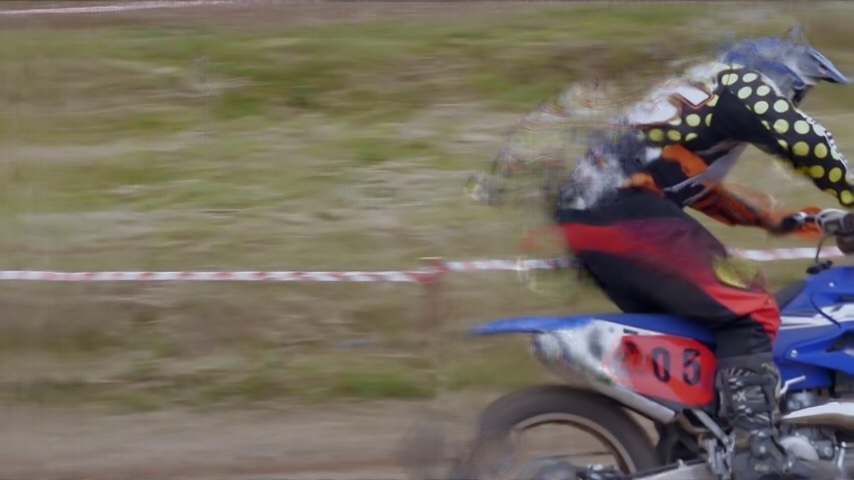}&
\includegraphics[width=0.13\linewidth, trim=300 0 0 0, clip]{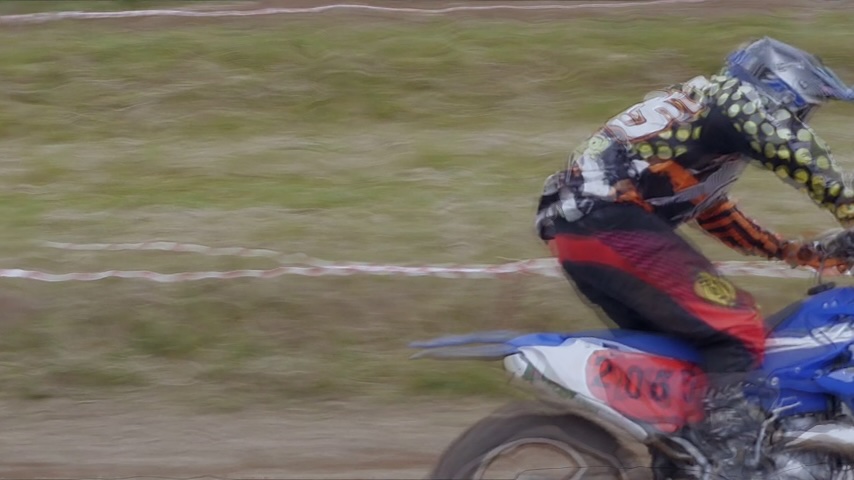}&
\includegraphics[width=0.13\linewidth, trim=300 0 0 0, clip]{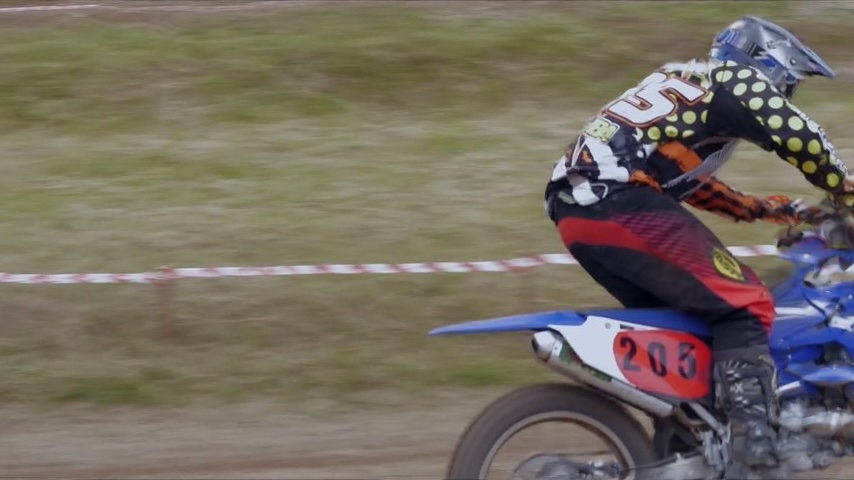}&
\includegraphics[width=0.13\linewidth, trim=300 0 0 0, clip]{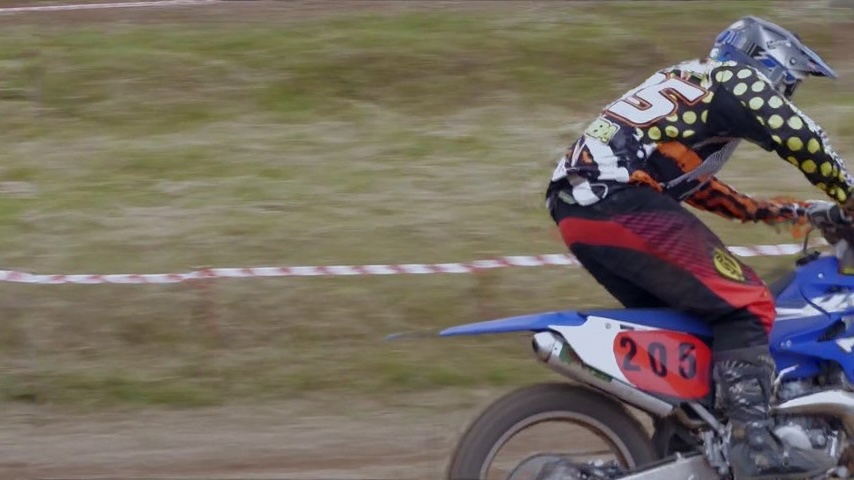}&
\includegraphics[width=0.13\linewidth, trim=300 0 0 0, clip]{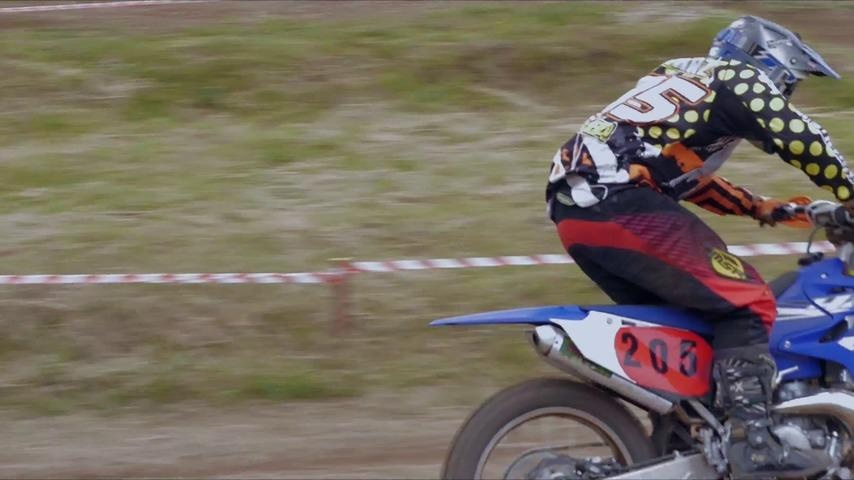}\\

\includegraphics[width=0.2\linewidth, trim=0 0 0 0, clip]{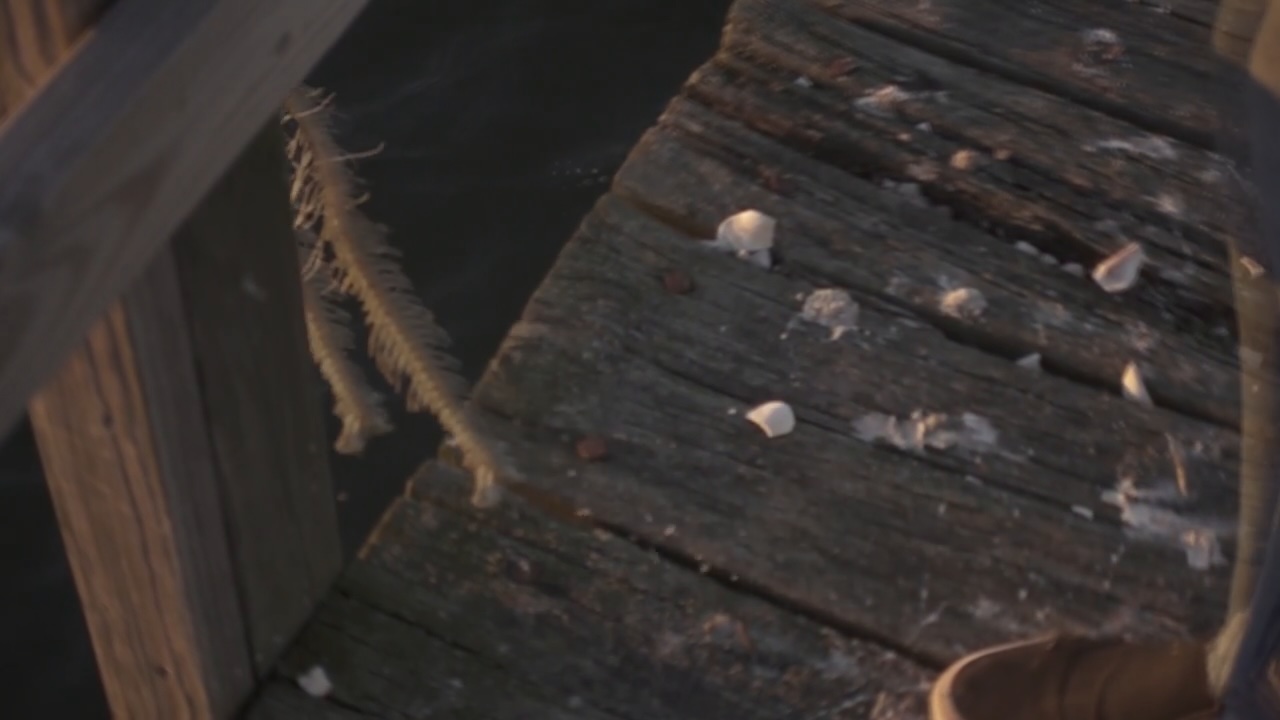} &
\includegraphics[width=0.13\linewidth, trim=200 150 450 25, clip]{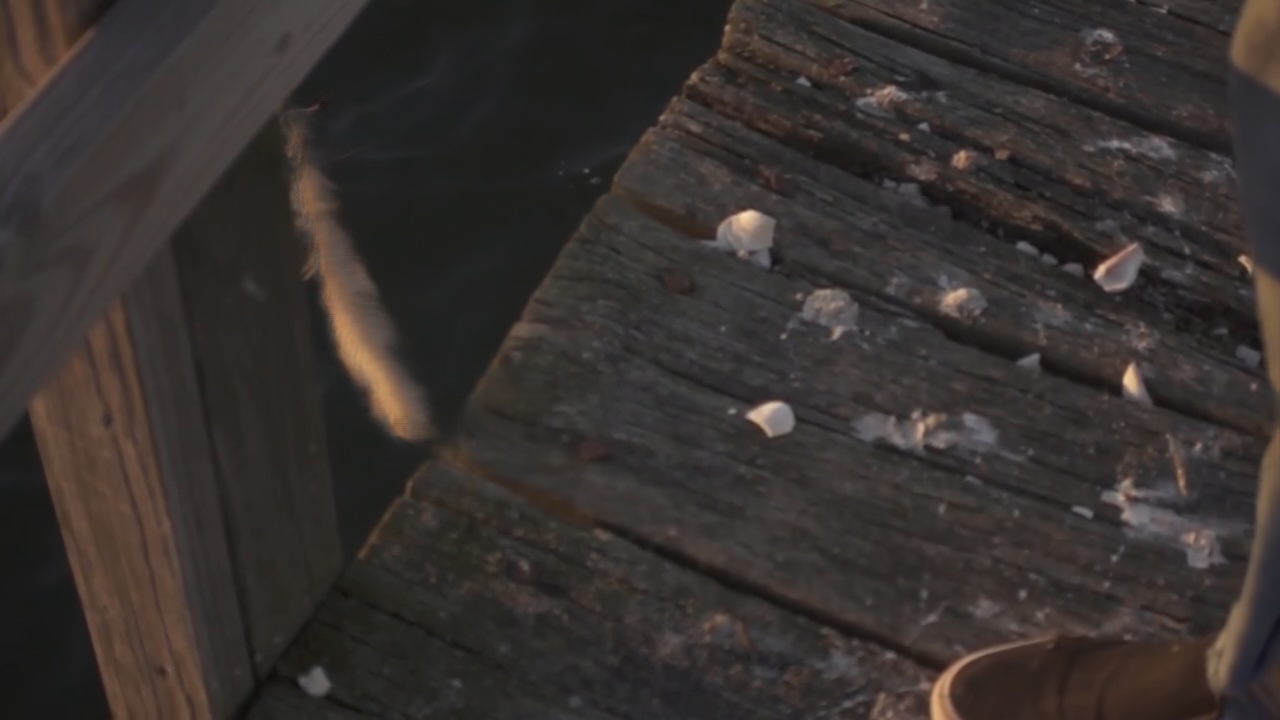}&
\includegraphics[width=0.13\linewidth, trim=200 150 450 25, clip]{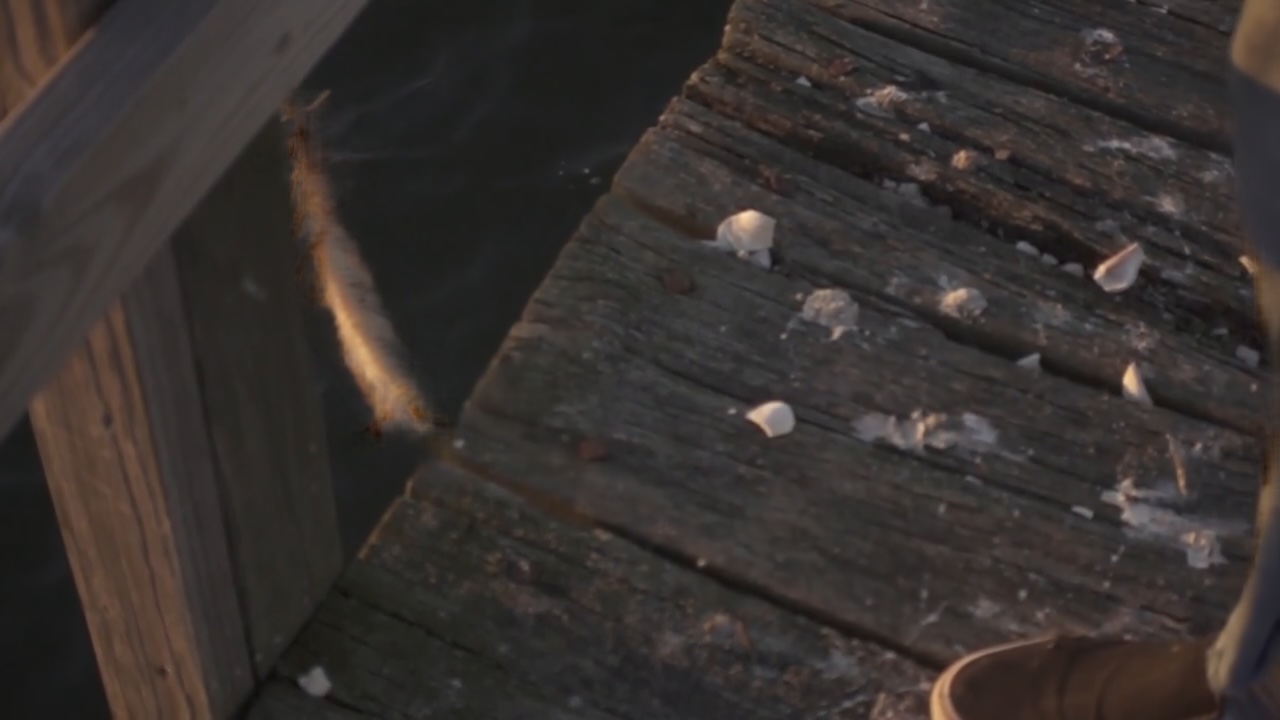}&
\includegraphics[width=0.13\linewidth, trim=200 150 450 25, clip]{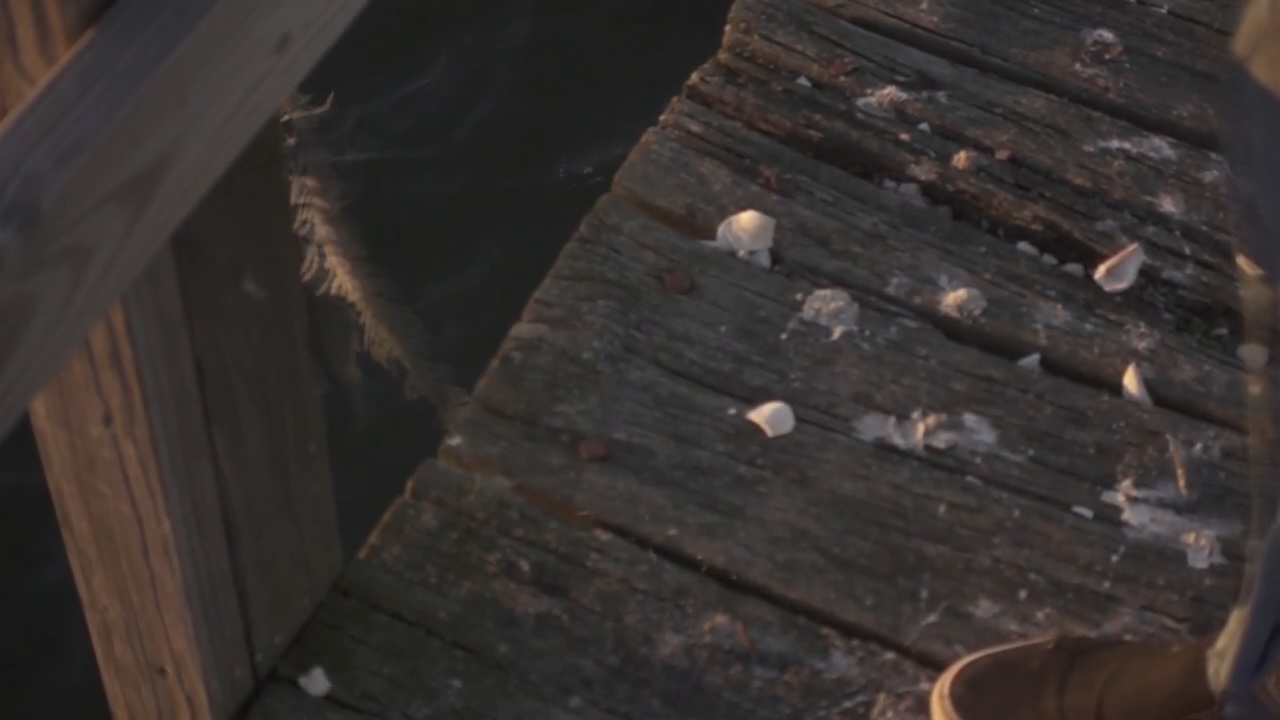}&
\includegraphics[width=0.13\linewidth, trim=200 150 450 25, clip]{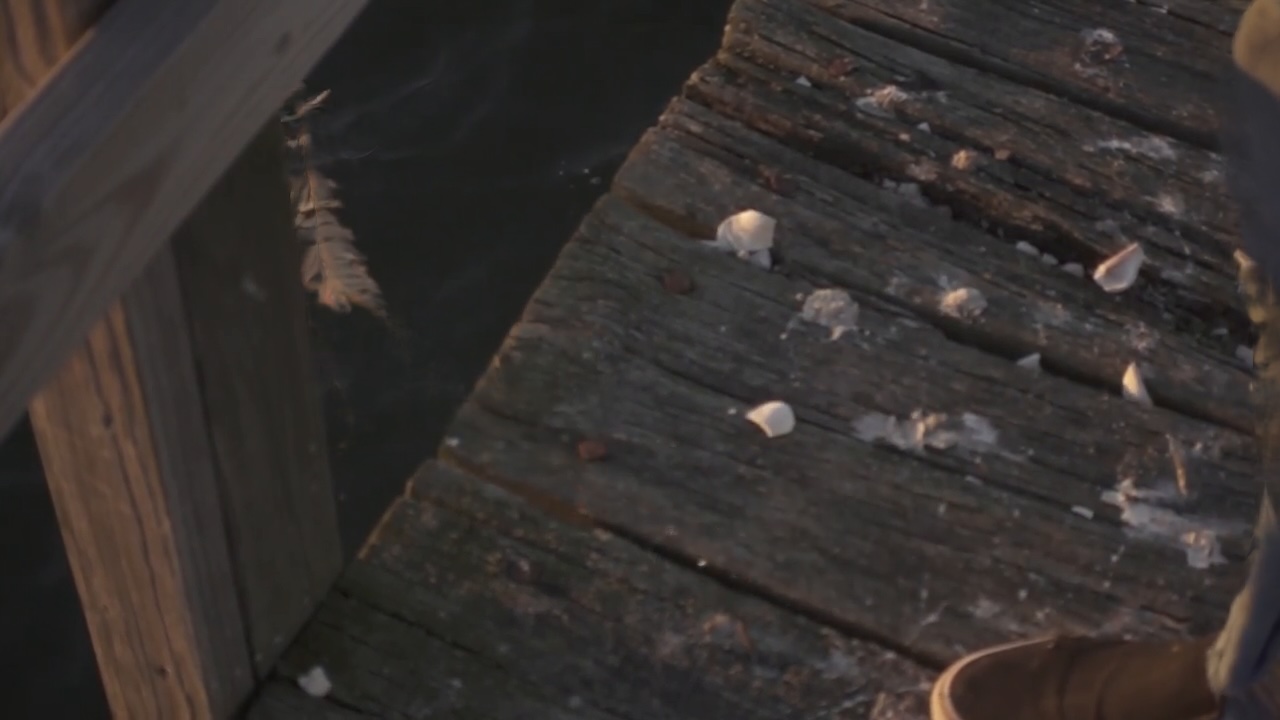}&
\includegraphics[width=0.13\linewidth, trim=200 150 450 25, clip]{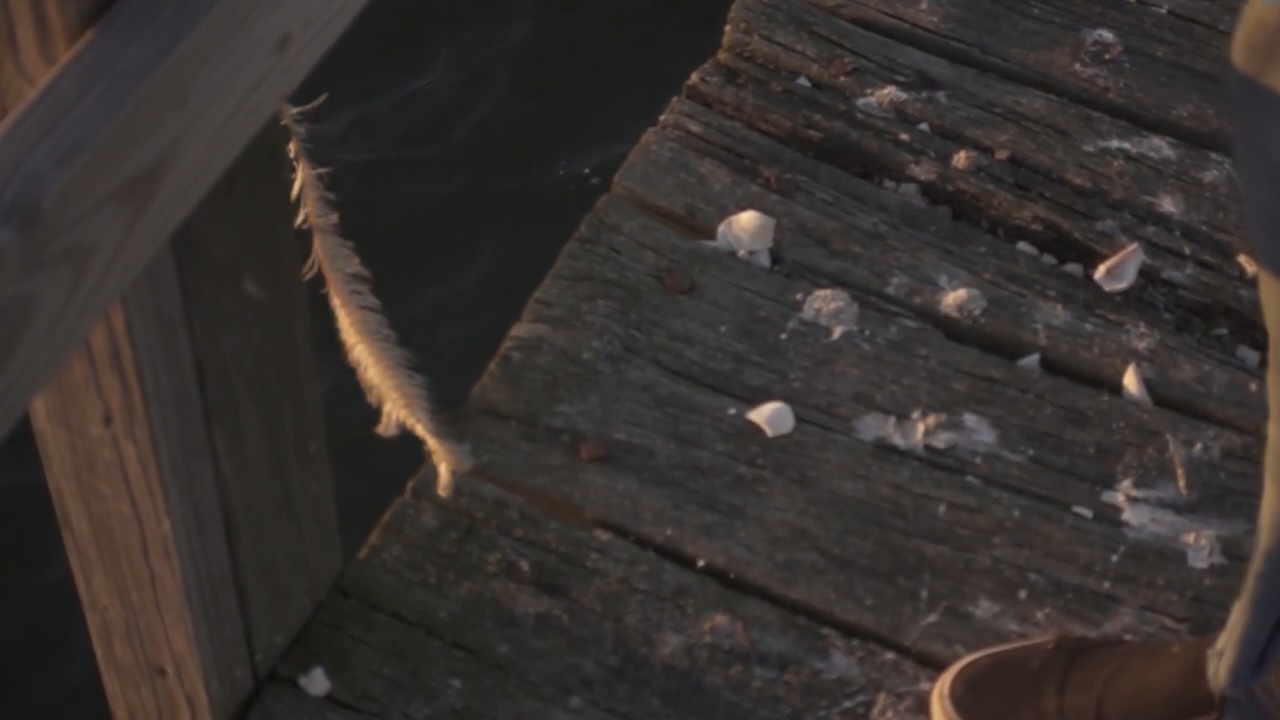}&
\includegraphics[width=0.13\linewidth, trim=200 150 450 25, clip]{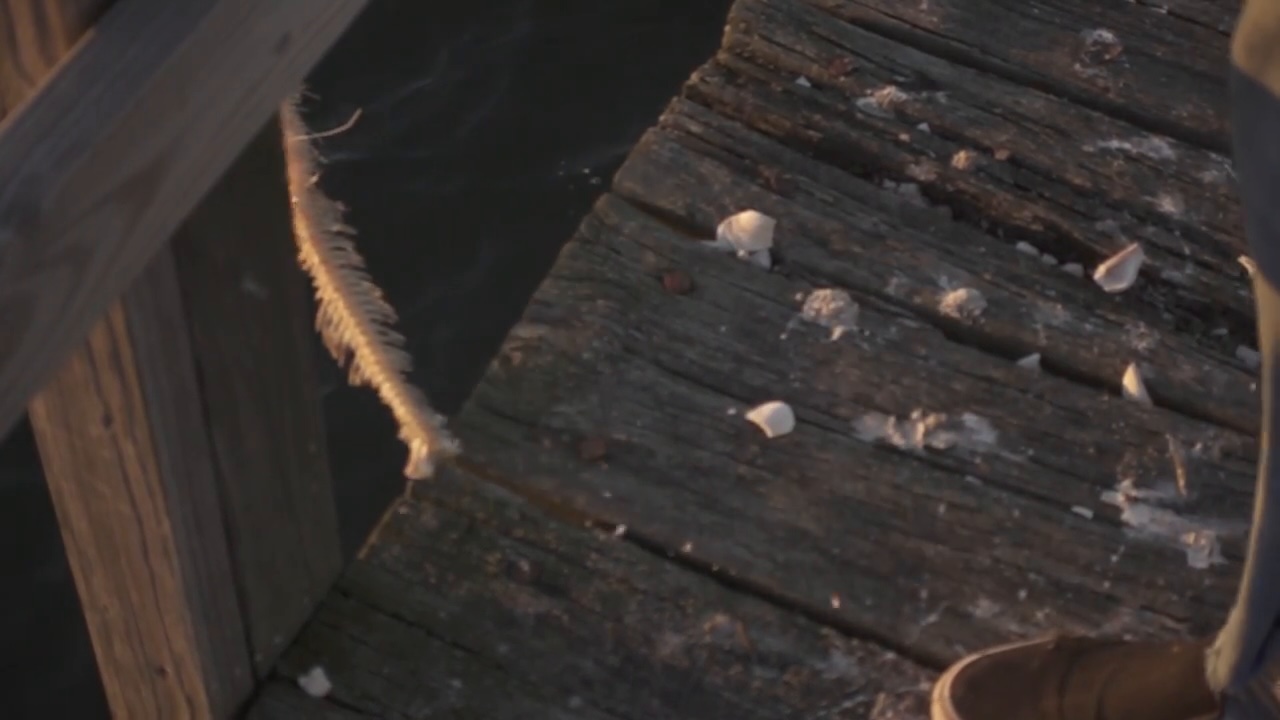}\\

\includegraphics[width=0.2\linewidth, trim=0 0 0 0, clip]{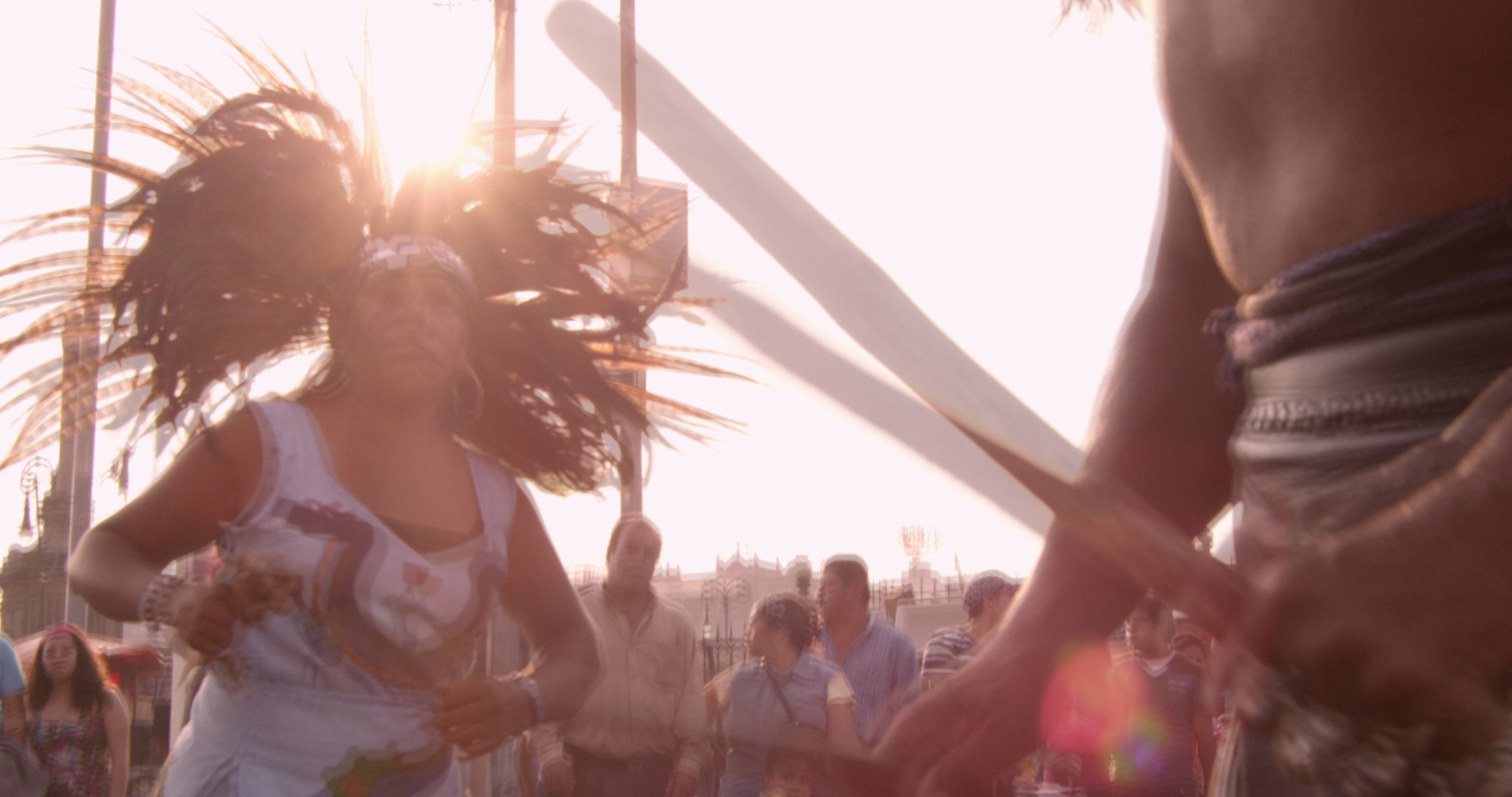} &
\includegraphics[width=0.13\linewidth, trim=850 50 50 75, clip]{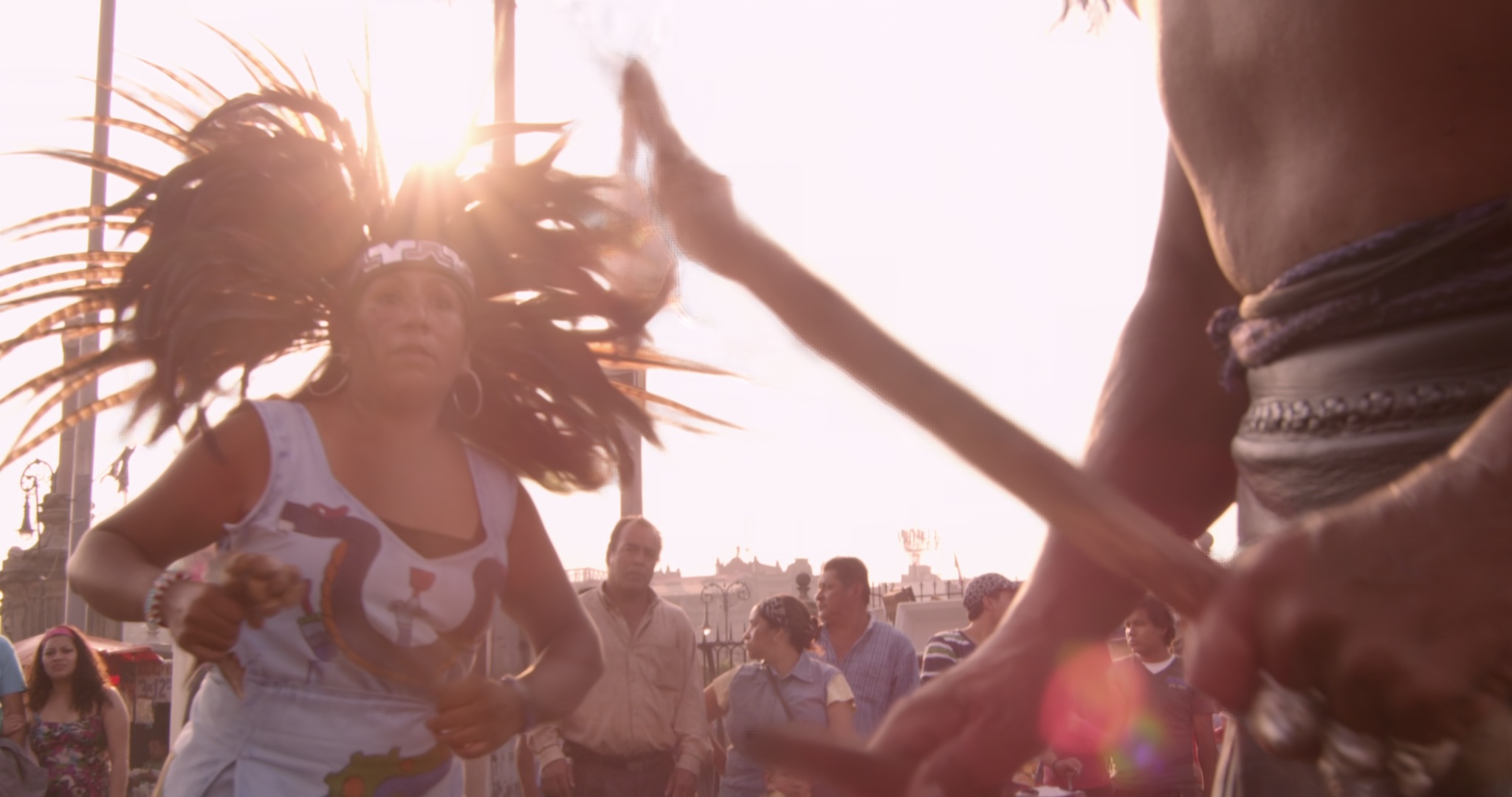}&
\includegraphics[width=0.13\linewidth, trim=850 50 50 75, clip]{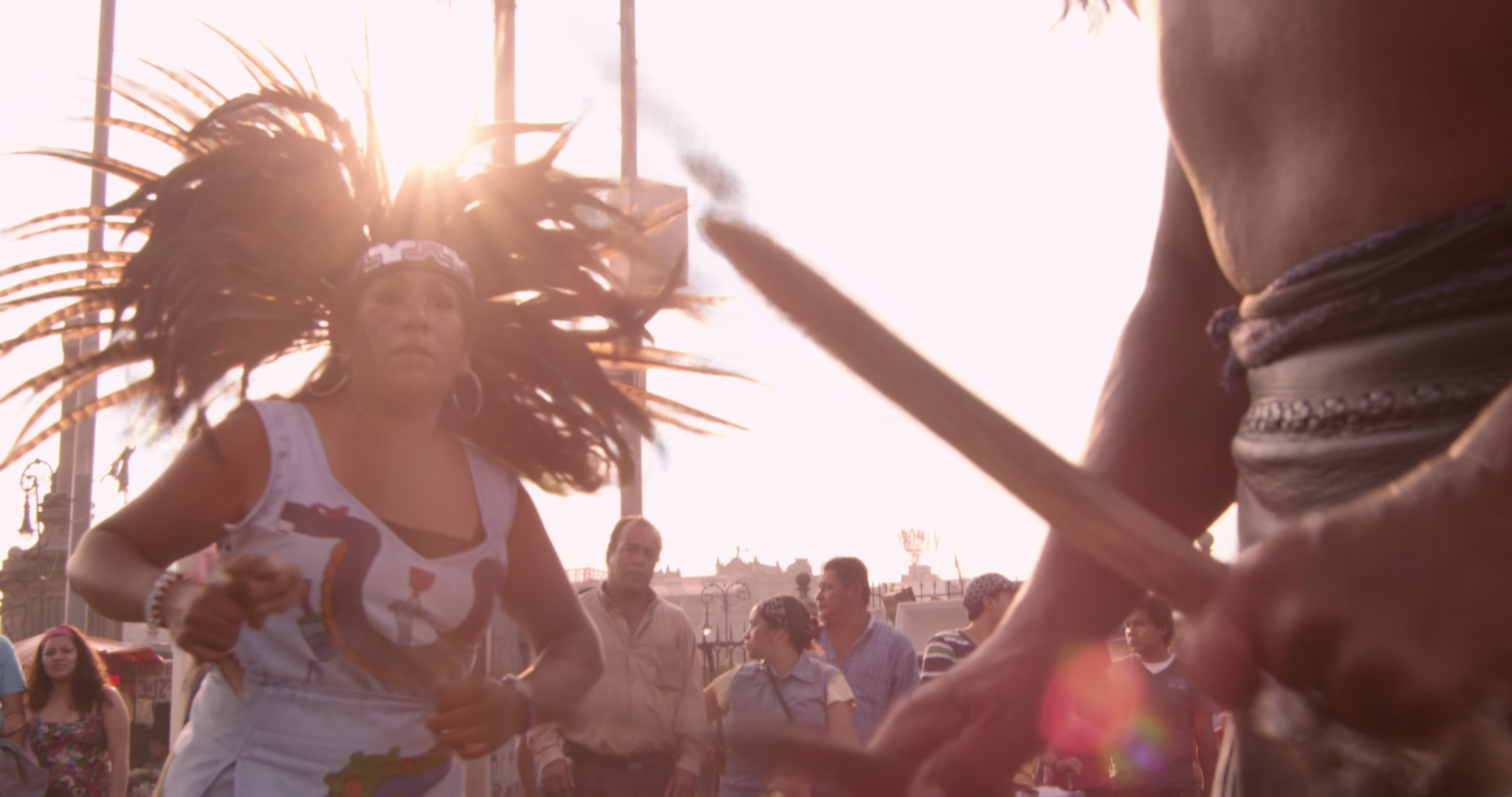}&
\includegraphics[width=0.13\linewidth, trim=850 50 50 75, clip]{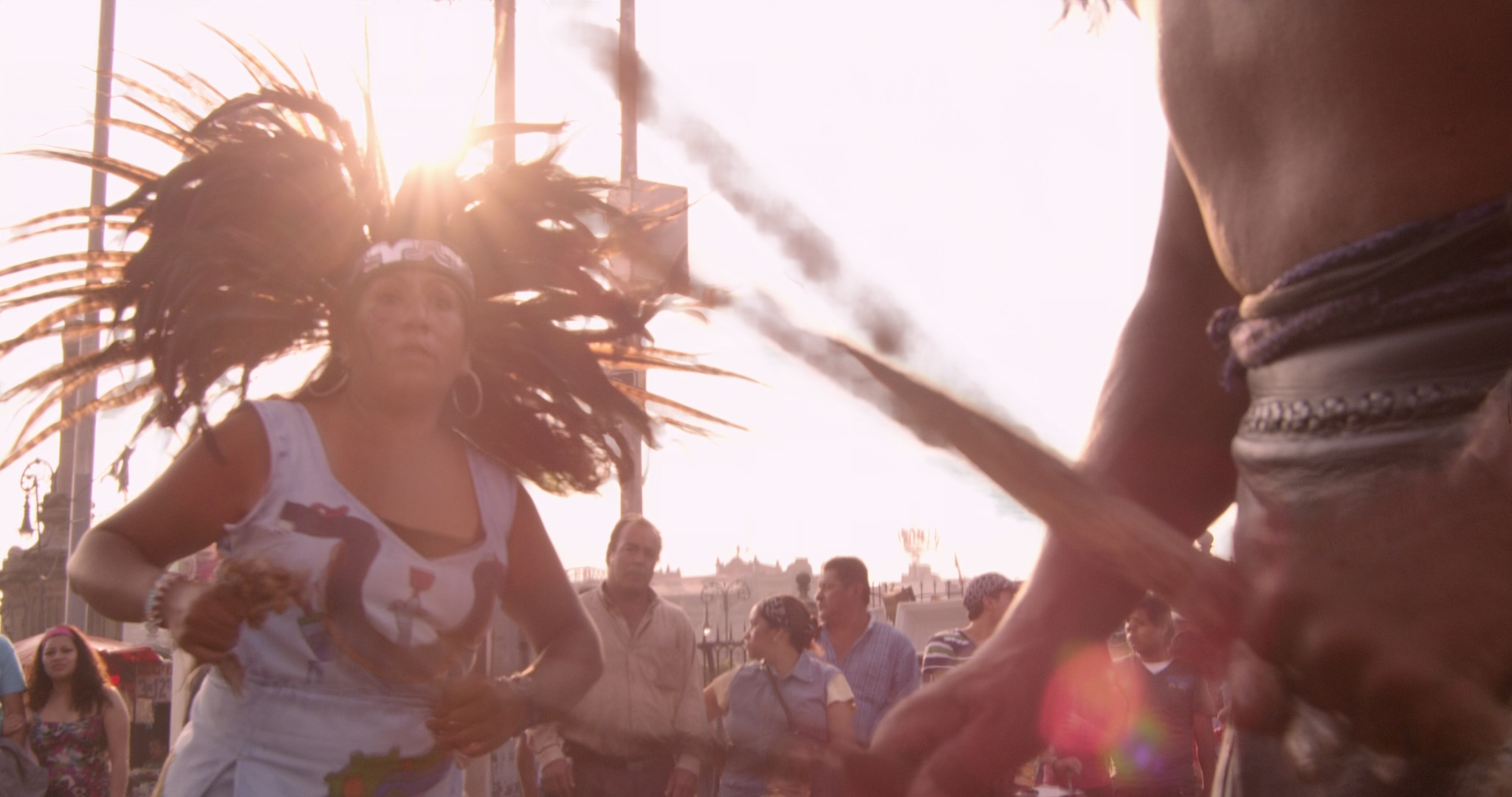}&
\includegraphics[width=0.13\linewidth, trim=850 50 50 75, clip]{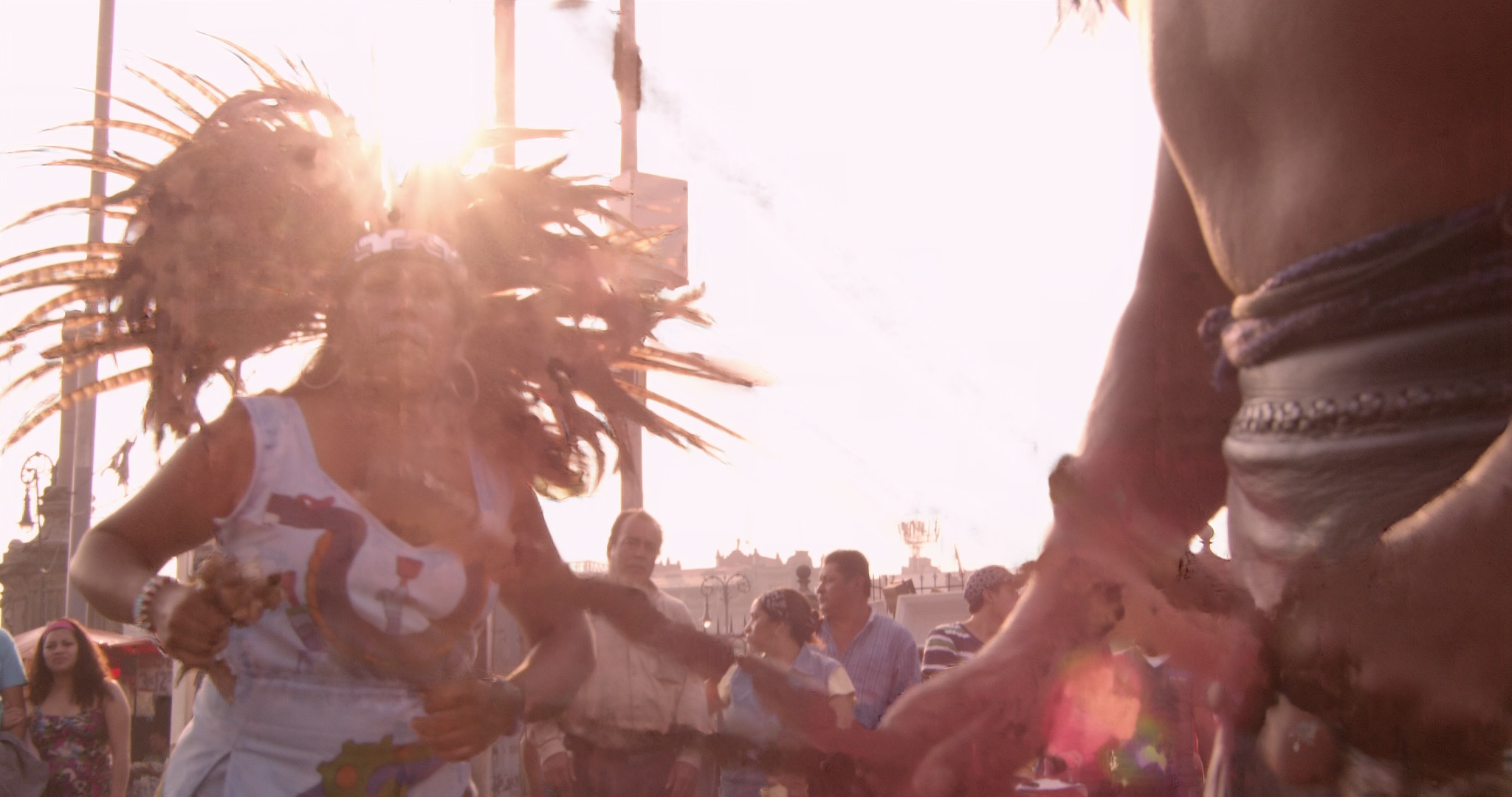}&
\includegraphics[width=0.13\linewidth, trim=850 50 50 75, clip]{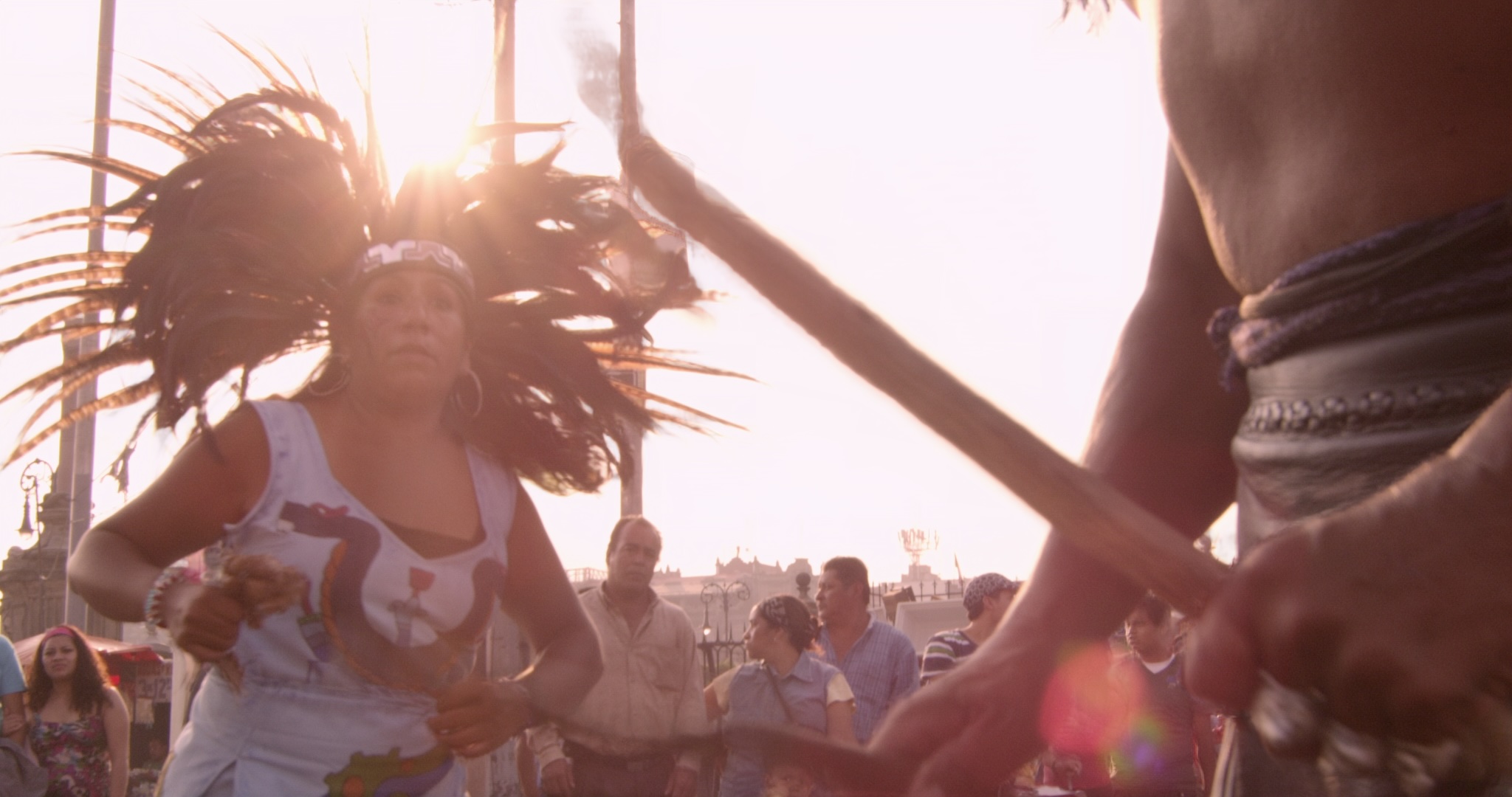}&
\includegraphics[width=0.13\linewidth, trim=850 50 50 75, clip]{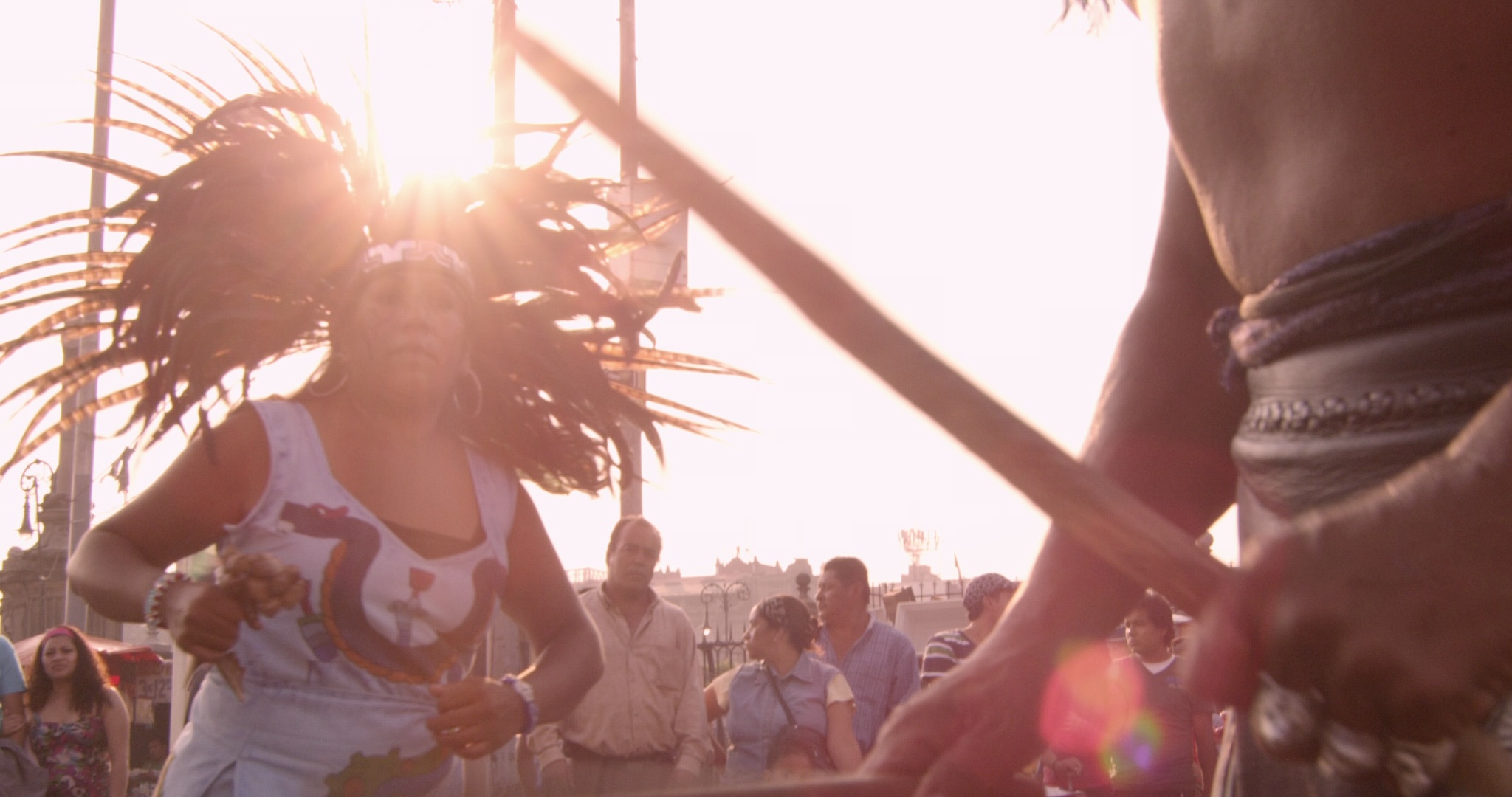}\\

\includegraphics[width=0.2\linewidth, trim=00 135 00 150, clip]{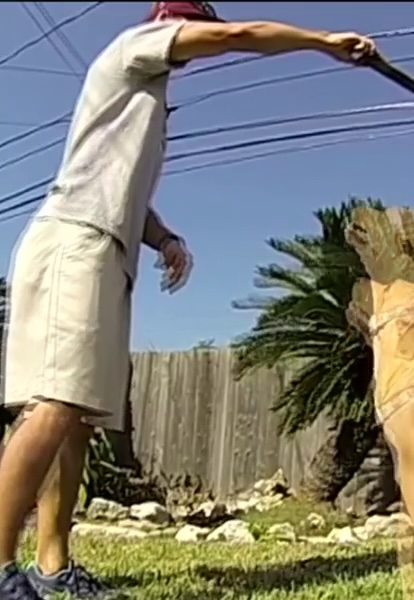} &
\includegraphics[width=0.13\linewidth, trim=200 150 0 200, clip]{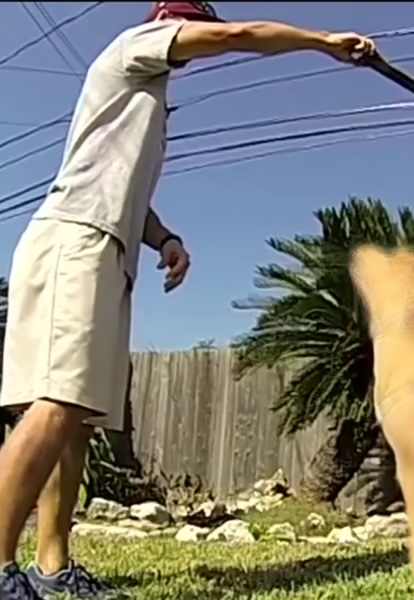}&
\includegraphics[width=0.13\linewidth, trim=200 150 0 200, clip]{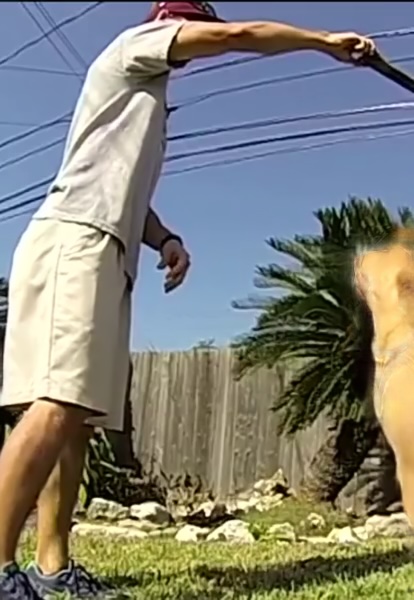}&
\includegraphics[width=0.13\linewidth, trim=200 150 0 200, clip]{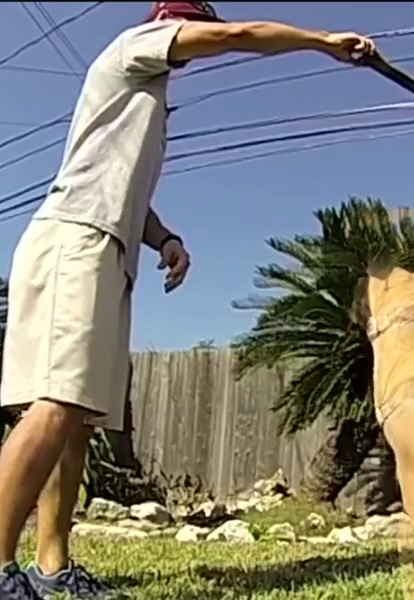}&
\includegraphics[width=0.13\linewidth, trim=200 150 0 200, clip]{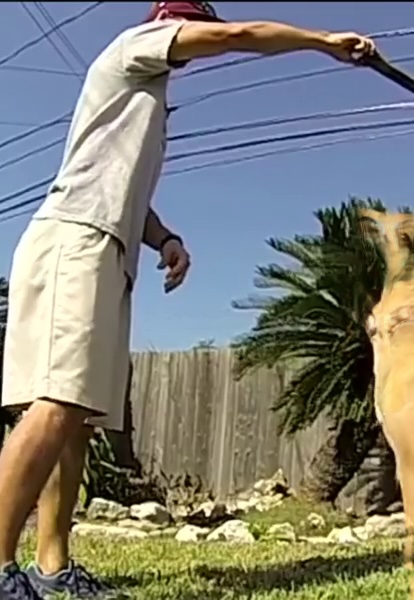}&
\includegraphics[width=0.13\linewidth, trim=200 150 0 200, clip]{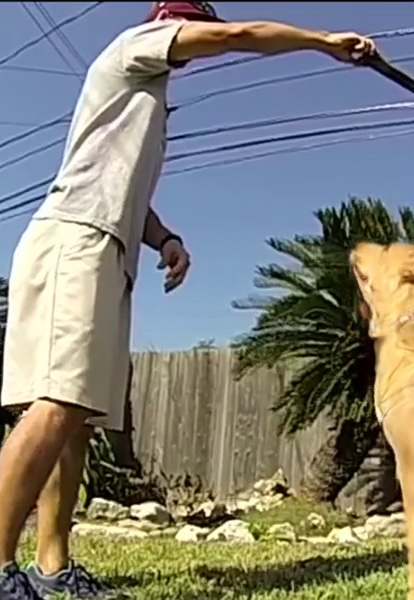}&
\includegraphics[width=0.13\linewidth, trim=200 150 0 200, clip]{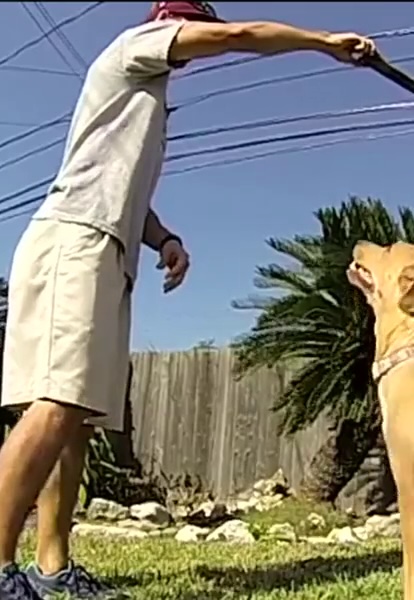}\\

   {\small Input overlay}  & {\small AMT-G~\cite{li2023amt}}       & {\small SGM-VFI~\cite{liu2024sgm-vfi}}    & {\small LDMVFI~\cite{danier2024ldmvfi}}       & {\small CBBD~\cite{lyu2024cbbd}} & {\small \bf Ours}      & {\small Ground truth} \\
\end{tabular}

    }
    \vspace{-3mm}
    \caption{{\bf Qualitative results on SNU-FILM, Xiph and DAVIS.} For complex motions, most non-diffusion-based methods (AMT, SGM-VFI) produce blurry results. However, most diffusion-based methods (LDMVFI, CBBD) struggle in handling large motions. Our method archives the best accuracy and produces high-quality results in most cases, thanks to the proposed hierarchical flow diffusion models.
    }
    \vspace{-3mm}
    \label{fig:compare_qualitative}
\end{figure*}
\vspace{-3mm}

\section{Experiments}
\label{se:exp}

We evaluate our method in this section. We first discuss implementation details of our method and the experimental settings, and then compare it with state-of-the-art methods, followed by systematic ablation studies.

\noindent \textbf{Implementation details.}
We train our method on Vimeo90k~\cite{xue2019vimeo}, which consists of 50k triplets designed for video frame interpolation.
We train the flow diffusion models with the finest resolution of 256$\times$256. For data augmentation, we randomly crop 256$\times$256 patches and perform random rotation, flipping, and frame order reversing, following previous methods~\cite{kong2022ifrnet, li2023amt, lyu2024cbbd}.
We use AdamW optimizer~\cite{AdamW_2019_iclr} and anneal the learning rate from 4e-4 to 4e-5 based on One-Cycle strategy~\cite{onecycle}, for both the training of the encoder-decoder synthesizer and the diffusion models.

We first train the encoder-decoder synthesizer for 200 epochs with a batch size of 64
and then freeze the synthesizer and train the diffusion model for another 200 epochs with the same batch size. In the final stage, we fine-tune the synthesizer and the diffusion model jointly using the photometric loss discussed in Sec.~\ref{sec:joint_optimize} for 100 epochs with a batch size of 32.

We use 3 pyramid levels for hierarchical diffusion models from the coarsest level $k_1=4$ to the finest level $k_0=2$.
The balancing loss weight in photometric loss is set to $\lambda_1=0.1$ and $\lambda_2=20$.
We use 1000 denoising steps in total for hierarchical diffusion models during training, with the step numbers on each scale roughly the same.
To accelerate the denoising process, we follow DDIM~\cite{song2021ddim} to set $\sigma_t$ in Eq.~\ref{eq:denoising} to 0, formulating it as a deterministic generative process, and perform 6 sampling steps during inference. 

During inference, we first resize the input image to a resolution having its shorter dimension equals to 256, and then extract multiscale features, to run the hierarchical diffusion models. After getting the predicted flow, we then resize it to the original image resolution, and feed it into the image synthesizer to get the final interpolated results at the raw resolution.

\noindent \textbf{Evaluation strategy.}
We train our model only on Vimeo90K, and evaluate it on Vimeo90K and other datasets, including SNUFILM~\cite{choi2020snufilm} which consists of four subsets (easy, medium, hard, and extreme) with different level of motion magnitude, Xiph~\cite{montgomery1994xiph} which contains 392 triplets in 4K resolution, and DAVIS~\cite{perazzi2016davis} with 2847 triplets with a fixed resolution of 854$\times$480. For dataset Xiph, we follow the same preprocessing step as in~\cite{niklaus2020softmax} to generate two datasets, Xiph-4K and Xiph-2K, and evaluate on them, respectively.

We evaluate the interpolated frames in LPIPS~\cite{zhang2018lpips} and FID~\cite{heusel2017fid}, which have a better correlation with human perception than PSNR and SSIM~\cite{lyu2024cbbd}. LPIPS calculates the mean squared error (MSE) distance between deep feature embeddings of image pairs, and FID computes the Fréchet distance between the feature distributions of the ground truth and predicted images.

\begin{table*}
    \centering

\begin{tabular}{l|cc|cc|ccc|ccc}
         \toprule
         \multirow{2}{*}{Method} & \multicolumn{2}{c|}{SNUFILM-hard}  & \multicolumn{2}{c|}{SNUFILM-extreme} & \multicolumn{3}{c|}{Sintel-clean}  & \multicolumn{3}{c}{Sintel-final}\\
         ~                 & LPIPS    & FID       & LPIPS     & FID   & EPE    & LPIPS  & FID    & EPE     &LPIPS   & FID  \\
         \midrule
         Vanilla              & 0.0625     & 22.283    & 0.1199    & 46.414  & 9.19 & 0.1089    & 36.079
         & 9.43 & 0.0939    & 38.011\\   
        \textbf{Ours}     & 0.0405    & 15.320     &  0.0839       & 27.032   & 5.70    & 0.0789    & 23.476    & 6.46  & 0.0730    & 28.453    \\
         Oracle            & {\bf 0.0264}    & {\bf 11.050}     &  {\bf 0.0424}       & {\bf 20.710}     & {\bf 2.67} & {\bf 0.0427}    &{\bf 19.083}    & {\bf 3.89}  & {\bf 0.0385}    & {\bf 21.512}            \\
         \midrule
         EMA-VFI        & 0.0579    & 20.679    & 0.1099    & 39.051 & 6.79     & 0.1049    & 28.657    & 7.28  & 0.0767    & 33.948\\
         EMA-VFI + {\bf Ours}            & {\bf 0.0521}    & {\bf  19.144}     &  {\bf 0.0968}       & {\bf 35.346}  & 6.79 & {\bf 0.0863}    & {\bf 26.649}    & 7.28  & {\bf 0.0754}    & {\bf 29.908}      \\
         \bottomrule
\end{tabular}
    \vspace{-3mm}
    \caption{{\bf Relation between optical flow and frame interpolation.}
    ``Vanilla" is the setting without our flow diffusion but relying on RAFT to compute the forward and backward flow between the two input frames and multiply by a factor of 0.5 to produce the bilateral flow for the image synthesizer. ``Oracle'' is the upper bound setting that uses the ground truth interpolation frame to compute the bilateral flow with RAFT. ``EMA-VFI + Ours'' uses the bilateral flow results of EMA-VFI~\cite{zhang2023ema_vfi} as the input of our image synthesizer. Our method achieves much better results in both bilateral flow estimation and frame interpolation.
    }
    \label{tab:effect_flow}
    \vspace{-3mm}
\end{table*}

\subsection{Comparison to the State of the Art}
We compare our method with state-of-the-art methods, including VFIformer~\cite{lu2022vfiformer}, AMT~\cite{li2023amt}, SGM-VFI~\cite{liu2024sgm-vfi}, EMA-VFI~\cite{zhang2023ema_vfi}, URPNet~\cite{jin2023urpnet}, and Per-VFI~\cite{wu2024pervfi}. We also compare our method with the recent diffusion-based methods, including LDMVFI~\cite{danier2024ldmvfi}, MADiff~\cite{huang2024madiff}, and CBBD~\cite{lyu2024cbbd}.
We report the quantitative results in Table~\ref{tab:compre_snufilm},~\ref{tab:compare_xiph}, and~\ref{tab:compare_others_benchmark}. 
On most datasets, our method outperforms most state-of-the-art methods significantly, especially on more challenging benchmarks such as the extreme subset of SNUFILM and Xiph-4K.
Fig.~\ref{fig:compare_qualitative} shows qualitative comparison results on SNUFILM, Xiph and DAVIS. Most competitors frequently produce results with blurred results or significant artifacts. By contrast, our method is superior in handling large movements, recovering subtle details, and producing high-quality frame interpolation results.

In addition to the accuracy, we evaluate the efficiency of our method. Since different methods report the running time on different machines in their paper, we run LDMVFI, CBBD, SGM-VFI, and our method on the same workstation with an RTX-4090 GPU, and set the input image pair at a resolution of 1024$\times$1024.
As discussed in Fig.~\ref{fig:teaser}, LDMVFI, CBBD, SGM-VFI, and our method finish the processing in 8.3s, 2.1s, 0.19s, and 0.20s, respectively. Our method is 10+ times faster than the diffusion-based methods LDMVFI and CBBD, and on par with SGM-VFI in efficiency. While note that, our method achieves much higher accuracy than SGM-VFI, as in Table~\ref{tab:compre_snufilm} and~\ref{tab:compare_xiph}.

Our framework has 0.35M, 0.59M, and 46.96M parameters for the encoder, decoder, and hierarchical diffusion network, respectively, and it only consumes $\sim$2.9 GB VRAM in float32 mode during inference for a typical 1024x1024 image pair, which is significantly more resource-friendly than other diffusion-based method LDMVFI ($\sim$6.9 GB) and CBBD ($\sim$8.4 GB).

\begin{figure}
    \centering
    \begin{tabular}{cc}
        \includegraphics[width=0.4\linewidth]{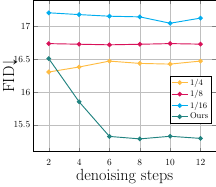} &
        \includegraphics[width=0.4\linewidth]{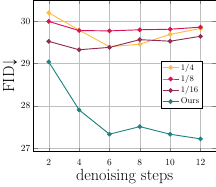} \\
        {\small (a) SNUFILM-hard}        & {\small (b) SNUFILM-extreme} \\
    \end{tabular}
    \vspace{-3mm}
    \caption{{\bf Analysis of hierarchical diffusion models.}
    We compare our hierarchical diffusion method with its single-level versions, with which the diffusion only happens at the corresponding level (denoted as ``1/16'', ``1/8'', and ``1/4'' respectively). Our hierarchical method improves consistently with more denoising steps.
    }
    \vspace{-3mm}
    \label{fig:compare_step_scale}
\end{figure}

\begin{table*}
    \centering


\begin{tabular}{c|cc|cc|cccccc}
        \toprule
         \multirow{2}{*}{Resolution}   & \multicolumn{2}{|c|}{SNUFILM-hard}  & \multicolumn{2}{|c|}{SNUFILM-extreme}      & \multicolumn{6}{c}{Runtime}\\
         ~   & LPIPS        & FID                & LPIPS        & FID                   & Encoder   & scale-1/16  & scale-1/8   & scale-1/4   & Decoder   & Total \\
         \midrule
         128 $\times$128   & 0.0427    & 16.737            & 0.0887        & 29.774                & {\bf 2.57}       & {\bf 17.43}      & {\bf 18.48}     & {\bf 18.77}     & {\bf 7.93}     & {\bf 65.31}   \\
         256 $\times$ 256   & \underline{0.0405}    & \underline{15.320}            & \underline{0.0842}        & \underline{27.356}               & \underline{2.94}      & \underline{18.23}       & \underline{18.56}     & \underline{18.90}     & \underline{39.31}     & \underline{97.82}    \\
         512 $\times$ 512   & {\bf 0.0397}    & {\bf 15.026}            & {\bf 0.0836}        & {\bf 26.925}                & 7.45      & 18.24      & 18.78     & 28.64     & 64.96     &  138.06 \\
         \bottomrule
         
\end{tabular}
    \vspace{-3mm}
    \caption{{\bf Inference effect of different training resolutions.}
    We evaluate the inference effect of our model trained with different resolutions. Higher training resolution leads to better performance overall, but resulting in increased inference time. We report the inference time of each stage of our method (in milliseconds), including the different levels in the hierarchical diffusion models (denoted as scale-1/16, scale-1/8, and scale-1/4 respectively). 256x256 gives the best balance between efficiency and accuracy.
    }
    \label{tab:reslution_effect}
    \vspace{-3mm}
\end{table*}

\subsection{Ablation Study}
\label{sec:abltation}

\noindent \textbf{Relation between optical flow and frame interpolation.}
We first conduct ablation studies on the relation between optical flow and frame interpolation on SNU-FILM.
Since there is no ground truth flow in VFI dataset, we also conduct the same ablation on a typical optical flow dataset MPI-Sintel~\cite{butler2012sintel} which includes GT flow and two tracks ``clean'' and ``final''. On Sintel, we report the bilateral flow results in end-point error (EPE).
As shown in Table.~\ref{tab:effect_flow}, ``Vanilla'' struggles to capture complex motion patterns in estimating bilateral flow and cannot produce reasonable interpolation results. Our method achieves much better results in both bilateral flow estimation and frame interpolation. 
On the other hand, we compare EMA-VFI and a version with the bilateral flow results of EMA-VFI directly as the input of our image synthesizer (``+ Ours''). With the same bilateral flow, our method is more accurate in frame interpolation, thanks to the proposed image synthesizer.

\noindent \textbf{Result analysis of hierarchical diffusion models.}
We compare our hierarchical diffusion method with its single-level versions, with which the diffusion only happens at the corresponding level. 
We evaluate their performance with different sampling steps on SUMFILM-hard and SNUFILM-extreme, as illustrated in Fig.~\ref{fig:compare_step_scale}. 
Our hierarchical diffusion strategy outperforms all single-level baselines significantly. 
The single-level baselines struggle to converge, and additional denoising steps bring little improvement. By contrast, our hierarchical diffusion consistently benefits from more denoising steps.
This also highlights the robustness of our hierarchical diffusion to different motion types and complex scenes.

\begin{figure}
    \centering
    \begin{tabular}{cc}
        \includegraphics[width=0.4\linewidth]{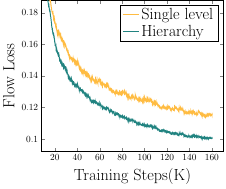} &
        \includegraphics[width=0.4\linewidth]{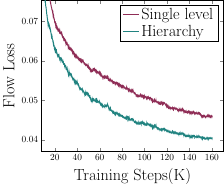} \\
        {\small (a) scale-1/4}      & {\small (b) scale-1/16} \\
    \end{tabular}
    \vspace{-3mm}
    \caption{
    {\bf Training analysis of hierarchical diffusion model.} The proposed hierarchical flow diffusion model has better convergence compared with versions with a fixed resolution on each pyramid level only, which helps to produce better interpolation results with our flow-guided image synthesizer.
    }
    \vspace{-3mm}
    \label{fig:loss_vis}
\end{figure}

\noindent \textbf{Training analysis of hierarchical diffusion models.}
We report the flow loss value of our hierarchical diffusion model during the training on Vimeo-90k, and compare it with versions with a fixed resolution on a single pyramid level, as shown in Fig.~\ref{fig:loss_vis}. The proposed hierarchical diffusions model converges better than the single-level versions, which helps to produce better interpolation results with our flow-guided image synthesizer.

\noindent \textbf{Effect of flow representation in diffusion.}
Previous methods~\cite{danier2024ldmvfi, lyu2024cbbd} use diffusion models to denoise the latent space directly. By contrast, we propose to use diffusion models to denoise an intermediate flow representation.
To study the effect of these two parametrization methods within a unified framework, we adapt our encoder-decoder based synthesizer to a latent-based version.
Since there is no intermediate flow, instead of using the warped feature pair, we directly concatenate the encoder features and the decoder feature on corresponding levels.
As shown in the 1st and 4th row of Table.~\ref{tab:compare_optimization}, without modeling an intermediate flow representation for the encoder-decoder, the performance deteriorates significantly, with the FID error increasing by 28\% (from 15.320 to 19.674) on SNUFILM-hard and 45\% (from 27.356 to 39.642) on SNUFILM-extreme.

\noindent \textbf{Effect of joint optimization.}
We jointly optimize the encoder-decoder synthesizer and the diffusion models. We report the result without joint optimization in the 2nd row of Table.~\ref{tab:compare_optimization}. The result deteriorates without joint optimization.

\noindent \textbf{Effect of parameter sharing in diffusion models.}
We share network parameters across the hierarchical diffusion models on different levels.
Alternatively, they can use separated network parameters on each level.
We evaluate our method in these two settings, and report the result in the last two rows of Table.~\ref{tab:compare_optimization}.
By sharing network parameters across different levels, we improve the robustness of the diffusion model and decrease the FID error on SNUFIM-hard by 6.6\% (from 16.324 to 15.320).

\begin{table}[]
    \centering
    \setlength{\tabcolsep}{3pt}

\begin{tabular}{ccc|cc|cc}
    \toprule
    \multirow{2}{*}{Flow} &     \multirow{2}{*}{Joint}   & \multirow{2}{*}{Share}& \multicolumn{2}{c|}{SNUFILM-hard}    & \multicolumn{2}{c}{SNUFILM-extreme}\\
    ~                         &  ~                                      & ~ & LPIPS     & FID       & LPIPS     & FID   \\
    \midrule
    -       & \cmark     & \cmark       & 0.0442         & 19.674        & 0.0956            & 39.642           \\
     \cmark     & -            & \cmark & \underline{0.0416}          & \underline{16.158}          & \underline{0.0865}           & \underline{28.767}          \\
     \cmark          & \cmark    & -    &0.0418           & 16.324          & 0.0870              & 29.313      \\
     \cmark     & \cmark       & \cmark & {\bf 0.0405}    & {\bf 15.320}    & {\bf 0.0842}      &{\bf 27.356}           \\
     \bottomrule
\end{tabular}

    \vspace{-3mm}
    \caption{{\bf Ablation studies of different design of our method.}
    ``Share" denotes sharing network parameters across hierarchical diffusion models working on different levels, ``Joint'' denotes jointly fine-tuning the synthesizer and the diffusion model in the last stage, and ``Flow'' denotes modeling the diffusion process to denoise an optical flow or directly denoise the latent space.
    }
    \vspace{-3mm}
    \label{tab:compare_optimization}
\end{table}

\noindent \textbf{Evaluation of different training resolutions.}
We train the diffusion model with an input image resolution of 256$\times$256 by default, and for inference, we resize the input image pair to have the shorter size equal to 256 to extract the condition latent. 
In Table.~\ref{tab:reslution_effect}, we evaluate the effect of input image resolution on our method.
Increasing the input resolution improves performance. 
We also list the running time of each component in our framework for running the diffusion model on different input image resolutions, and compare it with recent diffusion-based methods in Table.~\ref{tab:reslution_effect}.
We measure the runtime on a workstation with an NVIDIA RTX-4090 GPU, and as shown in the table, our method takes only 98 ms for 256$\times$256 resolution.

\begin{figure}
    \centering
    \setlength{\tabcolsep}{2pt}
    \begin{tabular}{cccc}
        \includegraphics[width=0.24\linewidth]{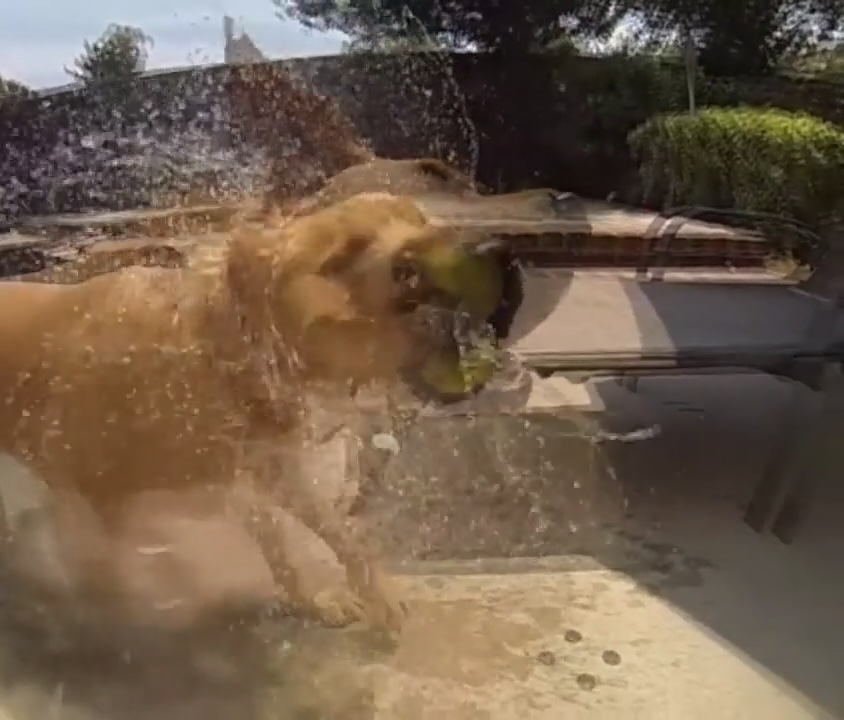} &
        \includegraphics[width=0.24\linewidth]{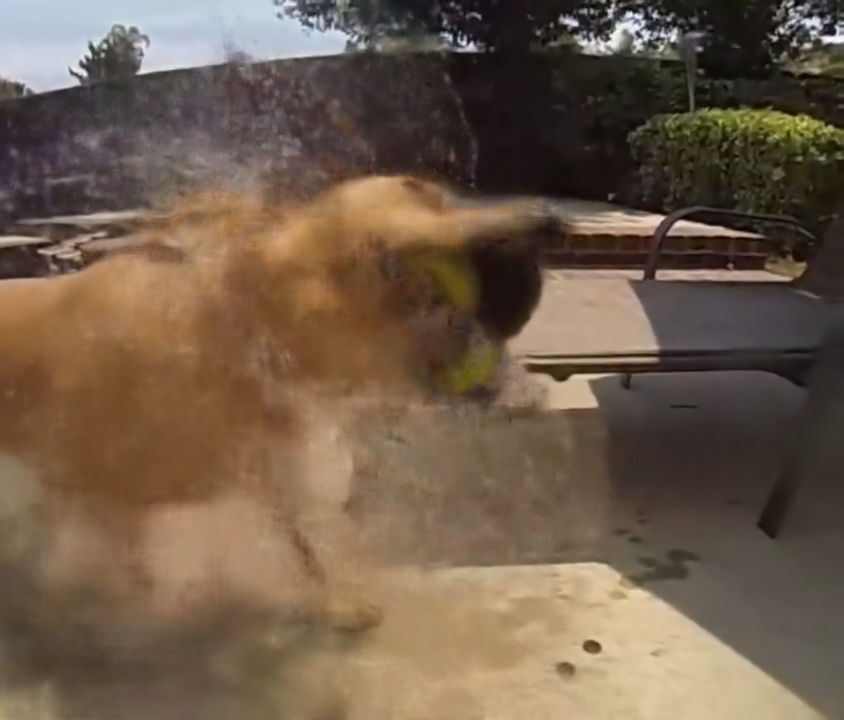} &
        \includegraphics[width=0.24\linewidth]{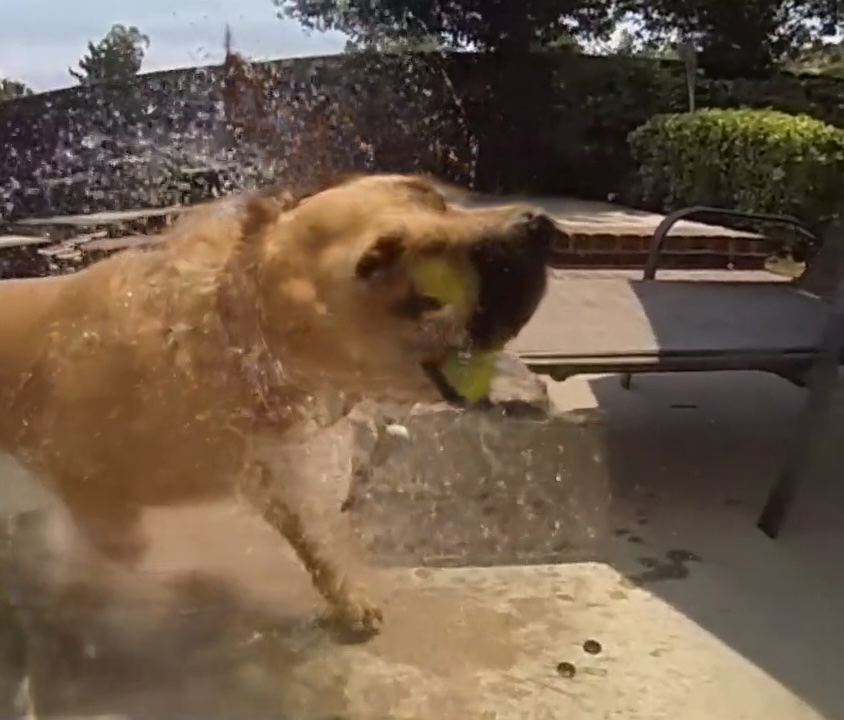} &
        \includegraphics[width=0.24\linewidth]{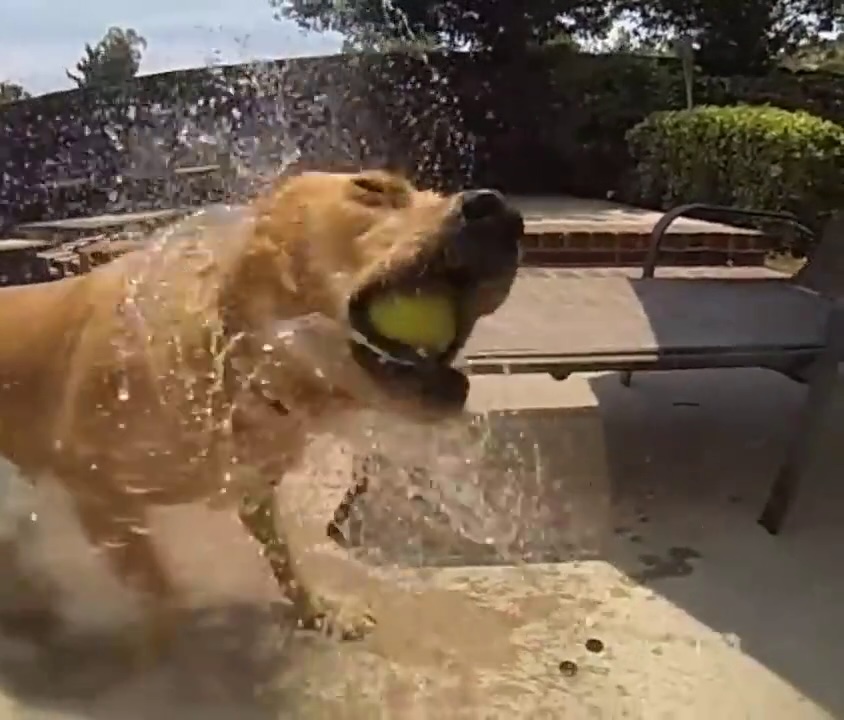} \\

        \includegraphics[width=0.24\linewidth, trim=0 250 70 200, clip]{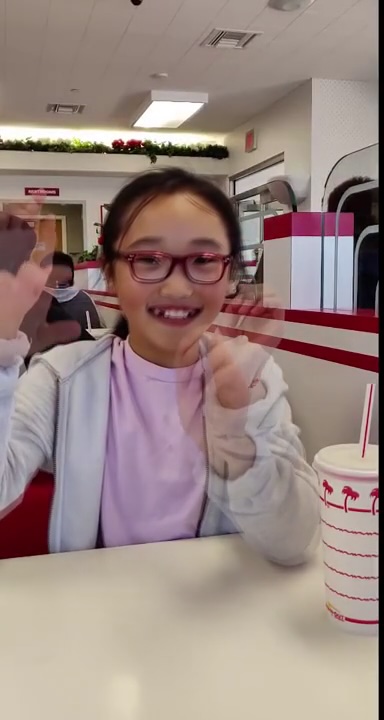} &
        \includegraphics[width=0.24\linewidth, trim=0 250 70 200, clip]{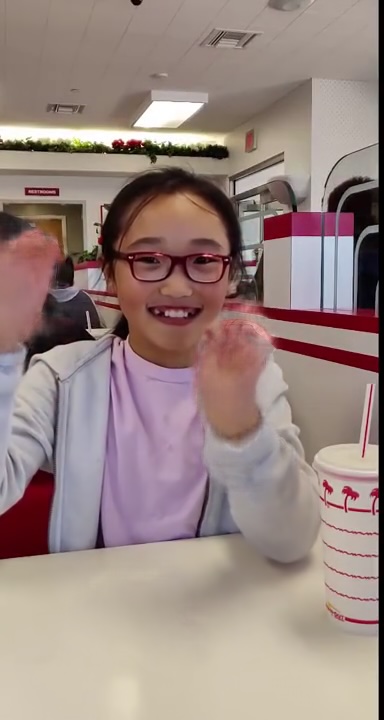} &
        \includegraphics[width=0.24\linewidth, trim=0 250 70 200, clip]{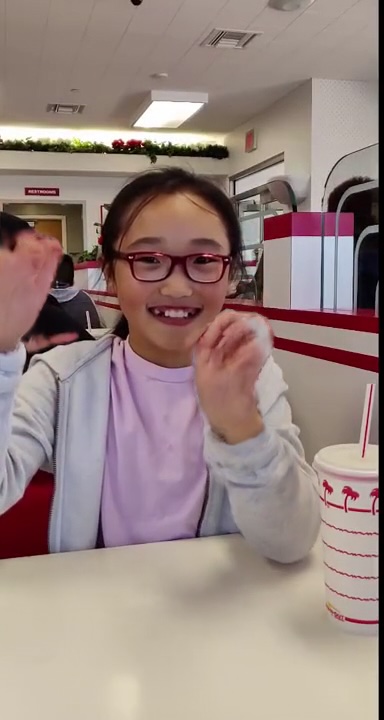} &
        \includegraphics[width=0.24\linewidth, trim=0 250 70 200, clip]{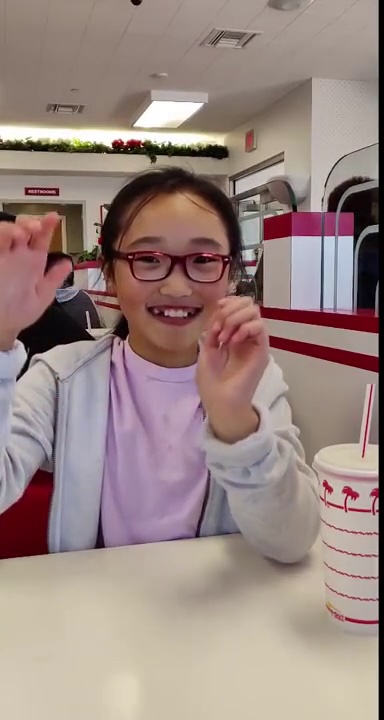} \\
        {\small Input overlay}  & {\small SGM-VFI}      & {\small \bf Ours}     & {\small Ground truth} \\
\end{tabular}
    \vspace{-3mm}
    \caption{{\bf Limitation discussion.} Our method suffers in scenarios with extreme motion patterns. However, it is still better than the state of the art, and can recover most of the details that are missed with SGM-VFI in these cases.}
    \vspace{-3mm}
    \label{fig:failure_case}
\end{figure}

\noindent \textbf{Limitation discussion.}
Our method produces accurate results in scenarios with complex motion and large displacement, while there are still extreme cases where it produces noticeable artifacts, as shown in Fig.~\ref{fig:failure_case}.  While, note that, our method is still better than the state of the art, and can recover most of the details that are missed with SGM-VFI in those cases. We attribute this to the relatively small-scale training data, and plan to collect datasets in larger scale with diverse motion to further improve the performance.

\section{Conclusion}
\label{sec:conclusion}
We have introduced a hierarchical flow diffusion model for video frame interpolation. Instead of formulating frame interpolation as a denoising procedure in the latent space, we proposed to model optical flow explicitly from coarse to fine by hierarchical diffusion models, which has much smaller search space in each denoising step, and can handle complex motions and large displacements. In experiments, our approach demonstrates the effectiveness of the hierarchical diffusion models by generating high-quality interpolated frames, which outperforms state-of-the-art methods, and 10+ times faster than other diffusion-based methods.

\clearpage

{
    \small
    \bibliographystyle{ieeenat_fullname}
    \bibliography{main}

\begin{thebibliography}{56}
\providecommand{\natexlab}[1]{#1}
\providecommand{\url}[1]{\texttt{#1}}
\expandafter\ifx\csname urlstyle\endcsname\relax
  \providecommand{\doi}[1]{doi: #1}\else
  \providecommand{\doi}{doi: \begingroup \urlstyle{rm}\Url}\fi

\bibitem[Black and Anandan(1993)]{black1993framework}
Michael~J Black and Padmanabhan Anandan.
\newblock {A Framework for the Robust Estimation of Optical Flow}.
\newblock In \emph{International Conference on Computer Vision}, 1993.

\bibitem[Butler et~al.(2012)Butler, Wulff, Stanley, and Black]{butler2012sintel}
Daniel~J Butler, Jonas Wulff, Garrett~B Stanley, and Michael~J Black.
\newblock {A Naturalistic Open Source Movie for Optical Flow Evaluation}.
\newblock In \emph{European Conference on Computer Vision}, 2012.

\bibitem[Chen and Koltun(2016)]{Fullflow_2016_cvpr}
Qifeng Chen and Vladlen Koltun.
\newblock {Full Flow: Optical Flow Estimation by Global Optimization over Regular Grids}.
\newblock In \emph{Conference on Computer Vision and Pattern Recognition}, 2016.

\bibitem[Chen et~al.(2023{\natexlab{a}})Chen, Sun, Song, and Luo]{chen2023diffusiondet}
Shoufa Chen, Peize Sun, Yibing Song, and Ping Luo.
\newblock {DiffusionDet: Diffusion Model for Object Detection}.
\newblock In \emph{International Conference on Computer Vision}, 2023{\natexlab{a}}.

\bibitem[Chen et~al.(2023{\natexlab{b}})Chen, ZHANG, and Hinton]{chen2023Analogbits}
Ting Chen, Ruixiang ZHANG, and Geoffrey Hinton.
\newblock {Analog Bits: Generating Discrete Data using Diffusion Models with Self-Conditioning}.
\newblock In \emph{International Conference on Learning Representations}, 2023{\natexlab{b}}.

\bibitem[Choi et~al.(2020)Choi, Kim, Han, Xu, and Lee]{choi2020snufilm}
Myungsub Choi, Heewon Kim, Bohyung Han, Ning Xu, and Kyoung~Mu Lee.
\newblock {Channel Attention is All You Need for Video Frame Interpolation}.
\newblock In \emph{AAAI Conference on Artificial Intelligence}, 2020.

\bibitem[Danier et~al.(2024)Danier, Zhang, and Bull]{danier2024ldmvfi}
Duolikun Danier, Fan Zhang, and David Bull.
\newblock {LDMVFI: Video Frame Interpolation with Latent Diffusion Models}.
\newblock In \emph{AAAI Conference on Artificial Intelligence}, 2024.

\bibitem[Dhariwal and Nichol(2021)]{dhariwal2021diffusion}
Prafulla Dhariwal and Alexander Nichol.
\newblock {Diffusion Models Beat Gans on Image Synthesis}.
\newblock \emph{Advances in Neural Information Processing Systems}, 2021.

\bibitem[Gatys et~al.(2016)Gatys, Ecker, and Bethge]{gatys2016style}
Leon~A Gatys, Alexander~S Ecker, and Matthias Bethge.
\newblock {Image Style Transfer Using Convolutional Neural Networks}.
\newblock In \emph{Conference on Computer Vision and Pattern Recognition}, 2016.

\bibitem[Hai et~al.(2023{\natexlab{a}})Hai, Song, Li, Ferstl, and Hu]{hai2023pfcflow}
Yang Hai, Rui Song, Jiaojiao Li, David Ferstl, and Yinlin Hu.
\newblock {Pseudo Flow Consistency for Self-Supervised 6D Object Pose Estimation}.
\newblock In \emph{International Conference on Computer Vision}, 2023{\natexlab{a}}.

\bibitem[Hai et~al.(2023{\natexlab{b}})Hai, Song, Li, and Hu]{hai2023scflow}
Yang Hai, Rui Song, Jiaojiao Li, and Yinlin Hu.
\newblock {Shape-Constraint Recurrent Flow for 6D Object Pose Estimation}.
\newblock In \emph{Conference on Computer Vision and Pattern Recognition}, 2023{\natexlab{b}}.

\bibitem[He et~al.(2016)He, Zhang, Ren, and Sun]{resnet}
Kaiming He, Xiangyu Zhang, Shaoqing Ren, and Jian Sun.
\newblock {Deep Residual Learning for Image Recognition}.
\newblock In \emph{Conference on Computer Vision and Pattern Recognition}, 2016.

\bibitem[Hertz et~al.(2023)Hertz, Mokady, Tenenbaum, Aberman, Pritch, and Cohen-or]{hertz2023prompttoprompt}
Amir Hertz, Ron Mokady, Jay Tenenbaum, Kfir Aberman, Yael Pritch, and Daniel Cohen-or.
\newblock {Prompt-to-Prompt Image Editing with Cross-Attention Control}.
\newblock In \emph{International Conference on Learning Representations}, 2023.

\bibitem[Heusel et~al.(2017)Heusel, Ramsauer, Unterthiner, Nessler, and Hochreiter]{heusel2017fid}
Martin Heusel, Hubert Ramsauer, Thomas Unterthiner, Bernhard Nessler, and Sepp Hochreiter.
\newblock {GANs Trained by a Two Time-Scale Update Rule Converge to a Local Nash Equilibrium}.
\newblock \emph{Advances in Neural Information Processing Systems}, 30, 2017.

\bibitem[Ho et~al.(2020)Ho, Jain, and Abbeel]{ho2020ddpm}
Jonathan Ho, Ajay Jain, and Pieter Abbeel.
\newblock {Denoising Diffusion Probabilistic Models}.
\newblock \emph{Advances in Neural Information Processing Systems}, 2020.

\bibitem[Hu et~al.(2016)Hu, Song, and Li]{Hu_2016_CVPR}
Yinlin Hu, Rui Song, and Yunsong Li.
\newblock {Efficient Coarse-To-Fine PatchMatch for Large Displacement Optical Flow}.
\newblock In \emph{Conference on Computer Vision and Pattern Recognition}, 2016.

\bibitem[Hu et~al.(2017)Hu, Li, and Song]{hu2017robust}
Yinlin Hu, Yunsong Li, and Rui Song.
\newblock {Robust Interpolation of Correspondences for Large Displacement Optical Flow}.
\newblock In \emph{Conference on Computer Vision and Pattern Recognition}, 2017.

\bibitem[Huang et~al.(2022{\natexlab{a}})Huang, Shi, Zhang, Wang, Cheung, Qin, Dai, and Li]{huang2022flowformer}
Zhaoyang Huang, Xiaoyu Shi, Chao Zhang, Qiang Wang, Ka~Chun Cheung, Hongwei Qin, Jifeng Dai, and Hongsheng Li.
\newblock {FlowFormer: A Transformer Architecture for Optical Flow}.
\newblock In \emph{European Conference on Computer Vision}, 2022{\natexlab{a}}.

\bibitem[Huang et~al.(2022{\natexlab{b}})Huang, Zhang, Heng, Shi, and Zhou]{huang2022rife}
Zhewei Huang, Tianyuan Zhang, Wen Heng, Boxin Shi, and Shuchang Zhou.
\newblock {Real-Time Intermediate Flow Estimation for Video Frame Interpolation}.
\newblock In \emph{European Conference on Computer Vision}, 2022{\natexlab{b}}.

\bibitem[Huang et~al.(2024)Huang, Yu, Yang, Qin, Zheng, Zheng, Zhou, Wang, and Yang]{huang2024madiff}
Zhilin Huang, Yijie Yu, Ling Yang, Chujun Qin, Bing Zheng, Xiawu Zheng, Zikun Zhou, Yaowei Wang, and Wenming Yang.
\newblock {Motion-aware Latent Diffusion Models for Video Frame Interpolation}.
\newblock In \emph{ACM International Conference on Multimedia}, 2024.

\bibitem[Jiang et~al.(2018)Jiang, Sun, Jampani, Yang, Learned-Miller, and Kautz]{jiang2018superslomo}
Huaizu Jiang, Deqing Sun, Varun Jampani, Ming-Hsuan Yang, Erik Learned-Miller, and Jan Kautz.
\newblock {Super SloMo: High Quality Estimation of Multiple Intermediate Frames for Video Interpolation}.
\newblock In \emph{Conference on Computer Vision and Pattern Recognition}, 2018.

\bibitem[Jin et~al.(2023)Jin, Wu, Chen, Chen, Koo, and Hahm]{jin2023urpnet}
Xin Jin, Longhai Wu, Jie Chen, Youxin Chen, Jayoon Koo, and Cheul-hee Hahm.
\newblock {A Unified Pyramid Recurrent Network for Video Frame Interpolation}.
\newblock In \emph{Conference on Computer Vision and Pattern Recognition}, 2023.

\bibitem[Kawar et~al.(2023)Kawar, Zada, Lang, Tov, Chang, Dekel, Mosseri, and Irani]{Kawar2023imagic}
Bahjat Kawar, Shiran Zada, Oran Lang, Omer Tov, Huiwen Chang, Tali Dekel, Inbar Mosseri, and Michal Irani.
\newblock {Imagic: Text-Based Real Image Editing With Diffusion Models}.
\newblock In \emph{Conference on Computer Vision and Pattern Recognition}, 2023.

\bibitem[Ke et~al.(2024)Ke, Obukhov, Huang, Metzger, Daudt, and Schindler]{ke2024marigold}
Bingxin Ke, Anton Obukhov, Shengyu Huang, Nando Metzger, Rodrigo~Caye Daudt, and Konrad Schindler.
\newblock {Repurposing Diffusion-Based Image Generators for Monocular Depth Estimation}.
\newblock In \emph{Conference on Computer Vision and Pattern Recognition}, 2024.

\bibitem[Kong et~al.(2022)Kong, Jiang, Luo, Chu, Huang, Tai, Wang, and Yang]{kong2022ifrnet}
Lingtong Kong, Boyuan Jiang, Donghao Luo, Wenqing Chu, Xiaoming Huang, Ying Tai, Chengjie Wang, and Jie Yang.
\newblock {IFRNet: Intermediate Feature Refine Network for Efficient Frame Interpolation}.
\newblock In \emph{Conference on Computer Vision and Pattern Recognition}, 2022.

\bibitem[Li et~al.(2021)Li, Niklaus, Snavely, and Wang]{li2021neural}
Zhengqi Li, Simon Niklaus, Noah Snavely, and Oliver Wang.
\newblock {Neural Scene Flow Fields for Space-Time View Synthesis of Dynamic Scenes}.
\newblock In \emph{Conference on Computer Vision and Pattern Recognition}, 2021.

\bibitem[Li et~al.(2023)Li, Zhu, Han, Hou, Guo, and Cheng]{li2023amt}
Zhen Li, Zuo-Liang Zhu, Ling-Hao Han, Qibin Hou, Chun-Le Guo, and Ming-Ming Cheng.
\newblock {AMT: All-Pairs Multi-Field Transforms for Efficient Frame Interpolation}.
\newblock In \emph{Conference on Computer Vision and Pattern Recognition}, 2023.

\bibitem[Liu et~al.(2024)Liu, Zhang, Zhao, and Wang]{liu2024sgm-vfi}
Chunxu Liu, Guozhen Zhang, Rui Zhao, and Limin Wang.
\newblock {Sparse Global Matching for Video Frame Interpolation with Large Motion}.
\newblock In \emph{Conference on Computer Vision and Pattern Recognition}, 2024.

\bibitem[Loshchilov and Hutter(2018)]{AdamW_2019_iclr}
Ilya Loshchilov and Frank Hutter.
\newblock {Decoupled Weight Decay Regularization}.
\newblock In \emph{International Conference on Learning Representations}, 2018.

\bibitem[Lu et~al.(2022)Lu, Wu, Lin, Lu, and Jia]{lu2022vfiformer}
Liying Lu, Ruizheng Wu, Huaijia Lin, Jiangbo Lu, and Jiaya Jia.
\newblock {Video Frame Interpolation with Transformer}.
\newblock In \emph{Conference on Computer Vision and Pattern Recognition}, 2022.

\bibitem[Luo et~al.(2024)Luo, Li, Yang, Liu, Fan, and Liu]{luo2024flowdiffuser}
Ao Luo, Xin Li, Fan Yang, Jiangyu Liu, Haoqiang Fan, and Shuaicheng Liu.
\newblock {FlowDiffuser: Advancing Optical Flow Estimation with Diffusion Models}.
\newblock In \emph{Conference on Computer Vision and Pattern Recognition}, 2024.

\bibitem[Lyu et~al.(2024)Lyu, Li, Jiao, and Chen]{lyu2024cbbd}
Zonglin Lyu, Ming Li, Jianbo Jiao, and Chen Chen.
\newblock {Frame Interpolation with Consecutive Brownian Bridge Diffusion}.
\newblock In \emph{ACM International Conference on Multimedia}, 2024.

\bibitem[Mayer et~al.(2016)Mayer, Ilg, Hausser, Fischer, Cremers, Dosovitskiy, and Brox]{mayer2016large}
Nikolaus Mayer, Eddy Ilg, Philip Hausser, Philipp Fischer, Daniel Cremers, Alexey Dosovitskiy, and Thomas Brox.
\newblock {A Large Dataset to Train Convolutional Networks for Disparity, Optical Flow, and Scene Flow Estimation}.
\newblock In \emph{Conference on Computer Vision and Pattern Recognition}, 2016.

\bibitem[Mehl et~al.(2023)Mehl, Schmalfuss, Jahedi, Nalivayko, and Bruhn]{mehl2023spring}
Lukas Mehl, Jenny Schmalfuss, Azin Jahedi, Yaroslava Nalivayko, and Andr{\'e}s Bruhn.
\newblock {Spring: A High-Resolution High-Detail Dataset and Benchmark for Scene Flow, Optical Flow and Stereo}.
\newblock In \emph{Conference on Computer Vision and Pattern Recognition}, 2023.

\bibitem[Montgomery and Lars(1994)]{montgomery1994xiph}
Christopher Montgomery and H Lars.
\newblock Xiph.org video test media (derf's collection).
\newblock \url{https://media.xiph.org/video/derf}, 1994.
\newblock Online resource.

\bibitem[Nam et~al.(2024)Nam, Lee, Kim, Kim, Cho, Kim, and Kim]{nam2024diffmatch}
Jisu Nam, Gyuseong Lee, Sunwoo Kim, Hyeonsu Kim, Hyoungwon Cho, Seyeon Kim, and Seungryong Kim.
\newblock {Diffusion Model for Dense Matching}.
\newblock In \emph{International Conference on Learning Representations}, 2024.

\bibitem[Niklaus and Liu(2020)]{niklaus2020softmax}
Simon Niklaus and Feng Liu.
\newblock {Softmax Splatting for Video Frame Interpolation}.
\newblock In \emph{Conference on Computer Vision and Pattern Recognition}, 2020.

\bibitem[Peebles and Xie(2023)]{peebles2023dit}
William Peebles and Saining Xie.
\newblock {Scalable Diffusion Models with Transformers}.
\newblock In \emph{International Conference on Computer Vision}, 2023.

\bibitem[Perazzi et~al.(2016)Perazzi, Pont-Tuset, McWilliams, Van~Gool, Gross, and Sorkine-Hornung]{perazzi2016davis}
Federico Perazzi, Jordi Pont-Tuset, Brian McWilliams, Luc Van~Gool, Markus Gross, and Alexander Sorkine-Hornung.
\newblock {A Benchmark Dataset and Evaluation Methodology for Video Object Segmentation}.
\newblock In \emph{Conference on Computer Vision and Pattern Recognition}, 2016.

\bibitem[Reda et~al.(2022)Reda, Kontkanen, Tabellion, Sun, Pantofaru, and Curless]{reda2022film}
Fitsum Reda, Janne Kontkanen, Eric Tabellion, Deqing Sun, Caroline Pantofaru, and Brian Curless.
\newblock {FILM: Frame Interpolation for Large Motion}.
\newblock In \emph{European Conference on Computer Vision}, 2022.

\bibitem[Rombach et~al.(2022)Rombach, Blattmann, Lorenz, Esser, and Ommer]{rombach2022stablediffusion}
Robin Rombach, Andreas Blattmann, Dominik Lorenz, Patrick Esser, and Bj{\"o}rn Ommer.
\newblock {High-Resolution Image Synthesis with Latent Diffusion Models}.
\newblock In \emph{Conference on Computer Vision and Pattern Recognition}, 2022.

\bibitem[Saxena et~al.(2024)Saxena, Herrmann, Hur, Kar, Norouzi, Sun, and Fleet]{saxena2024ddvm}
Saurabh Saxena, Charles Herrmann, Junhwa Hur, Abhishek Kar, Mohammad Norouzi, Deqing Sun, and David~J Fleet.
\newblock {The Surprising Effectiveness of Diffusion Models for Optical Flow and Monocular Depth Estimation}.
\newblock \emph{Advances in Neural Information Processing Systems}, 2024.

\bibitem[Smith and Topin(2019)]{onecycle}
Leslie~N Smith and Nicholay Topin.
\newblock {Super-Convergence: Very Fast Training of Neural Networks Using Large Learning Rates}.
\newblock In \emph{Artificial intelligence and machine learning for multi-domain operations applications}. SPIE, 2019.

\bibitem[Song et~al.(2021)Song, Meng, and Ermon]{song2021ddim}
Jiaming Song, Chenlin Meng, and Stefano Ermon.
\newblock {Denoising Diffusion Implicit Models}.
\newblock In \emph{International Conference on Learning Representations}, 2021.

\bibitem[Sun et~al.(2018)Sun, Yang, Liu, and Kautz]{sun2018pwc}
Deqing Sun, Xiaodong Yang, Ming-Yu Liu, and Jan Kautz.
\newblock {PWC-Net: CNNs for Optical Flow Using Pyramid, Warping, and Cost Volume}.
\newblock In \emph{Conference on Computer Vision and Pattern Recognition}, 2018.

\bibitem[Sun et~al.(2021)Sun, Vlasic, Herrmann, Jampani, Krainin, Chang, Zabih, Freeman, and Liu]{sun2021autoflow}
Deqing Sun, Daniel Vlasic, Charles Herrmann, Varun Jampani, Michael Krainin, Huiwen Chang, Ramin Zabih, William~T Freeman, and Ce Liu.
\newblock {AutoFlow: Learning a Better Training Set for Optical Flow}.
\newblock In \emph{Conference on Computer Vision and Pattern Recognition}, 2021.

\bibitem[Teed and Deng(2020)]{teed2020raft}
Zachary Teed and Jia Deng.
\newblock {RAFT: Recurrent All-Pairs Field Transforms for Optical Flow}.
\newblock In \emph{European Conference on Computer Vision}, 2020.

\bibitem[Voleti et~al.(2022)Voleti, Jolicoeur-Martineau, and Pal]{voleti2022mcvd}
Vikram Voleti, Alexia Jolicoeur-Martineau, and Chris Pal.
\newblock {MCVD - Masked Conditional Video Diffusion for Prediction, Generation, and Interpolation}.
\newblock \emph{Advances in Neural Information Processing Systems}, 2022.

\bibitem[Wang et~al.(2024)Wang, Lipson, and Deng]{wang2024sea}
Yihan Wang, Lahav Lipson, and Jia Deng.
\newblock {SEA-RAFT: Simple, Efficient, Accurate RAFT for Optical Flow}.
\newblock \emph{European Conference on Computer Vision}, 2024.

\bibitem[Wu et~al.(2018)Wu, Singhal, and Krahenbuhl]{wu2018video}
Chao-Yuan Wu, Nayan Singhal, and Philipp Krahenbuhl.
\newblock {Video Compression through Image Interpolation}.
\newblock In \emph{European Conference on Computer Vision}, 2018.

\bibitem[Wu et~al.(2024)Wu, Tao, Li, Wang, Liu, and Zheng]{wu2024pervfi}
Guangyang Wu, Xin Tao, Changlin Li, Wenyi Wang, Xiaohong Liu, and Qingqing Zheng.
\newblock {Perception-Oriented Video Frame Interpolation via Asymmetric Blending}.
\newblock In \emph{Conference on Computer Vision and Pattern Recognition}, 2024.

\bibitem[Xue et~al.(2019)Xue, Chen, Wu, Wei, and Freeman]{xue2019vimeo}
Tianfan Xue, Baian Chen, Jiajun Wu, Donglai Wei, and William~T Freeman.
\newblock {Video Enhancement with Task-Oriented Flow}.
\newblock \emph{International Journal of Computer Vision}, 127:\penalty0 1106--1125, 2019.

\bibitem[Yue et~al.(2024)Yue, Wang, and Loy]{yue2024resshift}
Zongsheng Yue, Jianyi Wang, and Chen~Change Loy.
\newblock {Efficient Diffusion Model for Image Restoration by Residual Shifting}.
\newblock \emph{Advances in Neural Information Processing Systems}, 2024.

\bibitem[Zhang et~al.(2023)Zhang, Zhu, Wang, Chen, Wu, and Wang]{zhang2023ema_vfi}
Guozhen Zhang, Yuhan Zhu, Haonan Wang, Youxin Chen, Gangshan Wu, and Limin Wang.
\newblock {Extracting Motion and Appearance via Inter-Frame Attention for Efficient Video Frame Interpolation}.
\newblock In \emph{Conference on Computer Vision and Pattern Recognition}, 2023.

\bibitem[Zhang et~al.(2018)Zhang, Isola, Efros, Shechtman, and Wang]{zhang2018lpips}
Richard Zhang, Phillip Isola, Alexei~A Efros, Eli Shechtman, and Oliver Wang.
\newblock {The Unreasonable Effectiveness of Deep Features as a Perceptual Metric}.
\newblock In \emph{Conference on Computer Vision and Pattern Recognition}, 2018.

\bibitem[Zhou et~al.(2023)Zhou, Yang, and Yang]{zhou2023pyramid}
Dewei Zhou, Zongxin Yang, and Yi Yang.
\newblock {Pyramid Diffusion Models For Low-light Image Enhancement}.
\newblock In \emph{International Joint Conference on Artificial Intelligence}, 2023.

\end{thebibliography}
}

\end{document}


\maketitle



\noindent \textbf{Evaluation on Middlebury.} 
We compare our method against state-of-the-art methods on Middlebury, and report the results in Table~\ref{tab:compare_middlebury}. In addition to the metrics LPIPS and FID used in the main paper, we report numbers in PSNR and SSIM on this dataset. To evaluate the efficiency, we run most of the methods on the same workstation with an NVIDIA RTX-4090 GPU, and report the inference time in seconds with a normalized image resolution 1024$\times$1024. Our method outperforms most methods significantly in LPIPS and FID, and 10+ times faster than other diffusion-based methods. Furthermore, our method is on par with other non-diffusion-based methods in efficiency, while produces much more accurate results. Note that, although most other methods are better in PSNR and SSIM in the table, these two metrics do not align well with the perceptual quality, as illustrated in Fig.~\ref{fig:psnr_vs_lpips}.

\begin{figure}[h]
    \centering
    \setlength{\tabcolsep}{2pt}
    \begin{tabular}{ccc}
        \includegraphics[width=0.31\linewidth]{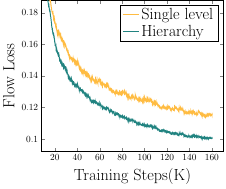} &
        \includegraphics[width=0.31\linewidth]{figures/loss_compare_scale8.pdf} &
        \includegraphics[width=0.31\linewidth]{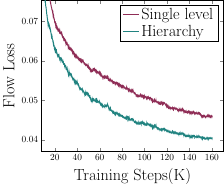} \\
        {\small (a) scale-1/4}  & {\small (b) scale-1/8}       & {\small (c) scale-1/16} \\
    \end{tabular}
    \vspace{-2mm}
    \caption{
    {\bf Training analysis of hierarchical diffusion model.} The proposed hierarchical flow diffusion model has better convergence compared with versions with a fixed resolution on each pyramid level only, which helps to produce better interpolation results with our flow-guided image synthesizer.
    }
    \vspace{-3mm}
    \label{fig:loss_vis}
\end{figure}

\noindent \textbf{Training analysis of hierarchical diffusion.}
We report the flow loss value of our hierarchical diffusion model during the training on Vimeo-90k, and compare it with three versions with a fixed resolution on each pyramid level only, as shown in Fig.~\ref{fig:loss_vis}. It shows that the proposed hierarchical diffusions model converges better than all the single-level versions, which helps to produce better interpolation results with our flow-guided image synthesizer.


\begin{table}[h]
    \centering
    \resizebox{1.0\linewidth}{!}{
    \begin{tabular}{l|ccccc}
    \toprule
    Methods      &  PSNR$\uparrow$ & SSIM $\uparrow$         & LPIPS$\downarrow$         & FID $\downarrow$ & Time (s)\\
    \midrule
    VFIformer   & {\bf 38.438}  & 0.986              & 0.0310         & 15.634    & 3.27\\
    EMA-VFI     & 38.320        & \underline{0.987}                 & 0.0151    & 9.439          & 0.56\\
    AMT-G       & \underline{38.395}& {\bf 0.988}                & 0.0151    & 7.895         & \textbf{0.17} \\
    SGM-VFI     & 37.931        & \underline{0.987}                     & 0.0177        & 8.983        & \underline{0.19}\\
    MADIFF      & 34.180        & 0.974                 & 0.0160    & 11.678            & - \\
    Per-VFI     & 34.344        & 0.976                        & 0.0153    & 11.429         & 0.45   \\  
    \midrule
    LDMVFI      & 34.230        & 0.974                & 0.0195    & 16.167               & 8.32\\
    CBBD        & 36.852        & 0.983                      & \underline{0.0093}            & \underline{8.101} & 2.07 \\ 
    {\bf Ours}  & 37.136        & 0.983                      &{\bf 0.0089}    & {\bf 6.961} & 0.20 \\
    \bottomrule
\end{tabular}
    }
     \vspace{-2mm}
    \caption{{\bf Results on Middlebury.}
    Our method is 10+ time faster than diffusion-based methods (second group), and on par with non-diffusion-based methods (first group) in efficiency. Moreover, our method outperforms all methods in LPIPS and FID, including both diffusion based and non-diffusion based methods.
    }
    \label{tab:compare_middlebury}
\end{table}



\begin{figure}[h]
    \centering
    \setlength{\tabcolsep}{3pt}

\begin{tabular}{cc}

    \includegraphics[width=0.49\linewidth, trim=100 250 650 0, clip]{figures/compare_results/snufilm_extreme/191/sgmvfi_191.png} &
    \includegraphics[width=0.49\linewidth, trim=100 250 650 0, clip]{figures/compare_results/snufilm_extreme/191/ours_191.png} \\  [-3pt]
    {\small \textcolor{purple}{PSNR:24.999}, \textcolor{purple}{SSIM:0.792}}  &
    {\small \textcolor{purple}{PSNR:23.880},  \textcolor{purple}{SSIM:0.755}} \\ [-2pt]
    {\small \textcolor{cyan}{LPIPS:0.0659}, \textcolor{cyan}{FID:20.629}}  &
    {\small \textcolor{cyan}{LPIPS:0.0462}, \textcolor{cyan}{FID:18.322}} \\ [2pt]

    \includegraphics[width=0.49\linewidth, trim=0 125 0 325, clip]{figures/compare_results/snufilm_extreme/121/sgmvfi_121.png} &
    \includegraphics[width=0.49\linewidth, trim=0 125 0 325, clip]{figures/compare_results/snufilm_extreme/121/ours_121.png} \\[-3pt]
    {\small \textcolor{purple}{PSNR:28.348}, \textcolor{purple}{SSIM:0.972}}  &
    {\small \textcolor{purple}{PSNR:27.314}, \textcolor{purple}{SSIM:0.969}} \\ [-2pt]
    {\small \textcolor{cyan}{LPIPS:0.0291}, \textcolor{cyan}{FID:20.629}}  &
    {\small \textcolor{cyan}{LPIPS:0.0224}, \textcolor{cyan}{FID:18.322}} \\ [2pt]
     {\small SGM-VFI}        & {\small \bf Ours}  \\
\end{tabular}
    \vspace{-0.5em}
    \caption{
    {\bf Problems with PSNR and SSIM.} We show two example results of SGM-VFI and our method, and report the numbers in different metrics accordingly. Although the results of SGM-VFI have better numbers in PSNR$\uparrow$ and SSIM$\uparrow$, their perceptual quality is worse (Note the blurry tail of the plane in the first row, and the problematic wood stick in the second example). LPIPS$\downarrow$ and FID$\downarrow$ align better with the perceptual quality (We compute FID based on an image set containing only the current two examples).
    }
    \label{fig:psnr_vs_lpips}
\end{figure}

\noindent \textbf{More visualization.} We show more qualitative results in Fig.~\ref{fig:more_vis_1} and~\ref{fig:more_vis_2}. As shown in the figure, our method is superior in handling large movements, recovering subtle details, and producing high-quality frame interpolation results.

\begin{figure*}
    \centering
    \setlength{\tabcolsep}{1pt}
    \input{figures/compare_result_supp_snufilm}
    \vspace{-3mm}
    \caption{{\bf More qualitative results on SNU-FILM.} Our method is superior in handling large movements, recovering subtle details, and producing high-quality frame interpolation results.}
    \label{fig:more_vis_1}
\end{figure*}

\begin{figure*}
    \centering
    \setlength{\tabcolsep}{1pt}
    \input{figures/compare_result_davis_supp}
    \vspace{-3mm}
    \caption{{\bf More qualitative results on DAVIS.} Our method is superior in handling large movements, recovering subtle details, and producing high-quality frame interpolation results.}
    \label{fig:more_vis_2}
\end{figure*}
